\documentclass[10pt,twocolumn,letterpaper]{article}

\usepackage[pagenumbers]{cvpr} 

\usepackage[dvipsnames]{xcolor}

\renewcommand{\paragraph}[1]{{\vspace{1mm}\noindent\textbf{#1}\xspace}}

\definecolor{cvprblue}{rgb}{0.21,0.49,0.74}
\usepackage[pagebackref,breaklinks,colorlinks,citecolor=cvprblue]{hyperref}
\usepackage{multirow}
\usepackage{tabularx}
\usepackage{array}
\usepackage{colortbl}
\usepackage{xcolor}
\usepackage{seqsplit}
\usepackage{graphicx}
\usepackage{booktabs}
\usepackage{pifont}
\usepackage{amssymb}
\usepackage{amsmath}

\DeclareMathOperator*{\argmax}{arg\,max}

\def\paperID{12260} 
\def\confName{CVPR}
\def\confYear{2024}

\title{{UniIR}\includegraphics[width=1.25em]{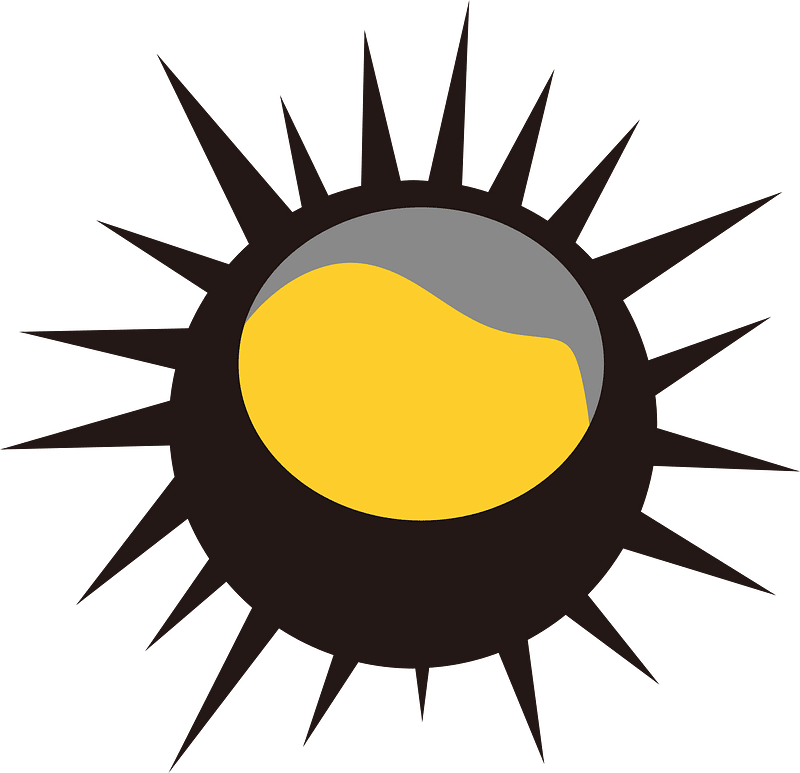}: Training and Benchmarking Universal Multimodal \\Information Retrievers}

\author{\normalsize \textbf{Cong Wei$^{\dagger}$, Yang Chen$^{\ddagger}$, Haonan Chen$^{\dagger}$, Hexiang Hu$^{\circ}$, Ge Zhang$^{\dagger}$, Jie Fu$^{\mathsection}$, Alan Ritter$^{\ddagger}$, Wenhu Chen$^{\dagger}$} \\
{\small$^{\dagger}$University of Waterloo\,
$^{\ddagger}$Georgia Institute of Technology\,
$^{\mathsection}$Hong Kong University of Science and Technology\,
$^{\circ}$Google DeepMind}
}

\begin{document}
\twocolumn[{%
    \renewcommand\twocolumn[1][]{#1}%
    \maketitle
    \centering
    \vspace{-8mm}
    \url{https://tiger-ai-lab.github.io/UniIR/}
    \vspace{2mm}
    \begin{center}
        \centering
        \includegraphics[width=0.87\textwidth]{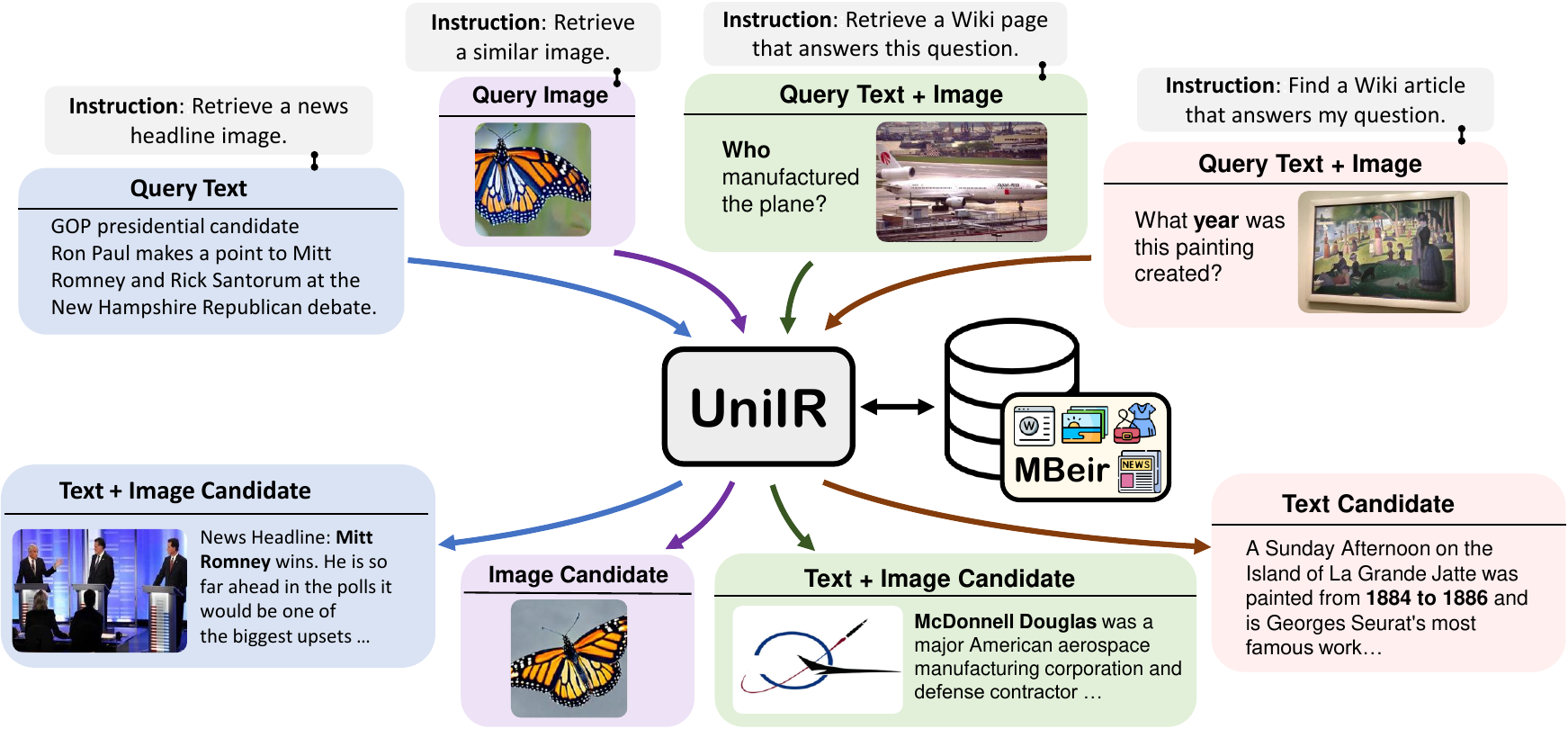}
        \captionof{figure}{We build a universal multimodal information retriever UniIR through instruction tuning. UniIR is capable of accepting any form of query and instruction to retrieve information in any modality.}
        \label{fig:teaser}
    \end{center}
}]
\begin{abstract}
Existing information retrieval (IR) models often assume a homogeneous format, limiting their applicability to diverse user needs, such as searching for images with text descriptions, searching for a news article with a headline image, or finding a similar photo with a query image. 
To approach such different information-seeking demands, we introduce UniIR, a unified instruction-guided multimodal retriever capable of handling eight distinct retrieval tasks across modalities. 
UniIR, a single retrieval system jointly trained on ten diverse multimodal-IR datasets, interprets user instructions to execute various retrieval tasks, demonstrating robust performance across existing datasets and zero-shot generalization to new tasks. 
Our experiments highlight that multi-task training and instruction tuning are keys to UniIR's generalization ability. 
Additionally, we construct the M-BEIR, a multimodal retrieval benchmark with comprehensive results, to standardize the evaluation of universal multimodal information retrieval.
\end{abstract}

\section{Introduction}
\label{sec:intro}
Information retrieval (IR) is a pivotal task that involves sourcing relevant information from vast data collections to meet specific user requirements~\cite{singhal2001modern}. This process has become increasingly important with the advent of generative AI~\cite{yasunaga2023retrieval,blattmann2022semiparametric, sheynin2023knndiffusion,chen2022re}, as it not only enables precise attribution but also mitigates the risk of inaccuracies and fabrications in generated content~\cite{ram2023context,asai2023retrieval}. Despite the crucial role of IR in the current technological landscape, much of the existing literature—particularly within the realm of multimodal IR—remains narrow in scope, focusing mainly on homogeneous retrieval scenarios with pre-defined format, and oftentimes within a single domain. For example, MSCOCO~\cite{lin2014microsoft} considers retrieving Flickr images via text caption, while EDIS~\cite{liu2023edis} considers retrieving news headline images with news title. Such a homogeneous setting is insufficient to accommodate users' diverse information-seeking needs, which often transcends domains and modalities. 
For instance, while some users may search for web images through textual queries, others might use a photo of a dress along with text input like ``similar styles'' or ``color in red'' to find similar fashion products for that specific dress.
The current suite of multimodal retrieval systems falls short in its capacity to accommodate these diverse user demands, limited to task-specific fine-tuning of a pre-trained CLIP~\cite{radford2021learning} model. In recognition of these limitations, a compelling need arises to conceptualize and develop a more flexible, general-purpose neural retriever that bridges different domains, modalities, and retrieval tasks to serve the diverse needs of users.

In this paper, we propose the UniIR framework to learn a single retriever to accomplish (possibly) any retrieval task. Unlike traditional IR systems, UniIR needs to follow the instructions to take a heterogeneous query to retrieve from a heterogeneous candidate pool with millions of candidates in diverse modalities. 
To train UniIR models, we construct M-BEIR, a benchmark of instruction following multimodal retrieval tasks building on existing 10 diverse datasets and unifying their queries and targets in a unified task formulation. The query instructions are curated to define the user's retrieval intention, thereby guiding the information retrieval process.
We train different UniIR models based on pre-trained vision-language models like CLIP~\cite{radford2021learning} and BLIP~\cite{li2022blip} on 300K training instances in M-BEIR with different multimodal fusion mechanisms (score-level fusion and feature-level fusion). We show that UniIR models are able to follow instructions precisely to retrieve desired targets from a heterogeneous candidate pool. Our best UniIR model is based on CLIP with score fusion, which not only achieves very competitive results on fine-tuned datasets but also generalizes to held-out datasets (Figure~\ref{fig:union_held_out_local}). 
Our ablation study reveals two insights: (1) Multi-task training in UniIR(BLIP) is beneficial, which leads to +9.7 improvement in terms of recall@5 over single-task training (Table~\ref{tab:multitask}); (2) Instruction tuning is critical to help models generalize to unseen retrieval datasets and leads to +10 improvement in terms of recall@5 (Figure~\ref{fig:union_held_out}). 

\paragraph{Our contributions} are summarized as follows:
\begin{itemize}
\item UniIR Framework: A universal multimodal information retrieval framework designed to integrate various multimodal retrieval tasks into a cohesive system.

\item M-BEIR: A large-scale multimodal retrieval benchmark that assembles 10 diverse datasets from multiple domains, encompassing 8 distinct multimodal retrieval tasks.

\item We introduce UniIR models, which are universal retrievers trained on M-BEIR, setting a foundational baseline for future research. Additionally, we evaluated the zero-shot performance of SOTA vision-language pre-trained models on the M-BEIR benchmark.

\end{itemize}

\section{UniIR Framework}
\label{sec:task}
\begin{table*}
    \centering
    \small
    \tabcolsep 5pt
    \resizebox{0.99\textwidth}{!}{%
    \begin{tabular}{@{}lllcr@{\;\;}r@{\;\;}r@{\;\;}r@{}}
    \toprule
        \textbf{Task} \tiny{(query $\to$ candidate)}        & \textbf{Dataset}             & \textbf{Instruction} (shown 1 out of 4)                                             & \textbf{Domain} & \textbf{Train} & \textbf{Dev} & \textbf{Test} & \textbf{Pool} \\
        \midrule
         \multirow{3}{*}{1. $q_t \to c_i$} 
         & VisualNews~\citep{liu2020visual}                 & Identify news-related image match with the description          & News      & 99K    & 20K        & 20K    & 542K   \\
         & MSCOCO~\citep{lin2014microsoft}                  & Find an everyday image match with caption                      & Misc.     & 100K   & 24.8K        & 24.8K    & 5K   \\
         & Fashion200K~\citep{han2017automatic}             & Based on fashion description, retrieve matched image             & Fashion   & 15K    & 1.7K         & 1.7K     & 201K   \\
         \midrule
         2. $q_{t} \to c_t$ & WebQA~\citep{chang2022webqa}   & Find an paragraph from Wikipedia to answer the question          & Wiki      & 16K    & 1.7K         & 2.4K     & 544K   \\
         \midrule
         \multirow{2}{*}{3. $q_t \to (c_i, c_t)$}
         & EDIS~\citep{liu2023edis}                         & Find a news image matching with the caption                     & News      & 26K    & 3.2K         & 3.2K     & 1M \\
         & WebQA~\citep{chang2022webqa}                     & Find a Wiki image that answer the question                      & Wiki      & 17K    & 1.7K         & 2.5K     & 403K   \\
         \midrule
         \multirow{3}{*}{4. $q_i \to c_t$}
         & VisualNews~\citep{liu2020visual}                 & Provide a news-related caption for the displayed image               & News      & 100K   & 20K        & 20K    & 537K   \\
         & MSCOCO~\citep{lin2014microsoft}                  & Find a caption describe the an image                             & Misc.     & 113K   & 5K         & 5K     & 25K   \\
         & Fashion200K~\citep{han2017automatic}             & Find a description for the fashion item in the image            & Fashion   & 15K    & 4.8K         & 4.8K     & 61K   \\
         \midrule
         5. $q_i  \to c_i$ & NIGHTS~\citep{fu2023learning}  & Find an image that is identical to the given image              & Misc.     & 16K    & 2K         & 2K     & 40K    \\
         \midrule
         \multirow{2}{*}{6. $(q_i, q_t) \to c_t$} 
         & OVEN~\citep{hu2023open}                          & Retrieve a Wiki text that answer the given query about the image  & Wiki      & 150K   & 50K        & 50K    & 676K   \\
         & InfoSeek~\citep{chen2023infoseek}                & Find an article that answers the given question about the image & Wiki      & 141K  & 11K        & 11K    & 611K \\
         \midrule
         \multirow{2}{*}{7. $(q_i, q_t) \to c_i$} 
         & FashionIQ~\citep{wu2021fashion}                  & Find an image to match the fashion image and style note          & Fashion   & 16K    & 2K      & 6K         & 74K    \\
         & CIRR~\citep{liu2021image}                        &  I'm looking for a similar everyday image with the described changes        & Misc.     & 26K    & 2K      & 4K         & 21K    \\
         \midrule
         \multirow{2}{*}{8. $(q_i, q_t) \to (c_i, c_t)$} 
         & OVEN~\citep{hu2023open}                          & Find a Wiki image-text pair to answer a question regarding an image    & Wiki      & 157K   & 14.7K      & 14.7K         & 335K   \\
         & InfoSeek~\citep{chen2023infoseek}                & Find a Wiki image-text pair to answers my question about this image    & Wiki      & 143K   & 17.6K      & 17.6K         & 481K \\
         \midrule
                       & 10 datasets & 64 instructions & 4 domains & 1.1M & 182K & 190K & 5.6M \\
         \bottomrule
    \end{tabular}
    }
    
    \caption{The overview of M-BEIR training/validation/test set. More detailed query instruction design can be found in Appendix.} 
    \label{tab:dataset}
    \vspace{-2pt}
\end{table*}
\subsection{Problem Definition}
In a universal multimodal search engine, users can initiate various search tasks based on their specific needs. These tasks involve different types of queries and retrieval candidates. The query $\mathbf{q}$ could be in text $q_{\text{t}}$, image $q_{\text{i}}$ or even image-text pair $(q_{\text{i}}, q_{\text{t}})$, while the retrieval candidate $\mathbf{c}$ could also be text $c_{\text{t}}$, image $c_{\text{i}}$ or an image-text pair $(c_{\text{i}}, c_{\text{t}})$. Eight existing retrieval tasks are being defined in Table~\ref{tab:dataset}. Please note that the compositional query, $(q_i, q_t)$, typically involves a text-based question $q_t$ about an image $q_i$. On the other hand, a compositional target, $(c_i, c_t)$, usually includes an image $c_i$ accompanied by a descriptive text $c_t$, providing contextual information.

To accommodate different retrieval intentions, we introduce a language task instruction $q_{\texttt{inst}}$ to represent the intention of the retrieval task. This instruction clearly defines what the search aims to find, whether seeking images, text, or a mix of both, and specifies the relevant domain. Further information can be found in Section~\ref{sec:dataset}. More formally, we aim to build a unified retriever model $f$ capable of taking any type of query to retrieve any type of target specified by the instruction $q_{\texttt{inst}}$:
\begin{align*}
    c^{\ast} = \argmax_{\mathbf{\{c\}} \in \mathcal{C}} [f(\mathbf{q}, q_{\texttt{inst}})^T \cdot f(\mathbf{c})]
\end{align*}
Here, $\mathcal{C}$ denotes the heterogeneous candidate pool, $f(\cdot)$ is the function we are optimizing for maximum dot-product retrieval, and $c^{\ast}$ is the predicted result.

By including task instructions, we unify different multimodal retrieval tasks into a single framework, thus enabling us to build a general-purpose multimodal retriever.
Furthermore, instruction fine-tuned language models have shown the capability to perform zero-shot generalization to unseen tasks by following instructions. 
However, applying this concept of zero-shot generalization to the multimodal retrieval domain faces challenges due to the lack of existing datasets tailored for this purpose. To address this gap, we are creating a comprehensive, unified dataset named M-BEIR, which is detailed in Section \ref{sec:dataset}. M-BEIR will serve as a foundational resource for exploring and advancing the capabilities of multi-modal retrieval models.

\subsection{UniIR Model}
In this section, we present the UniIR model, our unified multimodal information retrieval system. The UniIR model is adept at handling distinct retrieval tasks simultaneously. We experimented with two multimodal fusion mechanisms for UniIR, namely score-level fusion~\cite{liu2022universal} and feature-level fusion~\cite{hu2023open, liu2023edis}. To explore the effectiveness of these approaches, we adapted pre-trained models such as CLIP~\cite{radford2021learning} and BLIP~\cite{li2022blip} for our purposes as follows. \vspace{1ex}\\

\begin{figure}
  \centering
  \includegraphics[width=0.9\linewidth]{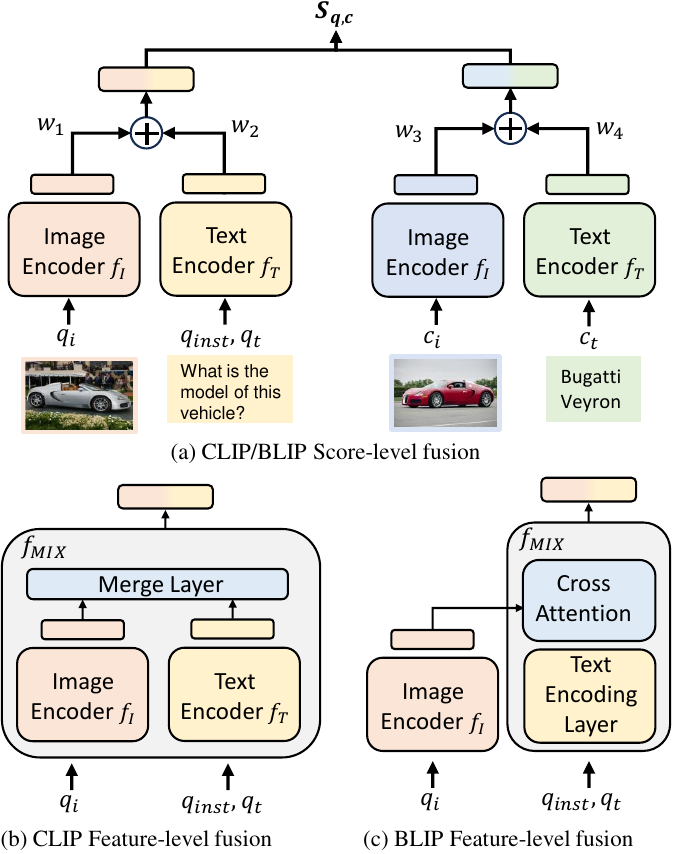} 
  \caption{(a) Score-level fusion encodes each modality into a single feature; (b) CLIP feature-level fusion (CLIP$_{FF}$) fuses two modalities into a single feature with a mix-modality transformer layer; (c) BLIP feature-level fusion (BLIP$_{FF}$) adopts cross-attention to output a single feature vector.}
  \label{fig:model}
  \vspace{-2ex}
\end{figure}

\paragraph{Score-level Fusion.} As illustrated in Figure~\ref{fig:model}(a), the score-level fusion variants for CLIP and BLIP (denoted as CLIP$_{\text{SF}}$ and BLIP$_{\text{SF}}$) employ distinct encoders for vision and text. Specifically, the vision encoder is marked as $f_i$ and the uni-modal text encoder as $f_t$.  In these methods, both image and text inputs (whether from a query or a target) are processed into two individual vectors. These vectors undergo a weighted sum to form a unified representation vector. This process is mathematically represented as $f(q_{\text{i}}, q_{\text{t}}, q_{\texttt{inst}})=w_1 f_I(q_{\text{i}}) + w_2 f_T (q_{\text{t}},q_{\texttt{inst}})$ for queries and $f(c_{\text{i}}, c_{\text{t}})=w_3 f_I(c_{\text{i}}) + w_4 f_T (c_{\text{t}})$ for targets. Therefore, the similarity score between a query $\mathbf{q}$ and a target $\mathbf{c}$ is calculated as a weighted sum of the within-modality and cross-modality similarity scores:
\begin{align*}
s_{\mathbf{q},\mathbf{c}} &= f(q_{\text{i}}, q_{\text{t}}, q_{\texttt{inst}})^T \cdot f(c_{\text{i}}, c_{\text{t}}) \\
                          &= w_1w_3 f_I(q_{\text{i}})^Tf_I(c_{\text{i}}) + w_2w_4 f_T(q_{\text{t}}, q_{\texttt{inst}})^Tf_T(c_{\text{t}}) \\
                          &+ w_1w_4 f_I(q_{\text{i}})^Tf_T(c_{\text{t}}) + w_2w_3 f_T(q_{\text{t}}, q_{\texttt{inst}})^Tf_I(c_{\text{i}})
\end{align*}
$w_1, w_2, w_3, w_4$ is a set of learnable parameters that reflects importance weights. \vspace{1ex}

\begin{figure*}[t!]
  \centering
  \vspace{-5mm}
  \includegraphics[width=\textwidth]{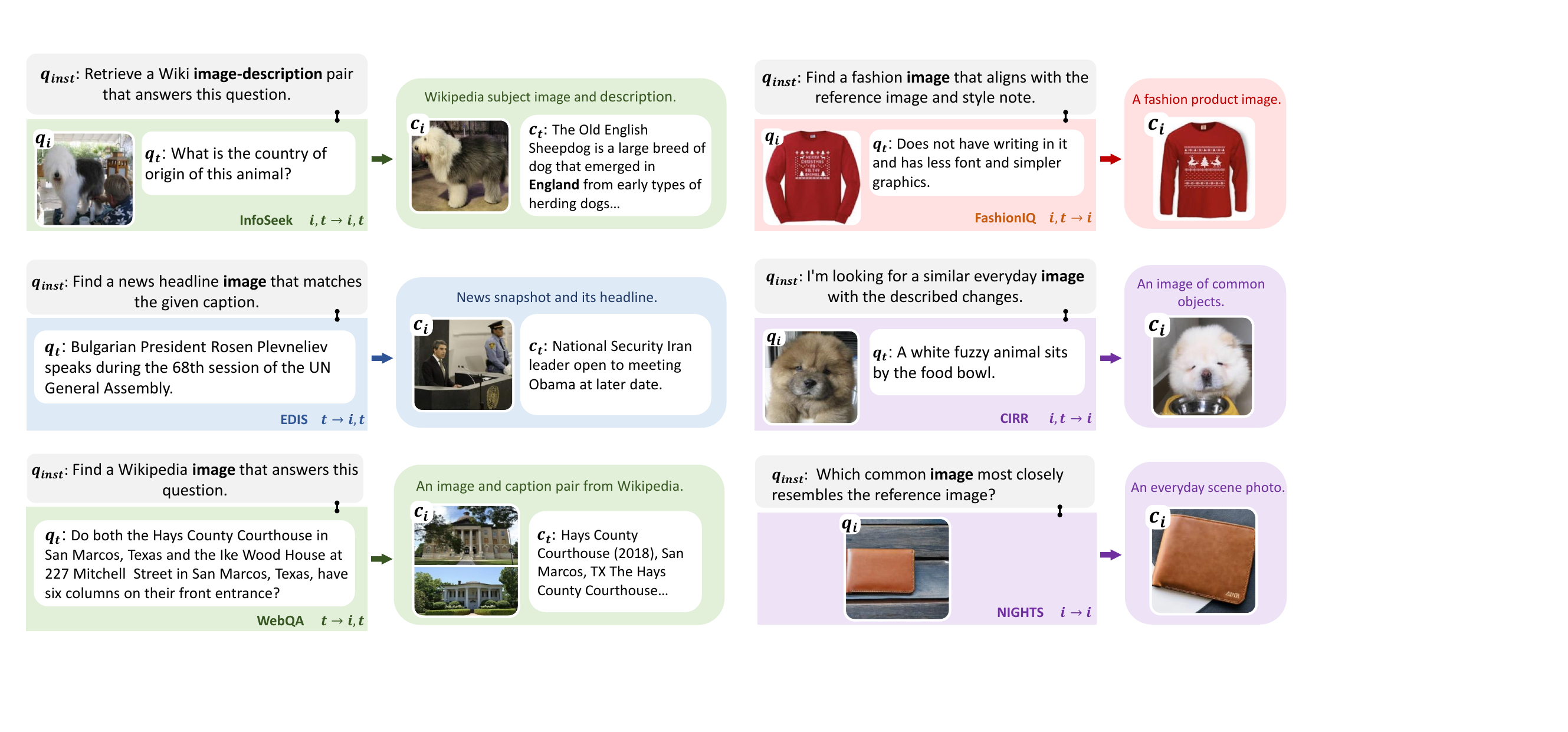}
  \caption{Examples of six query instances in the M-BEIR dataset. Each example query instance includes a query $\textbf{q}$, a human-annotated natural language instruction $q_{\text{inst}}$, and a positive(relevant) candidate $\mathbf{c}^+$.}
  \label{fig:MbeirDataExample}
\end{figure*}

\paragraph{Feature-level Fusion.} Contrasting the approach of processing uni-modal data separately, feature-level fusion integrates features during the encoding phase. This fusion method computes a unified feature vector for multi-modal queries or candidates using mixed-modality attention layers. As illustrated in Figure~\ref{fig:model} (b), for the CLIP feature-level fusion (CLIP$_{\text{FF}}$), we have enhanced the pre-trained vision encoder $f_I$ and text encoder $f_T$ with a 2-layer Multi-Modal Transformer, which follows the same architecture as T5 Transformer, forming a mixed-modality encoder $f_{\text{MIX}}$. In the case of BLIP feature-level fusion (BLIP$_{\text{FF}}$), the process begins with the extraction of image embeddings through the vision encoder $f_I$. These embeddings are then integrated with text embeddings through the cross-attention layers of BLIP's image-grounded text encoder, also labeled as $f_{\text{MIX}}$. In both CLIP$_{\text{FF}}$ and BLIP$_{\text{FF}}$, the output from $f_{\text{MIX}}$ is a comprehensive feature vector that combines information from both image and text modalities. The final representations for the query and target, denoted as $f_{\text{MIX}}(q_{\text{i}}, q_{\text{t}}, q_{\texttt{inst}})$ and $f_{\text{MIX}}(c_{\text{i}}, c_{\text{t}})$ respectively, are obtained separately but using the same $f_{MIX}$. The similarity score between the query and the target is then calculated by:
\begin{align*}
s_{\mathbf{q},\mathbf{c}} &= f_{\text{MIX}}(q_{\text{i}}, q_{\text{t}}, q_{\texttt{inst}})^T \cdot f_{\text{MIX}}(c_{\text{i}}, c_{\text{t}})
\end{align*}
We fine-tuned the above-detailed four types of model variants on the M-BEIR training data (detail in Section ~\ref{sec:dataset}), employing the query-target contrastive objective. To adhere to a uniform instruction tuning format, instructions $q_{\texttt{inst}}$ were integrated as prefixes to the text query $q_{\text{t}}$. See examples in Figure~\ref{fig:MbeirDataExample}.  We input padding tokens for queries or candidates missing either image or text modalities.

\definecolor{positive}{HTML}{008000} 
\definecolor{negative}{HTML}{FF0000}

\begin{table*}
    \centering
    \large
    \resizebox{\textwidth}{!}{%
    \begin{tabular}{ll|c|ccccc|lllll}
    \toprule
    &&\textbf{Zero-shot}&\multicolumn{5}{c|}{\textbf{Multi-task} (\ding{55} instruction)}& \multicolumn{5}{c}{\textbf{UniIR}(\checkmark instruction)}\\
    \cmidrule(lr){3-3}\cmidrule(lr){4-8}\cmidrule(lr){9-13}
    \textbf{Task} & \textbf{Dataset}  
    & BLIP2 & CLIP$_{\text{SF}}$ & CLIP$_{\text{FF}}$ & BLIP$_{\text{SF}}$ & BLIP$_{\text{FF}}$ &
    BLIP$_{\text{FF, 384}}$ 
    & CLIP$_{\text{SF}}$ {\large ($\Delta$)}& CLIP$_{\text{FF}}$ {\large ($\Delta$)} & BLIP$_{\text{SF}}$ {\large ($\Delta$)} & BLIP$_{\text{FF}}$ {\large ($\Delta$)} &
    BLIP$_{\text{FF, 384}}$ 
 {\large ($\Delta$)}\\
    \cmidrule(lr){1-2}    \cmidrule(lr){3-3}\cmidrule(lr){4-8}\cmidrule(lr){9-13}

\multirow{3}{*}{1. $q_t \to c_i$}
         & VisualNews               & 0.0 & 12.7 & 8.8 & 5.0 & 8.3 & 10.5  & \textbf{42.6}~\textcolor{positive}{\large (+29.9)} & \underline{28.8}~\textcolor{positive}{\large (+20.0)}& 20.9~\textcolor{positive}{\large (+15.8)}& 23.0~\textcolor{positive}{\large (+14.8)}& 26.5~\textcolor{positive}{\large (+16.0)}\\
         & MSCOCO                 & 0.0& 27.3 & 24.6 & 22.9 & 27.7 & 35.2  & \textbf{77.9}~\textcolor{positive}{\large (+50.6)} & 74.7~\textcolor{positive}{\large (+50.1)}& 71.6~\textcolor{positive}{\large (+48.7)}& 75.6~\textcolor{positive}{\large (+47.8)}& \underline{75.7}~\textcolor{positive}{\large (+40.4)}\\
         & Fashion200K             & 0.0& 5.9 & 5.9 & 5.7 & 9.0 & 13.1  & 17.8~\textcolor{positive}{\large (+11.9)} & 15.5~\textcolor{positive}{\large (+9.7)}& 24.3~\textcolor{positive}{\large (+18.6)}& \underline{25.4}~\textcolor{positive}{\large (+16.4)}& \textbf{26.7}~\textcolor{positive}{\large (+13.6)}\\
         \midrule
         2. $q_t \to c_t$
         & WebQA     & 35.2& 82.3 & 67.9 & 74.4 & 76.1 & 76.9  & \textbf{84.7}~\textcolor{positive}{\large (+2.5)} & 78.4~\textcolor{positive}{\large (+10.6)}& 78.9~\textcolor{positive}{\large (+4.4)}& \underline{79.5}~\textcolor{positive}{\large (+3.4)}& 79.2~\textcolor{positive}{\large (+2.4)}\\
         \midrule
         \multirow{2}{*}{3. $q_t \to$ ($c_i, c_t$)}
         & EDIS  & 0.0& 41.1 & 38.3 & 33.6 & 36.0 & 38.5  & \textbf{59.4}~\textcolor{positive}{\large (+18.3)} & 50.0~\textcolor{positive}{\large (+11.7)}& 47.2~\textcolor{positive}{\large (+13.6)}& 50.3~\textcolor{positive}{\large (+14.4)}& \underline{51.4}~\textcolor{positive}{\large (+12.9)}\\
         & WebQA & 0.0& 68.2 & 62.5 & 73.2 & 74.7 & 75.2  & 78.8~\textcolor{positive}{\large (+10.6)} & 75.3~\textcolor{positive}{\large (+12.8)}& 76.8~\textcolor{positive}{\large (+3.6)}& \textbf{79.7}~\textcolor{positive}{\large (+5.0)}& \underline{79.4}~\textcolor{positive}{\large (+4.2)}\\
         \midrule
         \multirow{3}{*}{4. $q_i \to c_t$}
         & VisualNews              & 0.0& 12.1 & 8.2 & 4.8 & 4.9 & 6.0  & \textbf{42.8}~\textcolor{positive}{\large (+30.7)} & \underline{28.6}~\textcolor{positive}{\large (+20.4)}& 19.4~\textcolor{positive}{\large (+14.6)}& 21.1~\textcolor{positive}{\large (+16.3)}& 22.9~\textcolor{positive}{\large (+16.9)}\\
         & MSCOCO                & 0.0& 84.6 & 80.8 & 74.9 & 76.9 & 81.4  & \textbf{92.3}~\textcolor{positive}{\large (+7.8)} & 89.0~\textcolor{positive}{\large (+8.2)}& 88.2~\textcolor{positive}{\large (+13.4)}& 88.8~\textcolor{positive}{\large (+11.9)}& \underline{90.1}~\textcolor{positive}{\large (+8.7)}\\
         & Fashion200K          & 0.0& 1.2 & 1.3 & 2.6 & 3.6 & 4.0  & 17.9~\textcolor{positive}{\large (+16.7)} & 13.7~\textcolor{positive}{\large (+12.4)}& 24.3~\textcolor{positive}{\large (+21.7)}& \underline{27.6}~\textcolor{positive}{\large (+24.1)}& \textbf{28.4}~\textcolor{positive}{\large (+24.4)}\\
         \midrule
         5. $q_i \to c_i$
         & NIGHTS     & 24.0& 31.0 & 30.8 & 32.9 & 31.3 & 32.5  & 32.0~\textcolor{positive}{\large (+1.0)} & 31.9~\textcolor{positive}{\large (+1.2)}& \underline{33.4}~\textcolor{positive}{\large (+0.4)}& 33.0~\textcolor{positive}{\large (+1.6)}& \textbf{33.7}~\textcolor{positive}{\large (+1.3)}\\
         \midrule
         \multirow{2}{*}{6. ($q_i, q_t$) $\to c_t$}
         & OVEN                        & 0.0& 36.8 & 31.6 & 33.2 & 37.7 & \underline{39.2}  & \underline{39.2}~\textcolor{positive}{\large (+2.4)} & 34.7~\textcolor{positive}{\large (+3.1)}& 35.2~\textcolor{positive}{\large (+2.0)}& 38.7~\textcolor{positive}{\large (+1.0)}& \textbf{40.7}~\textcolor{positive}{\large (+1.5)}\\
         & InfoSeek                & 0.0& 18.3 & 15.4 & 11.9 & 17.8 & 17.1  & \textbf{24.0}~\textcolor{positive}{\large (+5.8)} & 17.5~\textcolor{positive}{\large (+2.1)}& 16.7~\textcolor{positive}{\large (+4.8)}& \underline{19.7}~\textcolor{positive}{\large (+1.9)}& 19.2~\textcolor{positive}{\large (+2.0)}\\
         \midrule
         \multirow{2}{*}{7. ($q_i, q_t$) $\to c_i$}
         & FashionIQ                & 3.9& 22.8 & 19.7 & 26.1 & 28.1 & \underline{29.0}  & 24.3~\textcolor{positive}{\large (+1.5)} & 20.5~\textcolor{positive}{\large (+0.9)}& 26.2~\textcolor{positive}{\large (+0.1)}& 28.5~\textcolor{positive}{\large (+0.5)}& \textbf{29.8}~\textcolor{positive}{\large (+0.9)}\\
         & CIRR                     & 6.2& 32.0 & 32.7 & 36.7 & 45.1 & 47.4  & 43.9~\textcolor{positive}{\large (+11.9)} & 40.9~\textcolor{positive}{\large (+8.2)}& 43.0~\textcolor{positive}{\large (+6.3)}& \textbf{51.4}~\textcolor{positive}{\large (+6.3)}& \underline{51.1}~\textcolor{positive}{\large (+3.8)}\\
         \midrule
         \multirow{2}{*}{8. ($q_i, q_t$) $\to$ ($c_i, c_t$)}
         & OVEN                       & 13.8& 58.7 & 50.1 & 51.0 & 51.6 & 53.1  & \textbf{60.2}~\textcolor{positive}{\large (+1.5)} & 55.8~\textcolor{positive}{\large (+5.7)}& 51.8~\textcolor{positive}{\large (+0.8)}& 57.8~\textcolor{positive}{\large (+6.2)}& \underline{59.5}~\textcolor{positive}{\large (+6.4)}\\
          & InfoSeek               & 11.4& \underline{42.3} & 31.5 & 23.0 & 25.4 & 25.2  & \textbf{44.6}~\textcolor{positive}{\large (+2.4)} & 36.8~\textcolor{positive}{\large (+5.3)}& 25.4~\textcolor{positive}{\large (+2.5)}& 27.7~\textcolor{positive}{\large (+2.3)}& 31.1~\textcolor{positive}{\large (+5.9)}\\
         \midrule
         & Average     & 5.9& 36.1 & 31.9 & 32.0 & 34.6 & 36.5  & \textbf{48.9}~\textcolor{positive}{\large (+12.8)} & 43.3~\textcolor{positive}{\large (+11.4)}& 42.7~\textcolor{positive}{\large (+10.7)}& 45.5~\textcolor{positive}{\large (+10.9)}& \underline{46.6}~\textcolor{positive}{\large (+10.1)}\\

         \bottomrule
    \end{tabular}
    }
    \caption{Benchmarking universal information retrieval on M-BEIR (5.6M candidates) with Recall@5 (except using Recall@10 for Fashion200K, FashionIQ). 
    $\Delta$: UniIR - Multi-task. 384 refers to image resolution. \textbf{Bold}: top-1 performance. \underline{Underline}: top-2.}
    \label{tab:union_small}
    \vspace{-2pt}
\end{table*}
\section{M-BEIR Benchmark}
\label{sec:dataset}

To train and evaluate unified multimodal retrieval models, we build a large-scale retrieval benchmark named M-BEIR (\textbf{M}ultimodal \textbf{BE}nchmark for \textbf{I}nstructed \textbf{R}etrieval).
The M-BEIR benchmark comprises eight multimodal retrieval tasks and ten datasets from a variety of domains and image sources. Each task is accompanied by human-authored instructions, encompassing 1.5 million queries and a pool of 5.6 million retrieval candidates in total (see Table~\ref{tab:dataset}).

\subsection{Data Format}
To unify multimodal retrieval tasks, which consist of different modalities in the source query and target candidate, each task in M-BEIR includes queries $\mathcal{Q} = \{\mathbf{q_1}, \mathbf{q_2}, ...\}$, a set of candidates $\mathcal{C} = \{\mathbf{c_1}, \mathbf{c_2}, ...\}$, where $\mathbf{q}$ and $\mathbf{c}$ both support text and image modality, and a human-authored instruction $q_{\texttt{inst}}$ is provided to specify the intent of the retrieval task. Each query instance in the M-BEIR dataset includes a query  $\mathbf{q}$, an instruction $q_{\texttt{inst}}$, a list of relevant(positive) candidate data $\mathbf{c}^+$ and a list of potentially available irrelevant(negative) candidate data $\mathbf{c}^-$. See examples in Figure~\ref{fig:MbeirDataExample}. Every M-BEIR query instance has at least one positive candidate data and possibly no negative candidate data. Our default retrieval setting is that the model needs to retrieve the positive candidates from a heterogeneous pool of candidates in all different modalities and domains.

\subsection{Dataset Collection}
The M-BEIR benchmark encompasses various domains: everyday imagery, fashion items, Wikipedia entries, and news articles. It integrates 8 multimodal retrieval tasks by leveraging a variety of datasets.

\paragraph{Data Selection.} 
To build a unified instruction-tuned multimodal retrieval model and comprehensive evaluation benchmark, we aim to cover a wide range of multimodal tasks, domains, and datasets. These include retrieval-focused datasets (OVEN~\cite{hu2023open}, EDIS~\cite{liu2023edis}, CIRR~\cite{liu2021image} and FashionIQ~\cite{wu2021fashion}), image-caption datasets (MS-COCO~\cite{lin2014microsoft}, Fashion200K~\cite{han2017automatic}, VisualNews~\cite{liu2020visual}), image-similarity measurement dataset (NIGHTS~\cite{fu2023learning}), along with retrieval-based VQA datasets (InfoSeek~\cite{chen2023infoseek}, WebQA~\cite{chang2022webqa}). These datasets, originally designed for different purposes, are effectively repurposed as retrieval tasks within the M-BEIR benchmark. In the case of image-caption datasets, we repurpose the image-caption pair as the retrieval task following MS-COCO. For the other datasets, we adopt original queries and use the annotated gold candidates as positive candidates $\mathbf{c}^+$ and annotated hard negatives as irrelevant candidates $\mathbf{c}^-$. We also adopt the provided candidate pool. In total, M-BEIR covers 8 different multimodal retrieval tasks and 4 domains with a global pool of 5.6 million candidates. See Table \ref{sec:dataset} for the full dataset list. To ensure data balance in our benchmark, we trim down candidate pools and instances from the larger datasets such as VisualNews, OVEN, and InfoSeek, which originally contained 1 to 6 million instances, significantly larger than other datasets. To facilitate training, validation, and testing, we use the original dataset splits from each dataset. If the dataset only releases a validation set, we hold out a part of the training data to use for validation and report results on the original validation set. Otherwise, we report results using the test set. More details on data processing can be found in the Appendix.

\paragraph{Instruction Annotation Guideline.}
One of the key components of the success of instruction-tuning is the diverse instructions that specify the intention of the task~\cite{wei2021finetuned,chung2022scaling}. To design instructions for multimodal retrieval tasks, we took inspiration from the instruction schema in TART~\cite{asai2022task}. Our M-BEIR instruction describes a multimodal retrieval task by intent, domain, query modality, and target candidate modality.
Specifically, intent describes how the retrieved resources are related to the query. The domain defines the expected resource of the target candidate, such as Wikipedia or fashion products. For a text-to-image retrieval dataset like Fashion200K~\cite{han2017automatic}, our instruction would be: ``Based on the following fashion description, retrieve the best matching image.'' More examples in Table~\ref{tab:dataset} and Figure~\ref{fig:MbeirDataExample}. Following the instruction annotation guideline, we authored 4 instructions for each query in every retrieval task. The full list of instructions is in the Appendix Table~\ref{tab:instructions} and Table~\ref{tab:instructions2}.

\subsection{Evaluation Metrics}
We follow the standard retrieval evaluation metric, recall@k, used for MSCOCO and report results for all datasets. Specifically, we adhere to the recall implementation of CLIP~\cite{radford2021learning}/BLIP~\cite{li2022blip} for MSCOCO, which counts the retrieved instance as correct if it overlaps with relevant instances. We mainly report Recall@5 for all datasets except Fashion200K and FashionIQ, following the prior work~\cite{wu2021fashion} to report Recall@10. Full results of Recall@1/5/10 can be found in the Appendix Table~\ref{tab:union}.

\definecolor{lightblue}{RGB}{204,229,255} 
\definecolor{green}{RGB}{0,180,0} 

\begin{figure*}[t!]
\centering
\scriptsize
\begin{tabular}{>{\columncolor{lightblue}}m{1.5cm} >{\columncolor{lightblue}}m{1cm} m{2.2cm} m{3.8cm} m{2.4cm} m{3.8cm}}
    \toprule
    \textbf{Dataset} & \textbf{Domain} & \textbf{Task} & \textbf{Query Instruction  $q_{\texttt{inst}}$} & \textbf{Query Image $q_i$} & \textbf{Query Text $q_t$} \\
    \midrule
    EDIS 
    & News
    & 3. $q_t$ $\to$ ($c_i, c_t$)
    & Find a news headline image that matches the provided caption. 
    & -
    & Barack Obama with Germany's chancellor Angela Merkel at the Brandenburg Gate Berlin on 19 June.
\end{tabular}

    \begin{tabular}{>{\columncolor{lightblue}}m{1.2cm}m{2.7cm}m{2.7cm}m{2.7cm}m{2.7cm}m{2.7cm}}
            \toprule
            \textbf{Model} & \textbf{Rank 1 ($c_i, c_t$)} & \textbf{Rank 2 ($c_i, c_t$)} & \textbf{Rank 3 ($c_i, c_t$)} & \textbf{Rank 4 ($c_i, c_t$)} & \textbf{Rank 5 ($c_i, c_t$)} \\
            \midrule
                  UniIR
                  \newline (CLIP$_{\text{SF}}$)
                  \newline \checkmark inst
                  &  
                  \Large \textcolor{red}{\texttimes}
                  \includegraphics[width=18mm,height=18mm]{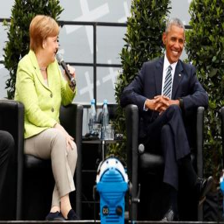}
                  \newline \scriptsize Obama Speaks at a Berlin Event With Angela Merkel.
                  &  
                  \large \textcolor{green}{\checkmark}
                  \includegraphics[width=18mm,height=18mm]{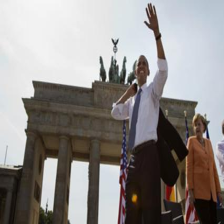} 
                  \newline \scriptsize Obama's Berlin visit to coincide with Trump in Brussels - Barack Obama.
                  &  
                  \Large \textcolor{red}{\texttimes}
                  \includegraphics[width=18mm,height=18mm]{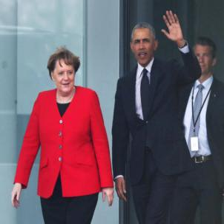} 
                  \newline \scriptsize Obama meets Germany's Merkel at chancellery in Berlin.
                  & 
                  \large \textcolor{green}{\checkmark}
                  \includegraphics[width=18mm,height=18mm]{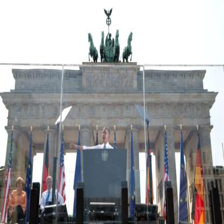} 
                  \newline \scriptsize President Obama Speaks to the People of Berlin from the Brandenburg Gate.
                  & 
                  \large \textcolor{green}{\checkmark}
                  \includegraphics[width=18mm,height=18mm]{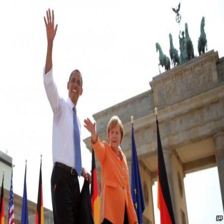} 
                  \newline \scriptsize When Barack Obama visited Berlin two years ago, he charmed a city.
                  \\
            \midrule
                  Multi-task
                  \newline (CLIP$_{\text{SF}}$)
                  \newline \ding{55} inst
                  &  
                  \Large \textcolor{red}{\texttimes}
                  \scriptsize President Obama stands next to German Chancellor Angela Merkel in front of Brandenburg Gate in Berlin on June 19.
                  &  
                  \Large \textcolor{red}{\texttimes}
                  \scriptsize President Obama and German Chancellor Angela Merkel in 2011.
                  &  
                  \large \textcolor{green}{\checkmark}
                  \includegraphics[width=18mm,height=18mm]{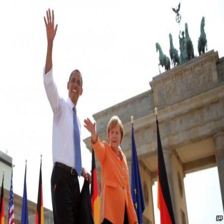}
                  \scriptsize When Barack Obama visited Berlin two years ago, he charmed a city.
                  & 
                  \Large \textcolor{red}{\texttimes}
                  \scriptsize  Barack Obama with German Chancellor Angela Merkel at the G20 summit in November.
                  & 
                  \Large \textcolor{red}{\texttimes}
                  \scriptsize US president Barack Obama at the Brandenburg Gate.
                  \\
            \midrule
                  Zero-shot
                  \newline (BLIP2)
                  \newline \ding{55} inst
                  &  
                  \Large \textcolor{red}{\texttimes}
                  \tiny President Obama stands next to German Chancellor Angela Merkel in front of Brandenburg Gate in Berlin on June 19.
                  &  
                  \Large \textcolor{red}{\texttimes}
                  \tiny US President Barack Obama waves next to German Chancellor Angela Merkel before they deliver speeches to invited guests in front of the Brandenburg Gate at Pariser Platz in Berlin on June 19 2013 during the official visit of the US President Barack Obama walks in ...
                  &  
                  \Large \textcolor{red}{\texttimes}
                  \tiny Barack Obama with German chancellor Angela Merkel at the G20 summit in November.
                  & 
                  \Large \textcolor{red}{\texttimes}
                  \tiny BERLIN GERMANY JUNE 19 US President Barack Obama meets German Chancellor Angela Merkel for bilateral talks at the Chancellery on June 19 2013 in Berlin Germany Obama is visiting Berlin for the first time during his presidency and his speech at the Brandenburg Gate...
                  & 
                  \Large \textcolor{red}{\texttimes}
                  \tiny President Obama walks with Germany s Chancellor Angela Merkel in St Petersburg Russia on Sept 6.
                  \\
            \bottomrule
    \end{tabular}
    \caption{Visualization of top 5 retrieved candidates from M-BEIR with 3 models on EDIS. Without instructions, zero-shot and multi-task training models mostly retrieve the wrong modality (text-only). UniIR retrieves candidates accurately with the right modality (image, text).}
    \label{fig:retrieval_results_edis_main}

\end{figure*}

\section{Experiments}
\label{sec:experiment}
In our experiments, we assess a variety of multimodal retrieval models on the M-BEIR dataset, leveraging pre-trained vision-language transformer models. We use publicly available checkpoints, as listed in Table~\ref{tab:union_small}. 
Our evaluation encompasses both SoTA models, fine-tuned baselines and UniIR models(detailed in Section \ref{sec:task}) under two retrieval scenarios: (1) retrieving from the M-BEIR 5.6 million candidate pool, which consists of the retrieval corpus from all tasks, and (2) retrieving from a task-specific pool (with homogeneous candidates) provided by the original dataset, which is to enable fair comparison with existing SoTA retrievers. We name this task-specific pool as M-BEIR$_{\text{local}}$.
The retrieval process involves a two-step pipeline. Firstly, we extract multimodal feature vectors for all the queries and candidates in the pool. We then utilize FAISS~\citep{johnson2019billion}, a powerful library for efficient similarity searches in dense vector spaces, to index and retrieve candidates.

\subsection{Baselines}
\paragraph{Zero-shot SoTA Retriever.}
We utilize pre-trained vision-language models such as CLIP (L-14)\citep{radford2021learning}, SigLIP (L)\citep{zhai2023sigmoid}, BLIP (L)~\citep{li2022blip}, and BLIP2~\citep{li2023blip} as our baseline feature extractors. 
The caveat is that these models cannot understand the intent of the retrieval task as the input is only query $\textbf{q}$, thus, they are expected to achieve low performance in the standard setting (1) with a heterogeneous candidate pool. We do not use $q$ instructions as we found this even degrades zero-shot retrieval performance.

\paragraph{Single/Multi-task Fine-tuned Baselines.}
We fine-tune CLIP and BLIP on this specific dataset as our single-task baseline retrievers. We also fine-tune CLIP and BLIP jointly on all M-BEIR training data without incorporating instructions as our multi-task baseline retrievers. The model only takes in $\textbf{q}$ and $\textbf{c}$ using the query-target contrastive training objective to maximize the positive pair similarity while minimizing negative pair similarity. 

\paragraph{Implementation Details.} For all the CLIP and BLIP variants, we employ the largest checkpoint, i.e., ViT-L14~\cite{dosovitskiy2020image}.
The default image resolution is $224 \times 224$ unless specified otherwise. We use a batch size of 105 for CLIP variants and 115 for BLIP variants. We adopt other hyperparameters as reported in the original implementations. For score fusion methods, we set $w_1=w_2=w_3=w_4=1$ by default. All our experiments are conducted on a single node with 8 H100 GPUs. Further details can be found in the Appendix.

\subsection{Experimental Results}
We report the main results on M-BEIR in Table~\ref{tab:union_small}, where models retrieve candidates from the 5.6 M pool. We show that zero-shot models struggle to retrieve queried information from such a heterogeneous pool. We demonstrate instruction-tuning as a crucial component in Table~\ref{tab:union_instruction} and show multimodal fusion architecture design insights in Table~\ref{tab:union_architecture}. Furthermore, we also conduct experiments to understand the zero-shot generalization of UniIR, where we would train UniIR on a subset of datasets and evaluate it on the held-out test set, which makes UniIR fairly comparable with other zero-shot retrievers.

\paragraph{Zero-shot retrievers cannot comprehend retrieval intention.}
We first benchmark four open-sourced cross-modal embedding models and found the recall values on most tasks are near zero.  To demonstrate this, we have provided an example of BLIP2 in Table~\ref{tab:union_small}. These pre-trained models struggle to comprehend the task intention without the guidance of instruction. For example, in the text-to-image retrieval task on MSCOCO, all zero-shot models retrieve text instances from the global pool, leading to 0\% recall rate. This outcome is expected, given that similarity scores tend to be higher when the query and candidate come from the same modality. 
Furthermore, we observed that zero-shot models, for example, BLIP2 in Table~\ref{tab:union_small} cannot effectively fuse modalities as the recall of WebQA drops from 35.2\% to 0\% when the retrieval candidates are image-text pairs instead of text snippets. In Figure~\ref{fig:retrieval_results_edis_main}, we present examples where BLIP2 retrieves distracting candidates from the wrong modality for an EDIS query.

\definecolor{positive}{HTML}{008000} 
\definecolor{negative}{HTML}{FF0000}

\begin{table}[ht!]
    \centering
    \large
    \resizebox{0.49\textwidth}{!}{%
    \begin{tabular}{lcclccl}
    \toprule
     & \textbf{ZS} & \textbf{Multi.} & \textbf{UniIR} & \textbf{ZS} & \textbf{Multi.} & \textbf{UniIR}\\
     \cmidrule(lr){1-1}
     \cmidrule(lr){2-4}
     \cmidrule(lr){5-7}
     
 \textbf{Task} & CLIP & CLIP$_{\text{SF}}$ & CLIP$_{\text{SF}}$ {\small ($\Delta$)} & BLIP & BLIP$_{\text{FF}}$ & BLIP$_{\text{FF}}$ {\small ($\Delta$)} \\
        \midrule

1. & 0.0 & 15.3 & 46.1~\textcolor{positive}{\small (+30.8)} & 0.0 & 15.0 & 41.3~\textcolor{positive}{\small (+26.3)}\\
2. & 32.1 & 82.3 & 84.7~\textcolor{positive}{\small (+2.5)} & 38.1 & 76.1 & 79.5~\textcolor{positive}{\small (+3.4)}\\
3. & 6.1 & 54.6 & 69.1~\textcolor{positive}{\small (+14.5)} & 0.0 & 62.0 & 65.0~\textcolor{positive}{\small (+3.0)}\\
4. & 0.0 & 32.6 & 51.0~\textcolor{positive}{\small (+18.4)} & 0.0 & 28.4 & 45.9~\textcolor{positive}{\small (+17.4)}\\
5. & 25.3 & 31.0 & 32.0~\textcolor{positive}{\small (+1.0)} & 25.1 & 31.3 & 33.0~\textcolor{positive}{\small (+1.6)}\\
6. & 0.0 & 27.5 & 31.6~\textcolor{positive}{\small (+4.1)} & 0.0 & 27.8 & 29.2~\textcolor{positive}{\small (+1.5)}\\
7. & 4.9 & 27.4 & 34.1~\textcolor{positive}{\small (+6.7)} & 4.8 & 36.6 & 40.0~\textcolor{positive}{\small (+3.4)}\\
8. & 23.3 & 50.5 & 52.4~\textcolor{positive}{\small (+1.9)} & 9.0 & 38.5 & 42.7~\textcolor{positive}{\small (+4.2)}\\
\midrule
Avg. & 7.9 & 36.1 & 48.9~\textcolor{positive}{\small (+12.8)} & 5.7 & 34.6 & 45.5~\textcolor{positive}{\small (+10.9)}\\

\bottomrule
    \end{tabular}
    }
    \caption{\textbf{Experiments of instruction-tuning}. Retrieve from the M-BEIR (Recall@5). 
    $\Delta$: UniIR - Multi-task (Multi.).}
    \label{tab:union_instruction}
    \vspace{-2pt}
\end{table}
\paragraph{Instruction-tuning improves retrieval on M-BEIR.}
To understand the benefit of instruction-tuning in UniIR, we present a comparison of UniIR with multi-task fine-tuned baselines in Table~\ref{tab:union_instruction}. Despite having the same architecture, UniIR models show significant improvement over baselines on M-BEIR. The average Recall@5 has increased by 12.8 and 10.9, respectively. 
We also discovered that the largest improvement was observed in cross-modality retrieval tasks 1 and 3. Without instructions, the multi-task baselines struggle to understand the task intention and tend to retrieve candidates from the same modality as the query. However, Instruction-tuning does not significantly improve within-modality retrieval tasks like 2 and 5 as these do not require the embedding model to understand intent.
\begin{table}
    \centering
    \small
    \resizebox{0.8\columnwidth}{!}{
    \begin{tabular}{l|cc|cc}
    \toprule
         & \multicolumn{2}{c|}{\textbf{Multi-task}} &  \multicolumn{2}{c}{\textbf{UniIR}} \\
         \cmidrule(lr){2-3}\cmidrule(lr){4-5}
         \textbf{Error Types} & CLIP$_{\text{SF}}$ & BLIP$_{\text{FF}}$ &  CLIP$_{\text{SF}}$ & BLIP$_{\text{FF}}$ \\
         \midrule
         \ding{55} modality&  58.8\%& 50.9\% &  2.7\% & 15.2\%\\
         \ding{55} domain &  0.3\%&  0.5\%&  0.1\%&  0.0\%\\
         Other &  40.9\%&  48.6\%&  97.2\%&  84.8\%\\
         \bottomrule
    \end{tabular}
    }
    \caption{Error analysis on M-BEIR.}
    \label{tab:union_error}
    \vspace{-15pt}
\end{table}

\paragraph{UniIR can precisely follow instructions.}
To further demonstrate the advantages of UniIR over Multi-task finetuning baselines, we conducted an analysis of the retrieval error. The errors were classified into three categories: incorrect modality, incorrect domain, and other errors. The results are presented in Table~\ref{tab:union_error}. The Multi-task models showed a high error rate of 58.8\% and 50.9\% in retrieving instances with the wrong modality from the global pool. However, with instruction finetuning, UniIR models were able to successfully learn to retrieve intended modalities, resulting in a significant drop in error rate to 2.7\% and 15.2\%.
In Figure~\ref{fig:retrieval_results_edis_main}, we show examples of incorrect modality errors by visualizing the top 5 retrieved candidates using zero-shot, multi-task and UniIR models on one of EDIS queries. Specifically, the zero-shot model (BLIP2) and multi-task model (CLIP$_{\text{SF}}$) mostly retrieve distracting candidates from the wrong modality ($c_t$), while UniIR (CLIP$_{\text{SF}}$) retrieves all positive candidates from the right modality ($c_i, c_t$). 
More examples can be found in Appendix Figure~\ref{fig:retrieval_results_visualnews}-\ref{fig:retrieval_results_infoseek}.

\begin{table}[ht!]
    \centering
    \large
    \resizebox{0.49\textwidth}{!}{%
    \begin{tabular}{lcc|cl|cl}
    \toprule
      \textbf{}& \multicolumn{2}{c}{\textbf{Zero-shot}} & \multicolumn{4}{c}{\textbf{UniIR}} \\
      
      \cmidrule(lr){2-3} \cmidrule(lr){4-7}
      \textbf{Task} & CLIP & BLIP &CLIP$_{\text{FF}}$ & CLIP$_{\text{SF}}$ {\scriptsize ($\Delta$)} & BLIP$_{\text{SF}}$ & BLIP$_{\text{FF}}$ {\scriptsize ($\Delta$)}\\
        \midrule

1. & 0.0 & 0.0 & 39.7 & 46.1~\textcolor{positive}{\small (+6.4)} & 38.9 & 41.3~\textcolor{positive}{\small (+2.4)}\\
2. & 32.1 & 38.1 & 78.4 & 84.7~\textcolor{positive}{\small (+6.3)} & 78.9 & 79.5~\textcolor{positive}{\small (+0.7)}\\
3. & 6.1 & 0.0 & 62.7 & 69.1~\textcolor{positive}{\small (+6.4)} & 62.0 & 65.0~\textcolor{positive}{\small (+3.0)}\\
4. & 0.0 & 0.0 & 43.8 & 51.0~\textcolor{positive}{\small (+7.3)} & 44.0 & 45.9~\textcolor{positive}{\small (+1.9)}\\
5. & 25.3 & 25.1 & 31.9 & 32.0~\textcolor{positive}{\small (+0.1)} & 33.4 & 33.0~\textcolor{negative}{\small (--0.4)}\\
6. & 0.0 & 0.0 & 26.1 & 31.6~\textcolor{positive}{\small (+5.5)} & 26.0 & 29.2~\textcolor{positive}{\small (+3.2)}\\
7. & 4.9 & 4.8 & 30.7 & 34.1~\textcolor{positive}{\small (+3.4)} & 34.6 & 40.0~\textcolor{positive}{\small (+5.4)}\\
8. & 23.3 & 9.0 & 46.3 & 52.4~\textcolor{positive}{\small (+6.1)} & 38.6 & 42.7~\textcolor{positive}{\small (+4.1)}\\
\midrule
Avg & 7.9 & 5.7 & 43.3 & 48.9~\textcolor{positive}{\small (+5.7)} & 42.7 & 45.5~\textcolor{positive}{\small (+2.8)}\\

\bottomrule
    \end{tabular}
    }
    \caption{\textbf{Experiments of multimodal feature fusion architecture design}. Retrieve from the M-BEIR (Recall@5). 
    $\Delta$: difference between two model architectures.}
    \label{tab:union_architecture}
    \vspace{-2pt}
\end{table}

\definecolor{positive}{HTML}{008000} 
\definecolor{negative}{HTML}{FF0000}

\begin{table*}[ht!]
    \small
    \centering
    \small
    \vspace{-5mm}
    \resizebox{\textwidth}{!}{%
    \begin{tabular}{llcccc|ccl|ccl}
    \toprule
     &  &  \multicolumn{4}{c}{\textbf{SoTA Zero-Shot}} & \textbf{ST} & \textbf{MT} & \textbf{UniIR} & \textbf{ST} & \textbf{MT} & \textbf{UniIR}\\
    \cmidrule{1-2}\cmidrule(lr){3-6} \cmidrule(lr){7-7} \cmidrule(lr){8-8}\cmidrule(lr){9-9}\cmidrule(lr){10-10}\cmidrule(lr){11-11}\cmidrule(lr){12-12}
        \textbf{Task}        & \textbf{Dataset}  &CLIP & SigLIP & BLIP & BLIP2 &CLIP$_{\text{SF}}$ & CLIP$_{\text{SF}}$ & CLIP$_{\text{SF}}$  {\scriptsize ($\Delta_s$)}& BLIP$_{\text{FF}}$ & BLIP$_{\text{FF}}$ & BLIP$_{\text{FF}}$ {\scriptsize ($\Delta_s$)}\\
        \midrule
         
         \multirow{3}{*}{1. $q_t \to c_i$}
         & VisualNews               & 43.3 & 30.1 & 16.4 & 16.7 & 43.5 & 40.6  & 42.6~\textcolor{negative}{\footnotesize (--0.9)} & 20.0 & 22.8 & 23.4~\textcolor{positive}{\footnotesize (+3.4)}\\
         & MSCOCO                   & 61.1 & 75.7 & 74.4 & 63.8 & 80.4 & 79.9  & 81.1~\textcolor{positive}{\footnotesize (+0.7)} & 77.3 & 78.3 & 79.7~\textcolor{positive}{\footnotesize (+2.3)}\\
         & Fashion200K              & 6.6 & 36.5 & 15.9 & 14.0 & 10.7 & 16.8  & 18.0~\textcolor{positive}{\footnotesize (+7.4)} & 17.1 & 25.8 & 26.1~\textcolor{positive}{\footnotesize (+9.0)}\\
         \midrule
         2. $q_t \to c_t$
         & WebQA         & 36.2 & 39.8 & 44.9 & 38.6 & 81.7 & 83.7  & 84.7~\textcolor{positive}{\footnotesize (+3.1)} & 67.5 & 77.9 & 80.0~\textcolor{positive}{\footnotesize (+12.5)}\\
         \midrule
         \multirow{2}{*}{3. $q_t \to$ ($c_i, c_t$)}
         & EDIS                     & 43.3 & 27.0 & 26.8 & 26.9 & 58.8 & 57.4  & 59.4~\textcolor{positive}{\footnotesize (+0.6)} & 38.2 & 51.2 & 50.9~\textcolor{positive}{\footnotesize (+12.7)}\\
         & WebQA                    & 45.1 & 43.5 & 20.3 & 24.5 & 76.3 & 76.7  & 78.7~\textcolor{positive}{\footnotesize (+2.5)} & 67.8 & 79.2 & 79.8~\textcolor{positive}{\footnotesize (+11.9)}\\
         \midrule
         \multirow{3}{*}{4. $q_i \to c_t$}
         & VisualNews               & 41.3 & 30.8 & 17.2 & 15.0 & 42.7 & 40.0  & 43.1~\textcolor{positive}{\footnotesize (+0.4)} & 22.4 & 20.9 & 22.8~\textcolor{positive}{\footnotesize (+0.3)}\\
         & MSCOCO                   & 79.0 & 88.2 & 83.2 & 80.0 & 89.8 & 90.3  & 92.3~\textcolor{positive}{\footnotesize (+2.6)} & 86.0 & 85.8 & 89.9~\textcolor{positive}{\footnotesize (+3.9)}\\
         & Fashion200K              & 7.7 & 34.2 & 19.9 & 14.2 & 12.0 & 18.4  & 18.3~\textcolor{positive}{\footnotesize (+6.3)} & 15.6 & 27.4 & 28.9~\textcolor{positive}{\footnotesize (+13.3)}\\
         \midrule
         \multirow{1}{*}{5. $q_i \to c_t$}
         & NIGHTS                   & 26.1 & 28.9 & 27.4 & 25.4 & 33.5 & 31.1  & 32.0~\textcolor{negative}{\footnotesize (--1.5)} & 30.4 & 31.5 & 33.0~\textcolor{positive}{\footnotesize (+2.6)}\\
         \midrule
         \multirow{2}{*}{6. ($q_i, q_t$) $\to c_t$}
         & OVEN                     & 24.2 & 29.7 & 16.1 & 12.2 & 45.4 & 46.6  & 45.5~\textcolor{positive}{\footnotesize (+0.1)} & 33.8 & 42.8 & 41.0~\textcolor{positive}{\footnotesize (+7.2)}\\
         & InfoSeek                 & 20.5 & 25.1 & 10.2 & 5.5 & 23.5 & 28.3  & 27.9~\textcolor{positive}{\footnotesize (+4.4)} & 18.5 & 23.9 & 22.4~\textcolor{positive}{\footnotesize (+3.9)}\\
         \midrule
         \multirow{2}{*}{7. ($q_i, q_t$) $\to c_i$}
         & FashionIQ                & 7.0 & 14.4 & 2.3 & 4.4 & 25.9 & 23.2  & 24.4~\textcolor{negative}{\footnotesize (--1.5)} & 3.0 & 28.4 & 29.2~\textcolor{positive}{\footnotesize (+26.2)}\\
         & CIRR                     & 13.2 & 22.7 & 10.6 & 11.8 & 52.0 & 38.7  & 44.6~\textcolor{negative}{\footnotesize (--7.3)} & 13.9 & 48.6 & 52.2~\textcolor{positive}{\footnotesize (+38.2)}\\
         \midrule
         \multirow{2}{*}{8. ($q_i, q_t$) $\to$ ($c_i, c_t$)}
         & OVEN                     & 38.8 & 41.7 & 27.4 & 27.3 & 66.2 & 69.0  & 67.6~\textcolor{positive}{\footnotesize (+1.4)} & 49.9 & 56.3 & 55.8~\textcolor{positive}{\footnotesize (+5.9)}\\
         &  InfoSeek                & 26.4 & 27.4 & 16.6 & 15.8 & 47.4 & 49.2  & 48.9~\textcolor{positive}{\footnotesize (+1.5)} & 32.3 & 32.9 & 33.0~\textcolor{positive}{\footnotesize (+0.7)}\\
         \midrule
         - &  Average               & 32.5 & 37.2 & 26.8 & 24.8 & 49.4 & 49.4  & 50.6~\textcolor{positive}{\footnotesize (+1.2)} & 37.1 & 45.8 & 46.8~\textcolor{positive}{\footnotesize (+9.7)}\\
         \bottomrule
    \end{tabular}
    }
    \caption{Multi-task (MT) instruction-tuning experiments on M-BEIR$_{\text{local}}$. We report Recall@5 results of zero-shot retrieval, single-task (ST) fine-tuning, and UniIR (with or without instructions) on M-BEIR$_{\text{local}}$ except for Fashion200K and FashionIQ where we report Recall@10. 
    Retrieval is conducted from M-BEIR$_{\text{local}}$  (single) candidate pools. $\Delta_{s}$: absolute difference to single task fine-tuning.
    }
    \label{tab:multitask}
    \vspace{-10pt}
\end{table*}

\begin{figure}
    \centering
    \includegraphics[width=0.9\linewidth]{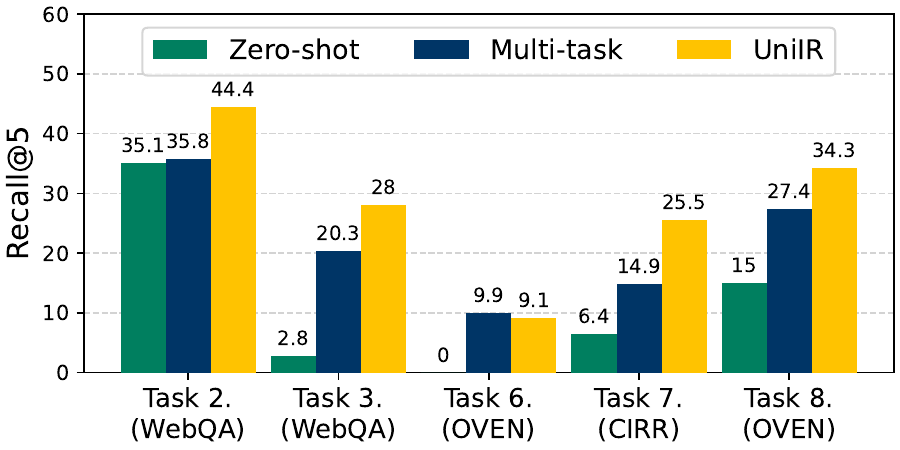}
    \caption{\textbf{Held-out dataset generalization experiments} on M-BEIR: we train a Multi-task and a UniIR model on 7 held-in datasets and test on 3 held-out datasets (WebQA, OVEN, CIRR) from the M-BEIR. Results are averaged over CLIP$_{\text{SF}}$ and BLIP$_{\text{FF}}$.}
    \label{fig:union_held_out}
    \vspace{-15pt}
\end{figure}

\paragraph{UniIR can generalize to unseen retrieval tasks.} 
During the multi-task fine-tuning stage of UniIR, we excluded three datasets (WebQA, OVEN, CIRR) and fine-tuned UniIR models and multi-task baselines on the remaining M-BEIR datasets. At test time on the M-BEIR global pool, we evaluated the zero-shot performance of all fine-tuned models, as well as SoTA pre-trained retrievers (CLIP and BLIP) on the three held-out datasets. In Figure~\ref{fig:union_held_out}, we compared the average performance of SoTA (CLIP and BLIP) retrievers, the average performance of multi-task fine-tuned baselines Multi-task(CLIP$_{\text{SF}}$) and Multi-task(BLIP$_{\text{FF}}$), and the average performance of UniIR (CLIP$_{\text{SF}}$) and UniIR (BLIP$_{\text{FF}}$). Our results indicate two main findings. Firstly, UniIR models outperform SoTA retriever baselines by a significant margin on held-out datasets during zero-shot evaluation. Secondly, we demonstrate that UniIR models, which incorporate instruction-tuning, exhibit superior generalization abilities on unseen tasks and datasets compared to their multi-task counterparts without instructions.

\paragraph{Aligning the model architecture with pre-training.}
In Table~\ref{tab:union_architecture}, we compare fusion architecture designs between score-fusion and feature-fusion, where score-fusion is native to the CLIP model (i.e., CLIP$_{\text{SF}}$) and feature-fusion is native to BLIP with its pre-trained cross-attention transformer encoder (i.e., BLIP$_{\text{FF}}$).
By adhering to the pre-training architecture design, we show that the fine-tuned UniIR models attain higher Recall@5 scores for each task, with an average improvement of 5.7 and 2.8 for CLIP and BLIP, respectively.
Furthermore, avoid adding randomly initialized layers during fine-tuning, such as the T5 mix-modality layers used in CLIP$_{\text{FF}}$. These layers are not pre-trained and can lead to overfitting - specifically for tasks like OVEN/InfoSeek (task 8). The comparison of CLIP$_{\text{FF}}$ and CLIP$_{\text{SF}}$ on task 8 reveals a significant drop in performance, from 52.4 to 46.3, for the latter.


\subsection{Comparison with Existing Methods}
To compare UniIR with existing retrievers, we also evaluate the homogeneous setting where the retriever only needs to retrieve from the task-specific pool, which is more consistent with the traditional IR setup. Additionally, we conducted held-out experiments to examine UniIR's zero-shot generalization ability on task-specific pools M-BEIR$_{\text{local}}$ in comparison to baseline models.

\paragraph{UniIR vs Zero-shot Retrievers.}
In Table~\ref{tab:multitask}, we demonstrate that while SigLIP attains the highest average value of zero-shot SoTA retrievers with an average value of 37.2\% on R@5, our UniIR models (CLIP$_{\text{SF}}$) and (BLIP$_{\text{FF}}$) surpass it by a significant margin, with average R@5 values of 50.6\% and 46.8\% respectively.

\paragraph{UniIR vs Single-task Tuning.}
Table~\ref{tab:multitask} demonstrates the advantages of multi-task instruction-tuning in the UniIR framework over single-task fine-tuning. Our findings indicate that UniIR (BLIP$_{\text{FF}}$) greatly outperforms its single-task counterpart by an average of 9.7\% on R@5, and exhibits significant improvements on task 7 compositional image retrieval such as CIRR with 48.6\% compared to 13.9\%. UniIR (CLIP$_{\text{SF}}$) also demonstrates an overall improvement of 1.2\%, particularly on Fashion200K and Infoseek. In contrast, we observed that the multi-task training without instructions would not lead to such improvements on average for CLIP$_{\text{SF}}$, as it remained at 49.4\%.

\begin{figure}
    \centering
    \includegraphics[width=0.9\linewidth]{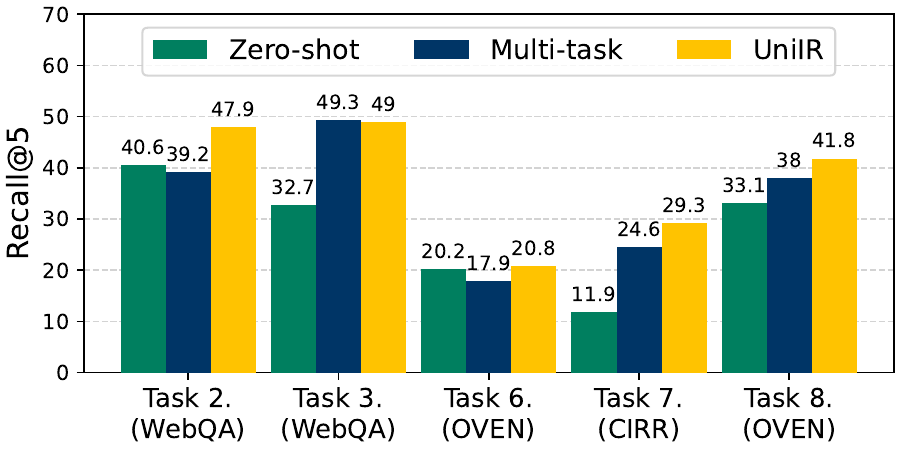}
    \caption{\textbf{Held-out dataset generalization experiments} on M-BEIR$_{\text{local}}$: we train a Multi-task and a UniIR model on 7 held-in datasets and test on 3 held-out datasets (WebQA, OVEN, CIRR). Results are averaged over CLIP$_{\text{SF}}$ and BLIP$_{\text{FF}}$.}
    \label{fig:union_held_out_local}
    \vspace{-15pt}
\end{figure}

\paragraph{Generalization Performance on Held-Out Datasets and Tasks}
In Figure~\ref{fig:union_held_out_local}, we showed the average zero-shot performance of SoTA CLIP and BLIP retrievers, the average zero-shot performance of multi-task fine-tuned baselines, and the average zero-shot performance of UniIR (CLIP$_{\text{SF}}$) and UniIR (BLIP$_{\text{FF}}$) on 3 held-out datasets on M-BEIR$_{\text{local}}$. The UniIR models exhibit superior generalization ability on unseen tasks and datasets. As shown in Figure~\ref{fig:union_held_out_local}, UniIR models consistently outperform the SoTA retrievers and multi-task training baselines over 3 held-out datasets across 5 tasks. On the other hand, Multi-task training without using instruction shows moderate improvements over the SoTA retriever baselines and performs even worse in tasks such as WebQA (task 2) and OVEN (task 6).


\section{Related Work }
\paragraph{Multimodal Information Retrieval.}
In recent years, the field of cross-modal information retrieval has seen significant exploration, with a particular emphasis on image-to-text matching.
Datasets such as MSCOCO~\citep{karpathy2015deep} and Flickr30k~\cite{plummer2015flickr30k} have become standard benchmarks for evaluating the progress of pre-trained vision-language models such as ALIGN~\cite{jia2021scaling}, VILT~\cite{kim2021vilt}, ALBEF~\cite{li2021align}, MURAL~\cite{jain2021mural}, and ImageBind~\cite{girdhar2023imagebind}. However, fine-grained image retrieval often hinges on the ability to articulate intents through text, presenting challenges in multimodal queries~\cite{changpinyo2021telling} such as ReMuQ~\cite{luo2023end}.
While the text-to-text retrieval benchmark BEIR~\cite{thakur2021beir} has advanced research in building generalized zero-shot text retrieval systems, a unified multimodal information retrieval benchmark covering a diverse range of tasks remains absent. 
We hope that the introduction of the M-BEIR will accelerate progress toward more general multimodal information retrieval models.

\paragraph{Retrieval-augmented Models.}
Retrieval augmentation has been studied extensively in the past few years. ORQA~\cite{lee2019latent}, RAELM~\cite{guu2020retrieval}, RAG~\cite{lewis2020retrieval} and FID~\cite{izacard2021leveraging} are among the earliest work to learn retriever and language model jointly from weakly supervised dataset. Later on, RETRO~\cite{borgeaud2022improving} scaled the retrieval-augmented training to large-scale language models to show great performance gain with a relatively small-sized language model. ATLAS~\cite{izacard2022few} further extends the idea of retrieval-augmented training to few-shot learning and shows performance gain on broader knowledge-intensive tasks. These works are mostly focused on retrieving text paragraphs to augment language models. Later on, RA-C3M\cite{yasunaga2023retrieval}, REVEAL~\cite{hu2023reveal}, and MuRAG~\cite{chen2022murag} have shown the advantage of retrieving multimodal content from Wikipedia to answer visually information-seeking questions.  The multimodal augmentation idea is also applied to image generation with several works like KNN-difussion\cite{sheynin2023knndiffusion} Re-imagen~\cite{chen2022re}, and RA-diffusion\cite{blattmann2022semiparametric}. The closest to our work is the TART~\cite{asai2022task}, which also aims to build a single retriever to retrieve different content based on the instruction. However, TART is still only focused on text-to-text modality.

\paragraph{Instruction tuning.}
Instruction-tuned models, where models are trained to follow user instructions, have emerged as a significant area of research in large language models (LLMs)~\cite{chowdhery2022palm, OpenAI2023GPT4TR, touvron2023llama}. FLAN and FLAN-T5~\cite{wei2021finetuned, chung2022scaling} have demonstrated capabilities to generalize to unseen natural language tasks~\cite{Mishra2021CrossTaskGV} and
InstructGPT~\cite{ouyang2022training} further illustrates how instruction tuning can align language models more closely with users' intentions. 
Recently, visual instruction tuning has been explored in vision-language tasks such as visual question answering with models such as InstructBLIP~\cite{Dai2023InstructBLIPTG} and LLaVA~\cite{liu2023visual} or datasets such as the MultiInstruct~\cite{xu2022multiinstruct}. 
On image diffusion models, InstructPix2Pix \cite{brooks2023instructpix2pix} and MagicBrush \cite{zhang2023magicbrush} dataset show how the diffusion model can follow instructions to edit images.
However, the most closely related retrieval-augmented models, such as InstructRETRO \cite{wang2023instructretro}, RA-DIT \cite{lin2023ra}, as well as embedding models like OneEmbedder \cite{su2022one} and TART \cite{asai2022task}, remain text-only.
In contrast, UniIR demonstrates promising cross-dataset generalization in multimodal retrieval, indicating potential advancements for integration into multimodal LLMs.
\section{Conclusion}
We presented UniIR, a framework to build universal multimodal information retrieval models. This framework enables one unified retriever to follow natural language instruction and accomplish diverse information retrieval tasks across different modalities.
We build the M-BEIR benchmark to enable the training and evaluation of UniIR models. We show that our proposed instruction-tuning pipeline can generalize well across different retrieval tasks and domains. However, the existing model performance is still relatively far from perfect indicating ample room for future improvement. We believe that large-scale pre-training algorithms with a stronger vision-language backbone model can build the foundation towards closing this gap and would leave this direction for future exploration.


\section*{Author Contributions}
CW and YC led the project. The authors had overlapping responsibilities, but the biggest contributions from each author were as follows:
\begin{itemize}
\item CW co-designed the experiment and, in particular, was responsible for collecting and processing the M-BEIR dataset, creating the codebase for the dataset utilities and UniIR models, creating the pipeline for training and evaluating retriever models as well as FAISS retrieval, conducting experiments for the main results and ablation studies, and writing the paper.
\item YC co-designed the experiment and, in particular, was responsible for designing the zero-shot model's codebase, conducting experiments for the zero-shot baselines, and writing the paper.
\item HC contributed to the code, particularly the BLIP model.
\item GZ and JF contributed to the final writeup.
\item HH, AR and WC contributed advice on the project, as well as feedback on writing and presentation.
\end{itemize}

\section*{Acknowledgments}
Yang Chen and Alan Ritter are supported by the NSF (IIS-2052498) and by the Office of the Director of National Intelligence (ODNI), Intelligence Advanced Research Projects Activity (IARPA), via the HIATUS Program contract \#2022-22072200004. The views and conclusions contained herein are those of the authors and should not be interpreted as necessarily representing the official policies, either expressed or implied, of NSF, ODNI, IARPA, or the U.S. Government. The U.S. Government is authorized to reproduce and distribute reprints for governmental purposes notwithstanding any copyright annotation therein.

{
    \small
    \bibliographystyle{ieeenat_fullname}
    \bibliography{main}
}

\clearpage
\setcounter{page}{1}
\maketitlesupplementary


\subsection{Visualization of Retrieval Results}
We present example retrieval results for all 10 datasets in the M-BEIR benchmark, shown in Figures \ref{fig:retrieval_results_visualnews} through \ref{fig:retrieval_results_infoseek}.
For each dataset, we show one example query and the top 5 candidates retrieved from M-BEIR (5.6 M) by UniIR (CLIP$_{\text{SF}}$), UniIR (BLIP$_{\text{FF}}$), and Multi-task fine-tuned baseline retrievers Multi-task (CLIP$_{\text{SF}}$) and Multi-task (BLIP$_{\text{FF}}$).
Our visualization further demonstrates the findings that we have discussed in Section~\ref{sec:experiment}. 

\paragraph{Instruction-tuning improves retrieval on M-BEIR.}
For most of the tasks, the multi-task baselines retrieved candidates from undesired modalities. As demonstrated in Figures~\ref{fig:retrieval_results_visualnews},~\ref{fig:retrieval_results_mscoco},~\ref{fig:retrieval_results_fashion200k} and \ref{fig:retrieval_results_edis}, the multi-task baselines only retrieved text candidates that had a similar meaning to the query and failed to understand the intent of the task. For instance, in the VisualNews example~\ref{fig:retrieval_results_visualnews}, the multi-task baselines are distracted by text that contains the entity names ``Obama" and ``Gates". Similarly, in the WebQA example~\ref{fig:retrieval_results_webqa}, baseline models are distracted by text that contains ``Saint Peter". On the other hand, UniIR models were able to accurately fetch candidates in the desired modalities.

\paragraph{Single-modality retrieval task.}
Even without instructions, the baseline models can perform well in single-modality retrieval tasks, such as similar image retrieval in Figure~\ref{fig:retrieval_results_nights}. This is consistent with the results presented in Section~\ref{sec:experiment}, as these tasks do not require a model to understand intent.

\paragraph{Vision and language composed image retrieval task.}
In Table~\ref{tab:union_small}, it is evident that UniIR (BLIP$_{\text{FF}}$) outperforms UniIR (CLIP$_{\text{SF}}$) in the composed image retrieval task (task 7). The composed image retrieval task generally involves modifying a reference image according to a textual description. Our hypothesis is that the ``cross-attention" layer in (BLIP$_{\text{FF}}$) can effectively merge the image and text embeddings, making it better suited for this task. Our presented examples~\ref{fig:retrieval_results_fashioniq} and~\ref{fig:retrieval_results_cirr} indicate that models based on BLIP$_{\text{FF}}$ can accurately retrieve the target candidate.

\paragraph{InfoSeek and OVEN tasks.}
InfoSeek task is more challenging as compared to the OVEN task, which mainly focuses on visual entity recognition. Infoseek queries require details from Wikipedia articles. As illustrated in Example~\ref{fig:retrieval_results_infoseek}, only UniIR (BLIP$_{\text{FF}}$) successfully retrieved the target candidate in the top five results. Although UniIR (CLIP$_{\text{SF}}$) retrieved 5 related candidates that correctly identified the lake in the query image, none of them contained the desired answer.

\subsection{Experiment Details}

\paragraph{Zero-shot SoTA Retriever.}
We utilize pre-trained vision-language models such as CLIP (L-14)\citep{radford2021learning}, SigLIP (L)\citep{zhai2023sigmoid}, BLIP (L)~\citep{li2022blip}, and BLIP2~\citep{li2023blip} as our baseline feature extractors. 
We use score-level fusion with $w_1=w_2=1$ to fuse multimodal features (i.e., element-wise addition). 
For task 8 (OVEN and InfoSeek),  which contains image-text queries and candidates, we exclusively use the image for the input query and candidate, as we found it achieves better performance in preliminary studies than using both modalities. In addition, we use the Wikipedia page title as the candidate with the prompt ``\texttt{a picture of [title]}'' as we found this approach gives better zero-shot performance compared to using 100 tokens from the Wikipedia page as the candidate.
Although BLIP and BLIP2 inherently support multimodal feature extraction, these features were not pre-trained specifically for retrieval tasks. 
Therefore, we apply the same approach as CLIP to extract features (dim=256). We use the CLIP (\texttt{ViT-L-14}), BLIP (\texttt{BLIP/models/model\_large.pth}) and BLIP2 (\texttt{BLIP2/blip2\_pretrained.pt}) models from the LAVIS~\cite{li-etal-2023-lavis} 
library with the supported \texttt{feature\_extractor} function. For SigLIP
(\texttt{timm/ViT-L-16-SigLIP-256}) model, we use the Open-CLIP~\cite{ilharco_gabriel_2021_5143773} library.

\paragraph{UniIR Models and Multi-task fine-tuned Models.}
We use the largest checkpoint, ViT-L14~\cite{dosovitskiy2020image}, for all CLIP and BLIP variants. The default image resolution is $224 \times 224$, and for UniIR (BLIP$_{\text{FF}}$), we also report the finetuned results with a resolution of $384 \times 384$, which is a commonly used setting for BLIP finetuning.
Our batch size is 105 for CLIP variants and 115 for BLIP variants on the $224 \times 224$ resolution. We train the model for 20 epochs using other hyperparameters as reported in the original implementations.
For score fusion methods, we set $w_1=w_2=w_3=w_4=1$ by default.
CLIP$_{\text{FF}}$ uses a 2-layer transformer architecture similar to the T5 Transformer~\cite{radford2021learning}, but with only 2 layers and 12 attention heads, with each head having 64 dimensions.
BLIP$_{\text{FF}}$ follows the original implementation of BLIP. The output from the image encoder is fed into transformer layers of the image-grounded text encoder through cross-attention layers, and then the output is treated as the fused feature.
We train all our models using in-batch query-candidate contrastive loss~\cite{radford2021learning} to maximize the positive pair similarity while minimizing negative pair similarity.
All of our experiments are conducted on a single node with 8 H100 GPUs.  


\subsection{Data Collection}
\paragraph{M-BEIR Format.}
Each query instance in the M-BEIR dataset includes a query $\mathbf{q}$, a list of relevant(positive) candidate data $\mathbf{c}^+$, and a list of potentially available irrelevant(negative) candidate data $\mathbf{c}^-$. In addition, a human-authored instruction $q_{\texttt{inst}}$ is provided with $\mathbf{q}$ to specify the intent of the retrieval task. It's important to note that every M-BEIR query instance has at least one positive candidate data and possibly no negative candidate data. 

\paragraph{VisualNews.}
We follow the preprocessing pipeline outlined in the Visual News dataset~\cite{liu2020visual} and randomly sampled 200K, 40K, and 40K image-caption pairs for our training, validation, and test sets, respectively. Then, converted them into M-BEIR format. As a result, we have 100K instances for task 1 $q_t \to c_i$ and 100K instances for task 4 $q_i \to c_t$ in the training set. We used all the images and captions in the Visual News dataset as the initial candidate pool, which amounted to a total of 2.5M candidates. We then trimmed down the candidate pool to 1M, ultimately arriving at a final M-BEIR$_\text{local}$ pool of 500K text and 500K images.

\paragraph{Fashion200K.}
The Fashion200K dataset~\cite{han2017automatic} consists of image and product description pairs. We use all the available images and descriptions as our M-BEIR$_{\text{local}}$ candidate pool, totalling 260K entries, with 200K images and 60K text descriptions. We randomly selected 30K image-description pairs from the original training set to form our training data. The original test data is evenly divided into a validation and test set. We converted the dataset into M-BEIR format. In total, we have 15K task 1 ($q_t \to c_i$) instances and 15K task 4 ($q_i \to c_t$) instances in training set.

\paragraph{MSCOCO.}
We used the Karpathy split~\cite{karpathy2015deep} for MSCOCO, which contains 113K/5K/5K images for train/validation/test. We directly converted MSCOCO into M-BEIR format. This results in 113K task 4 ($q_i \to c_t$) instance and 566K task 1 ($q_t \to c_i$) instances for the training set. We then trimmed the task 1 instance down to 100k. When evaluating models on the MSCOCO test split, we use all the images and captions from the original test set to construct the test split M-BEIR$_{\text{local}}$ candidate pool, which has 25K text and 5K images.

\paragraph{WebQA.}
We followed the full-scale retrieval setting in WebQA~\cite{chang2022webqa} and constructed the 940K M-BEIR$_{\text{local}}$ candidate pool. This pool contains 400K pairs of (image, text) and 540K text-only candidates. We convert the WebQA dataset into the M-BEIR format by using questions as queries. We randomly sample 5000 instances from the original training set as our test set.  As a result, the M-BEIR WebQA training set consists of 15K task 2 instances of $q_i \to c_t$ and 15K task 3 instances of $q_i \to (c_i, c_t)$.

\paragraph{EDIS.}
We followed the full candidate retrieval setting in EDIS~\cite{liu2023edis} and use all the 1M image-headline candidates in EDIS as the M-BEIR$_{\text{local}}$ candidate pool. We directly convert the original train/val/test split of 26K/3.2K/3.2K instances into M-BEIR format, which involves using the caption as the query and the image-headline pair as the positive candidate. This process results in 26K task 3 $q_i \to (c_i, c_t)$ instances in the training set.

\paragraph{NIGHTS.}
To convert the 20K triplets in the 2AFC NIGHTS dataset~\cite{fu2023learning} into M-BEIR format, we used the reference image as the query. For the positive candidate, we selected the image target aligned with human judgment, and for the negative candidate, we chose the other image target that disagreed with human judgment. This results in 16K/2K/2K task 5 $q_i \to c_i$ instances for train/val/test split. We use all the 40K images in the NIGHTS dataset as the M-BEIR$_{\text{local}}$ task-specific candidate pool.

\paragraph{FashionIQ.}
The FashionIQ dataset~\cite{wu2021fashion} consists of pairs of reference and target images alongside two sets of captions. However, since the captions are not detailed enough, we follow the approach of~\cite{liu2021image} and concatenated them into a single caption. To convert the FashionIQ dataset into M-BEIR format, we use the reference image and concatenated caption as a query and the target image as a positive candidate. All images in FashionIQ serve as the candidate pool for M-BEIR$_{\text{local}}$. We randomly sampled 1.7K instances from the converted training set as our validation set and used the original validation set as the test set.

\paragraph{CIRR.}
The CIRR dataset~\cite{liu2021image}, comprises pairs of reference and target images, along with a modification sentence that describes the changes made. We consider the reference image and modification sentence as the query and the target image as a positive candidate for a task 7 $(q_i, q_t) \to c_i$ instance. We use all the images in the CIRR dataset as the M-BEIR$_{\text{local}}$ candidate pool. For validation purposes, we randomly select 2K instances from the training set, and the original validation set is used as the test set.

\paragraph{OVEN.}
The OVEN dataset~\cite{hu2023open} has instances that include an image and a visual recognition text question. Additionally, it has a related image(potentially empty) and its corresponding text description as the target candidate. In order to convert this dataset into M-BEIR format, we treat the image and text question pair as a query, and the image and text description pair as a positive candidate. The text description in OVEN~\cite{hu2023open} is simply the title of the Wikipedia subject. However, in order to create the candidate text, we concatenate the Wikipedia title with the first 100 tokens of its summary. This allows for a more comprehensive understanding of the text description. This results in 4M training instances for task 6 $(q_i, q_t) \to c_t$ and 700K instances for task 8 $(q_i, q_t) \to (c_i, c_t)$. We trimmed the instances down to 120K each. We also trimmed down the original 6M candidates pool to 1M and adopted it as M-BEIR$_{\text{local}}$.

\paragraph{InfoSeek.}
The Infoseek dataset~\cite{chen2023infoseek} comprises image and text question pairs, along with an associated image (which may be empty) and a corresponding text description that contains the answer. To convert this dataset to M-BEIR format, we treat the image and text question pair as a query, and the image and text description pair as a positive candidate. However, the text description in Infoseek~\cite{chen2023infoseek} is an entire Wikipedia article with thousands of tokens. To create a proper retrieval task, we split the Wikipedia article into snippets of 100 tokens and use the snippet which contains the exact answer as the positive candidate, and the rest as negative candidates. This results in 400K training instances for Task 6 $(q_i, q_t) \to c_t$, and 300K instances for Task 8 $(q_i, q_t) \to (c_i, c_t)$. We trimmed the instances down to 140K each. We also trimmed down the original 6M candidates pool to 1M and adopted it as M-BEIR$_{\text{local}}$.

We removed distorted text or images and resized the image to 256 pixels for the shorter dimension in all the datasets. 

\paragraph{M-BEIR Candidates Pool.}
The M-BEIR (5.6M) candidate pool is created by combining M-BEIR$_{\text{local}}$ pools from all 10 datasets.

\subsection{Additional Experiment Analysis}
\paragraph{Zero-shot Retriever Results.}
In Table~\ref{tab:multitask}, we benchmark zero-shot retrieval models on each task and found that SigLIP performed the best on average. Interestingly, while the BLIP model performed on par with SigLIP or even better than CLIP on the MSCOCO dataset, which focuses on common objects, it underperformed significantly on other datasets in domains such as news, fashion, and Wikipedia. However, we observed that SigLIP significantly outperformed all other models on Fashion200K (36.5) and FashionIQ (14.4), while CLIP had a clear advantage on news datasets such as VisualNews (43.3). This suggests that covering a wide range of visual domains during pre-training, rather than focusing on a single domain, is crucial for achieving robust retrieval. M-BEIR presents a comprehensive evaluation benchmark to serve this purpose.

\paragraph{Held-out Dataset Generalization.}
In Table~\ref{tab:held_out_union} and~\ref{tab:held_out_local}, we report full results of each model used in Figure~\ref{fig:union_held_out} and ~\ref{fig:union_held_out_local} on M-BEIR and M-BEIR$_{\text{local}}$, respectively.
\definecolor{positive}{HTML}{008000} 
\definecolor{negative}{HTML}{FF0000}

\begin{table}[ht!]
    \centering
    \large
    \resizebox{0.49\textwidth}{!}{%
    \begin{tabular}{llcc|ccc|ccc}
    \toprule
    &   & \textbf{ZS} & \textbf{ZS} & \textbf{ZS}&
    \textbf{Multi.} & \textbf{UniIR} & \textbf{ZS} & \textbf{Multi.} & \textbf{UniIR}\\
    \cmidrule(lr){3-3} \cmidrule(lr){4-4}\cmidrule(lr){5-5} \cmidrule(lr){6-6}\cmidrule(lr){7-7}\cmidrule(lr){8-8}\cmidrule(lr){9-9}\cmidrule(lr){10-10}
        \textbf{Task}        & \textbf{Dataset}  & SigLIP & BLIP2 &  CLIP  &  CLIP$_{\text{SF}}$ & CLIP$_{\text{SF}}$  & BLIP & BLIP$_{\text{FF}}$ & BLIP$_{\text{FF}}$\\
         \midrule
         2.
         & WebQA     & 34.1 & 35.2 & 32.1  &35.0 &51.8 & 38.1 & 36.7 & 37.1\\
         \midrule
         3.
         & WebQA     & 2.2 & 0.0 &5.5  &10.2 &20.8  & 0.0 & 30.4 & 35.3\\
         \midrule
         6.
         & OVEN      & 0.0 & 0.0 &0.0  &8.9 &6.2  & 0.0 & 10.9 & 12.0\\
         \midrule
         7.
         & CIRR      & 7.1 & 6.2 &5.4  &14.1 &16.9  & 7.4 & 15.7 & 34.1\\
         \midrule
         8.
         & OVEN      & 27.2 & 13.8 &24.5  &34.4 &43.7 & 10.1 & 20.5 & 24.9\\
         \midrule
         - & Average & 14.1 & 11.0 & 13.5 &20.5 &27.9  & 11.1 & 22.8 & 28.7 \\
         \bottomrule
    \end{tabular}
    }
    \caption{\textbf{Held-out dataset generalization} experiments (Recall@5) on M-BEIR: we train a Multi-task (Multi.) and a UniIR model on 7 held-in datasets and test on 3 held-out datasets (WebQA, OVEN, CIRR).}
    \label{tab:held_out_union}
    \vspace{-2pt}
\end{table}
\definecolor{positive}{HTML}{008000} 
\definecolor{negative}{HTML}{FF0000}

\begin{table}[ht!]
    \centering
    \large
    \resizebox{0.49\textwidth}{!}{%
    \begin{tabular}{llcc|ccc|ccc}
    \toprule
    &   & \textbf{ZS} & \textbf{ZS} & \textbf{ZS} & 
    \textbf{Multi.} & \textbf{UniIR} & \textbf{ZS} & \textbf{Multi.} & \textbf{UniIR}\\
    \cmidrule(lr){3-3} \cmidrule(lr){4-4}\cmidrule(lr){5-5} \cmidrule(lr){6-6}\cmidrule(lr){7-7}\cmidrule(lr){8-8}\cmidrule(lr){9-9}\cmidrule(lr){10-10}
        \textbf{Task}        & \textbf{Dataset}  &  SigLIP & BLIP2 & CLIP  &  CLIP$_{\text{SF}}$ & CLIP$_{\text{SF}}$ & BLIP & BLIP$_{\text{FF}}$ & BLIP$_{\text{FF}}$\\
         \midrule
         2.
         & WebQA     & 39.8 & 38.6 &36.3  &37.0 &54.5& 44.9 & 41.3 & 41.3\\
         \midrule
         3.
         & WebQA     & 43.5 & 24.5 &45.1  &46.8 &47.7  & 20.3 & 51.7 & 50.3\\
         \midrule
         6.
         & OVEN      & 29.7 & 12.2 &24.2  &21.6 &25.0 & 16.1 & 14.2 & 16.5\\
         \midrule
         7.
         & CIRR      & 22.7 & 11.8 &13.2  &19.7 &21.4  & 10.6 & 29.4 & 37.1\\
         \midrule
         8.
         & OVEN      & 41.7 & 27.3 &38.8  &50.3 &52.8 & 27.4 & 25.6 & 30.7\\
         \midrule
         - & Average & 35.5 & 22.9 & 31.5 &35.1 &40.3& 23.8 & 32.4 & 35.2 \\
         \bottomrule
    \end{tabular}
    }
    \caption{\textbf{Held-out dataset generalization} experiments (Recall@5) on M-BEIR$_{\text{local}}$: we train a Multi-task (Multi.) and a UniIR model on 7 held-in datasets and test on 3 held-out datasets (WebQA, OVEN, CIRR).}
    \label{tab:held_out_local}
    \vspace{-2pt}
\end{table}
\definecolor{lightblue}{RGB}{204,229,255} 
\definecolor{lightpurple}{RGB}{230,230,250} 
\definecolor{lightred}{RGB}{255,204,204} 
\definecolor{lightgreen}{RGB}{204,255,204} 
\definecolor{green}{RGB}{0,180,0} 

\begin{figure*}[t!]
\centering
\scriptsize
\begin{tabular}{>{\columncolor{lightblue}}m{1.5cm} >{\columncolor{lightblue}}m{1cm} m{2.2cm} m{3.8cm} m{2.4cm} m{3.8cm}}
    \toprule
    \textbf{Dataset} & \textbf{Domain} & \textbf{Task} & \textbf{Query Instruction  $q_{\texttt{inst}}$} & \textbf{Query Image $q_i$} & \textbf{Query Text $q_t$} \\
    \midrule
    VisualNews 
    & News
    & 1. $q_t$ $\to$ $c_i$
    & Based on the caption, provide the most fitting image for the news story. 
    & -
    &
    Robert Gates and President Obama salute during a farewell ceremony for Gates as outgoing defense secretary Gates served in that position under Obama and President George W Bush He was known as a savvy manager of the bureaucracy and a ruthlessly efficient defense secretary In his new memoir Gates asserts that Obama had more than doubts about his course in Afghanistan.
\end{tabular}
\centering
    \scriptsize
    \begin{center}
    \begin{tabular}{>{\columncolor{lightblue}}m{1.2cm}m{2.7cm}m{2.7cm}m{2.7cm}m{2.7cm}m{2.7cm}}
            \toprule
            \textbf{Model} & \textbf{Rank 1 $c_i$} & \textbf{Rank 2 $c_i$} & \textbf{Rank 3 $c_i$} & \textbf{Rank 4 $c_i$} & \textbf{Rank 5 $c_i$} \\
            \midrule
                  UniIR
                  \newline (CLIP$_{\text{SF}}$)
                  \newline \checkmark inst
                  &  
                  \Large \textcolor{red}{\texttimes}
                  \includegraphics[width=20mm,height=20mm]{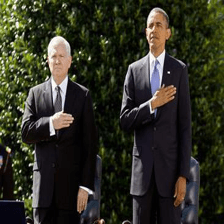}
                  &  
                  \Large \textcolor{red}{\texttimes}
                  \includegraphics[width=20mm,height=20mm]{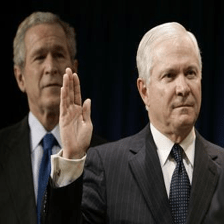} 
                  &  
                  \Large \textcolor{red}{\texttimes}
                  \includegraphics[width=20mm,height=20mm]{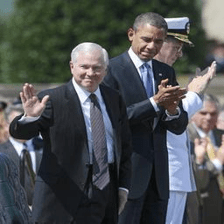} 
                  & 
                  \Large \textcolor{red}{\texttimes}
                  \includegraphics[width=20mm,height=20mm]{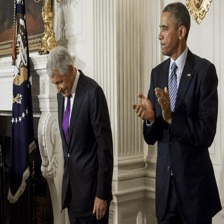} 
                  & 
                  \large \textcolor{green}{\checkmark}
                  \includegraphics[width=20mm,height=20mm]{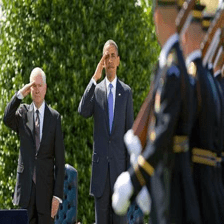}
                  \\
            \midrule
                  multi-task
                  \newline (CLIP$_{\text{SF}}$)
                  \newline \ding{55} inst
                  &  
                  \Large \textcolor{red}{\texttimes}
                  \tiny Robert Gates and President Obama salute during a farewell ceremony for Gates as outgoing defense secretary Gates served in that position under Obama and President George W Bush He was known as a savvy manager of the bureaucracy and a ruthlessly efficient defense secretary In his new memoir Gates asserts that Obama had more than doubts about his course in Afghanistan.
                  &  
                  \Large \textcolor{red}{\texttimes}
                  \tiny Former Defense Secretary Robert Gates acknowledges applause as President Obama speaks at Gates s farewell ceremony at the Pentagon in June 2011.
                  &  
                  \Large \textcolor{red}{\texttimes}
                  \tiny Former Defense secretary Robert Gates is seen with President Obama.
                  & 
                  \Large \textcolor{red}{\texttimes}
                  \tiny President Obama and Robert Gates.
                  & 
                  \Large \textcolor{red}{\texttimes}
                  \tiny Obama presents Gates with the Presidential Medal of Freedom The citation said Gates had selflessly dedicated his life to ensuring the security of the American people Gates cultivated an eventempered image in his years in government.
                  \\
            \midrule
                  UniIR
                  \newline (BLIP$_{\text{FF}}$)
                  \newline \checkmark inst
                  &  
                  \Large \textcolor{red}{\texttimes}
                  \includegraphics[width=20mm,height=20mm]{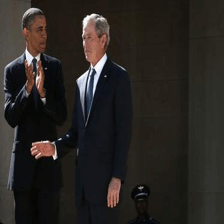}
                  &  
                  \Large \textcolor{red}{\texttimes}
                  \includegraphics[width=20mm,height=20mm]{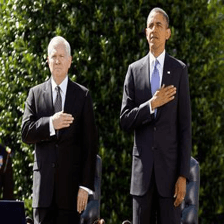} 
                  &  
                  \large \textcolor{green}{\checkmark}
                  \includegraphics[width=20mm,height=20mm]{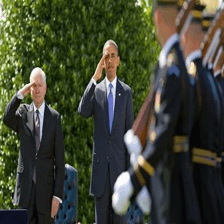} 
                  & 
                  \Large \textcolor{red}{\texttimes}                  
                  \includegraphics[width=20mm,height=20mm]{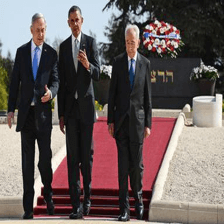} 
                  & 
                  \Large \textcolor{red}{\texttimes}   
                  \includegraphics[width=20mm,height=20mm]{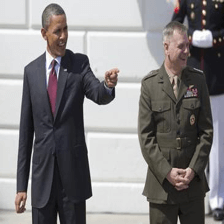}
                  \\
            \midrule
                  multi-task
                  \newline (BLIP$_{\text{FF}}$)
                  \newline \ding{55} inst
                  &  
                  \Large \textcolor{red}{\texttimes}
                  \tiny Robert Gates and President Obama salute during a farewell ceremony for Gates as outgoing defense secretary Gates served in that position under Obama and President George W Bush He was known as a savvy manager of the bureaucracy and a ruthlessly efficient defense secretary In his new memoir Gates asserts that Obama had more than doubts about his course in Afghanistan.
                  &  
                  \Large \textcolor{red}{\texttimes}
                  \tiny Former Defense Secretary Robert Gates acknowledges applause as President Obama speaks at Gates s farewell ceremony at the Pentagon in June 2011.
                  &  
                  \Large \textcolor{red}{\texttimes}
                  \includegraphics[width=20mm,height=20mm]{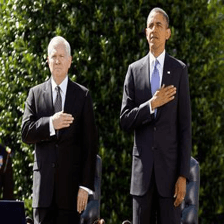}
                  & 
                  \Large \textcolor{red}{\texttimes}
                  \tiny Former Defense secretary Robert Gates is seen with President Obama.
                  & 
                  \Large \textcolor{red}{\texttimes}
                  \includegraphics[width=20mm,height=20mm]{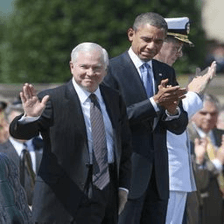}
                  \\
            \midrule
                  Zero-shot
                  \newline (BLIP2)
                  \newline \ding{55} inst
                  &  
                  \Large \textcolor{red}{\texttimes}
                  \tiny Robert Gates and President Obama salute during a farewell ceremony for Gates as outgoing defense secretary Gates served in that position under Obama and President George W Bush He was known as a savvy manager of the bureaucracy and a ruthlessly efficient defense secretary In his new memoir Gates asserts that Obama had more than doubts about his course in Afghanistan.
                  &  
                  \Large \textcolor{red}{\texttimes}
                  \tiny Former Defense Secretary Robert Gates acknowledges applause as President Obama speaks at Gates s farewell ceremony at the Pentagon in June 2011.
                  &  
                  \Large \textcolor{red}{\texttimes}
                  \tiny Former US president George W Bush center shakes hands with his father former President George HW Bush as they attend the opening ceremony of the George W Bush Presidential Center on April 25 2013 in Dallas Texas.
                  & 
                  \Large \textcolor{red}{\texttimes}
                  \tiny President Obama right and Senior Chief Special Warfare Operator Edward Byers Jr arrive for a ceremony at the White House where Byers received the Medal of Honor for his courageous actions in Afghanistan in 2012.
                  & 
                  \Large \textcolor{red}{\texttimes}
                  \tiny President Obama right and former president George W Bush arrive for a wreathlaying ceremony for victims of the 1998 bombing at the US Embassy in Dar es Salaam Tanzania.
                  \\
            \bottomrule
    \end{tabular}
    \caption{Visualization of the Top 5 Candidates Retrieved from M-BEIR (5.6M candidates) by Different Models for a VisualNews Query}
    \label{fig:retrieval_results_visualnews}
    \end{center}
\end{figure*}

\begin{figure*}[t!]
\centering
\scriptsize
\begin{tabular}{>{\columncolor{lightpurple}}m{1.5cm} >{\columncolor{lightpurple}}m{1cm} m{2.2cm} m{3.8cm} m{2.4cm} m{3.8cm}}
    \toprule
    \textbf{Dataset} & \textbf{Domain} & \textbf{Task} & \textbf{Query Instruction  $q_{\texttt{inst}}$} & \textbf{Query Image $q_i$} & \textbf{Query Text $q_t$} \\
    \midrule
    MSCOCO 
    & Common
    & 1. $q_t$ $\to$ $c_i$
    & Show me an image that best captures the following common scene description. 
    & -
    &
    An antilope is eating grass in between two zebra.
\end{tabular}

\centering
    \scriptsize
    \begin{center}
    \begin{tabular}{>{\columncolor{lightpurple}}m{1.2cm}m{2.7cm}m{2.7cm}m{2.7cm}m{2.7cm}m{2.7cm}}
            \toprule
            \textbf{Model} & \textbf{Rank 1 $c_i$} & \textbf{Rank 2 $c_i$} & \textbf{Rank 3 $c_i$} & \textbf{Rank 4 $c_i$} & \textbf{Rank 5 $c_i$} \\
            \midrule
                  UniIR
                  \newline (CLIP$_{\text{SF}}$)
                  \newline \checkmark inst
                  &  
                  \Large \textcolor{red}{\texttimes}
                  \includegraphics[width=20mm,height=20mm]{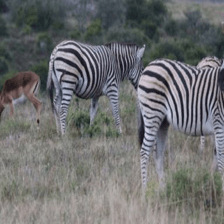}
                  &  
                  \large \textcolor{green}{\checkmark}
                  \includegraphics[width=20mm,height=20mm]{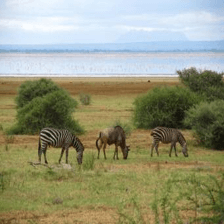} 
                  &  
                  \Large \textcolor{red}{\texttimes}
                  \includegraphics[width=20mm,height=20mm]{tables_and_figures/suppl/retrieval_results/CLIP_SF/Instruct/mscoco_ci_3.png} 
                  & 
                  \Large \textcolor{red}{\texttimes}
                  \includegraphics[width=20mm,height=20mm]{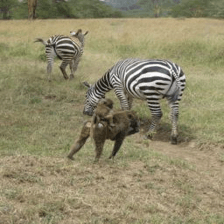} 
                  & 
                  \Large \textcolor{red}{\texttimes}
                  \includegraphics[width=20mm,height=20mm]{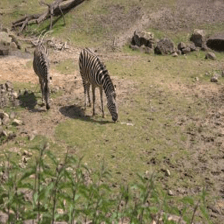} 
                  \\
            \midrule
                  multi-task
                  \newline (CLIP$_{\text{SF}}$)
                  \newline \ding{55} inst
                  &  
                  \Large \textcolor{red}{\texttimes}
                  \scriptsize An antilope is eating grass in between two zebra.
                  &  
                  \Large \textcolor{red}{\texttimes}
                  \scriptsize Two zebras and another animal grazing in the grass.
                  &  
                  \Large \textcolor{red}{\texttimes}
                  \scriptsize A couple of zebras stand next to a antelope.
                  & 
                  \Large \textcolor{red}{\texttimes}
                  \scriptsize Monkey standing behind two zebras as they graze.
                  & 
                  \Large \textcolor{red}{\texttimes}
                  \scriptsize Three zebras and two other animals grazing.
                  \\
            \midrule
                  UniIR
                  \newline (BLIP$_{\text{FF}}$)
                  \newline \checkmark inst
                  &  
                  \Large \textcolor{red}{\texttimes}
                  \includegraphics[width=20mm,height=20mm]{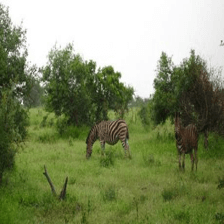}
                  &  
                  \Large \textcolor{red}{\texttimes}
                  \includegraphics[width=20mm,height=20mm]{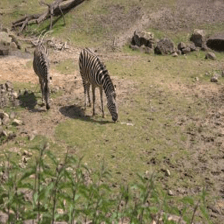} 
                  &  
                  \Large \textcolor{red}{\texttimes}                  
                  \includegraphics[width=20mm,height=20mm]{tables_and_figures/suppl/retrieval_results/BLIP_FF/Instruct/mscoco_3.png} 
                  & 
                  \Large \textcolor{red}{\texttimes}                  
                  \includegraphics[width=20mm,height=20mm]{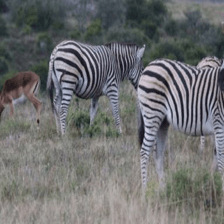} 
                  & 
                  \Large \textcolor{red}{\texttimes}   
                  \includegraphics[width=20mm,height=20mm]{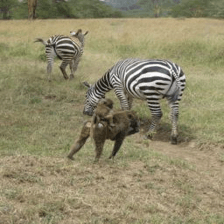}
                  \\
            \midrule
                  multi-task
                  \newline (BLIP$_{\text{FF}}$)
                  \newline \ding{55} inst
                  &  
                  \Large \textcolor{red}{\texttimes}
                  \newline \scriptsize An antilope is eating grass in between two zebra.
                  &  
                  \Large \textcolor{red}{\texttimes}
                  \newline \scriptsize Two zebras and another animal grazing in the grass.
                  &  
                  \Large \textcolor{red}{\texttimes}
                  \newline \scriptsize A couple of zebras graze on some grass.
                  & 
                  \Large \textcolor{red}{\texttimes}
                  \newline \scriptsize Two zebras are feeding on the grass by themselves.
                  & 
                  \Large \textcolor{red}{\texttimes}
                  \includegraphics[width=20mm,height=20mm]{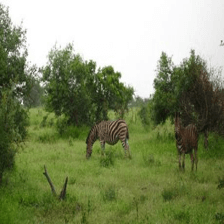}
                  \\
            \midrule
                  Zero-shot
                  \newline (BLIP2)
                  \newline \ding{55} inst
                  &  
                  \Large \textcolor{red}{\texttimes}
                  \scriptsize An antilope is eating grass in between two zebra.
                  &  
                  \Large \textcolor{red}{\texttimes}
                  \scriptsize A couple of zebra's leaned over eating grass in a field.
                  &  
                  \Large \textcolor{red}{\texttimes}
                  \scriptsize A zebra eating green grassy items for his lunch.
                  & 
                  \Large \textcolor{red}{\texttimes}
                  \scriptsize A zebra eating long clipped grass on the ground.
                  & 
                  \Large \textcolor{red}{\texttimes}
                  \scriptsize A zebra standing in a field grazing as another looks alert.
                  \\
            \bottomrule
    \end{tabular}
    \caption{Visualization of the Top 5 Candidates Retrieved from M-BEIR (5.6M candidates) by Different Models for an MSCOCO Query}
    \label{fig:retrieval_results_mscoco}
    \end{center}
\end{figure*}

\begin{figure*}[t!]
\centering
\scriptsize
\begin{tabular}{>{\columncolor{lightred}}m{1.5cm} >{\columncolor{lightred}}m{1cm} m{2.2cm} m{3.8cm} m{2.4cm} m{3.8cm}}
    \toprule
    \textbf{Dataset} & \textbf{Domain} & \textbf{Task} & \textbf{Query Instruction  $q_{\texttt{inst}}$} & \textbf{Query Image $q_i$} & \textbf{Query Text $q_t$} \\
    \midrule
    Fashion200K 
    & Fashion
    & 1. $q_t$ $\to$ $c_i$
    & Based on the following fashion description, retrieve the best matching image. 
    & -
    &
    Natural pebbled chevron alpaca drape front jacket.
\end{tabular}
\centering
    \scriptsize
    \begin{center}
    \begin{tabular}{>{\columncolor{lightred}}m{1.2cm}m{2.7cm}m{2.7cm}m{2.7cm}m{2.7cm}m{2.7cm}}
            \toprule
            \textbf{Model} & \textbf{Rank 1 $c_i$} & \textbf{Rank 2 $c_i$} & \textbf{Rank 3 $c_i$} & \textbf{Rank 4 $c_i$} & \textbf{Rank 5 $c_i$} \\
            \midrule
                  UniIR
                  \newline (CLIP$_{\text{SF}}$)
                  \newline \checkmark inst
                  &  
                  \Large \textcolor{red}{\texttimes}
                  \includegraphics[width=20mm,height=20mm]{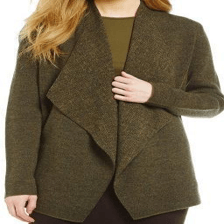}
                  &  
                  \Large \textcolor{red}{\texttimes}
                  \includegraphics[width=20mm,height=20mm]{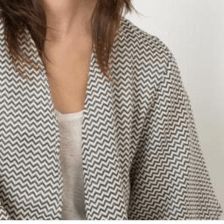} 
                  &  
                  \Large \textcolor{red}{\texttimes}
                  \includegraphics[width=20mm,height=20mm]{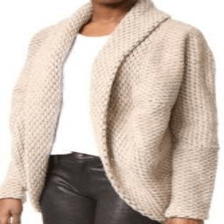} 
                  & 
                  \Large \textcolor{red}{\texttimes}
                  \includegraphics[width=20mm,height=20mm]{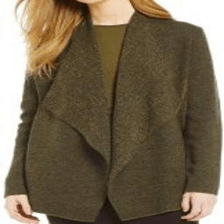} 
                  & 
                  \Large \textcolor{red}{\texttimes}
                  \includegraphics[width=20mm,height=20mm]{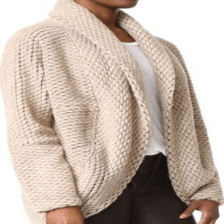} 
                  \\
            \midrule
                  multi-task
                  \newline (CLIP$_{\text{SF}}$)
                  \newline \ding{55} inst
                  &  
                  \Large \textcolor{red}{\texttimes}
                  \scriptsize Natural pebbled chevron alpaca drape front jacket.
                  &  
                  \Large \textcolor{red}{\texttimes}
                  \scriptsize Natural drape front open jacket.
                  &  
                  \Large \textcolor{red}{\texttimes}
                  \scriptsize Multicolor drape front jacket.
                  & 
                  \Large \textcolor{red}{\texttimes}
                  \scriptsize Multicolor crepe drape front jacket.
                  & 
                  \Large \textcolor{red}{\texttimes}
                  \scriptsize Multicolor drape front wool jacket.
                  \\
            \midrule
                  UniIR
                  \newline (BLIP$_{\text{FF}}$)
                  \newline \checkmark inst
                  &  
                  \Large \textcolor{red}{\texttimes}
                  \includegraphics[width=20mm,height=20mm]{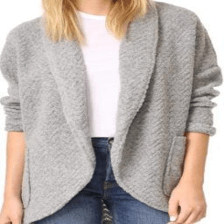}
                  &  
                  \Large \textcolor{red}{\texttimes}
                  \includegraphics[width=20mm,height=20mm]{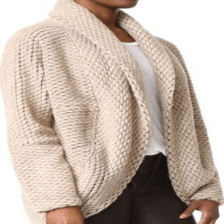} 
                  &  
                  \Large \textcolor{red}{\texttimes}                  
                  \includegraphics[width=20mm,height=20mm]{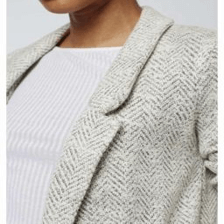} 
                  & 
                  \Large \textcolor{red}{\texttimes}                  
                  \includegraphics[width=20mm,height=20mm]{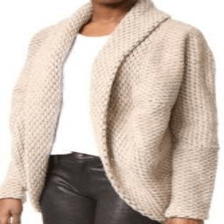} 
                  & 
                  \large \textcolor{green}{\checkmark}
                  \includegraphics[width=20mm,height=20mm]{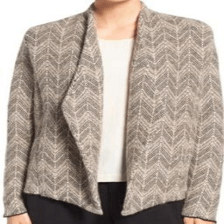}
                  \\
            \midrule
                  multi-task
                  \newline (BLIP$_{\text{FF}}$)
                  \newline \ding{55} inst
                  &  
                  \Large \textcolor{red}{\texttimes}
                  \newline \scriptsize Natural pebbled chevron alpaca drape front jacket.
                  &  
                  \Large \textcolor{red}{\texttimes}
                  \newline \scriptsize Beige textured terry draped jacket.
                  &  
                  \Large \textcolor{red}{\texttimes}
                  \includegraphics[width=20mm,height=20mm]{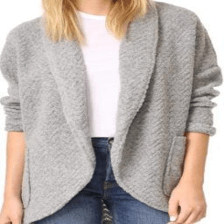}
                  & 
                  \Large \textcolor{red}{\texttimes}
                  \includegraphics[width=20mm,height=20mm]{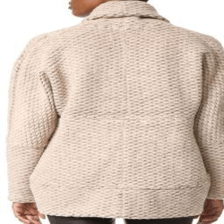} 
                  & 
                  \Large \textcolor{red}{\texttimes}
                  \newline \scriptsize Multicolor cashmere shawl collar jacket.
                  \\
            \midrule
                  Zero-shot
                  \newline (BLIP2)
                  \newline \ding{55} inst
                  &  
                  \Large \textcolor{red}{\texttimes}
                  \scriptsize Natural pebbled chevron alpaca drape front jacket.
                  &  
                  \Large \textcolor{red}{\texttimes}
                  \scriptsize Beige corded zigzag knit jacket.
                  &  
                  \Large \textcolor{red}{\texttimes}
                  \scriptsize Natural drape front open jacket.
                  & 
                  \Large \textcolor{red}{\texttimes}
                  \scriptsize Beige textured terry draped jacket.
                  & 
                  \Large \textcolor{red}{\texttimes}
                  \scriptsize Natural vegan sherpa drape jacket.
                  \\
            \bottomrule
    \end{tabular}
    \caption{Visualization of the Top 5 Candidates Retrieved from M-BEIR (5.6M candidates) by Different Models for a Fashion200K Query}
    \label{fig:retrieval_results_fashion200k}
    \end{center}
\end{figure*}

\begin{figure*}[t!]
\centering
\scriptsize
\begin{tabular}{>{\columncolor{lightblue}}m{1.5cm} >{\columncolor{lightblue}}m{1cm} m{2.2cm} m{3.8cm} m{2.4cm} m{3.8cm}}
    \toprule
    \textbf{Dataset} & \textbf{Domain} & \textbf{Task} & \textbf{Query Instruction  $q_{\texttt{inst}}$} & \textbf{Query Image $q_i$} & \textbf{Query Text $q_t$} \\
    \midrule
    EDIS 
    & News
    & 3. $q_t$ $\to$ ($c_i, c_t$)
    & Find a news image that matches the provided caption. 
    & -
    & Barack Obama with Germany's chancellor Angela Merkel at the Brandenburg Gate Berlin on 19 June.
\end{tabular}
\centering
    \scriptsize
    \begin{center}
    \begin{tabular}{>{\columncolor{lightblue}}m{1.2cm}m{2.7cm}m{2.7cm}m{2.7cm}m{2.7cm}m{2.7cm}}
            \toprule
            \textbf{Model} & \textbf{Rank 1 ($c_i, c_t$)} & \textbf{Rank 2 ($c_i, c_t$)} & \textbf{Rank 3 ($c_i, c_t$)} & \textbf{Rank 4 ($c_i, c_t$)} & \textbf{Rank 5 ($c_i, c_t$)} \\
            \midrule
                  UniIR
                  \newline (CLIP$_{\text{SF}}$)
                  \newline \checkmark inst
                  &  
                  \Large \textcolor{red}{\texttimes}
                  \includegraphics[width=20mm,height=20mm]{tables_and_figures/suppl/retrieval_results/CLIP_SF/Instruct/edis_1.png}
                  \newline \scriptsize Obama Speaks at a Berlin Event With Angela Merkel.
                  &  
                  \large \textcolor{green}{\checkmark}
                  \includegraphics[width=20mm,height=20mm]{tables_and_figures/suppl/retrieval_results/CLIP_SF/Instruct/edis_2.png} 
                  \newline \scriptsize Obama's Berlin visit to coincide with Trump in Brussels - Barack Obama.
                  &  
                  \Large \textcolor{red}{\texttimes}
                  \includegraphics[width=20mm,height=20mm]{tables_and_figures/suppl/retrieval_results/CLIP_SF/Instruct/edis_3.png} 
                  \newline \scriptsize Obama meets Germany's Merkel at chancellery in Berlin.
                  & 
                  \large \textcolor{green}{\checkmark}
                  \includegraphics[width=20mm,height=20mm]{tables_and_figures/suppl/retrieval_results/CLIP_SF/Instruct/edis_4.png} 
                  \newline \scriptsize President Obama Speaks to the People of Berlin from the Brandenburg Gate.
                  & 
                  \large \textcolor{green}{\checkmark}
                  \includegraphics[width=20mm,height=20mm]{tables_and_figures/suppl/retrieval_results/CLIP_SF/Instruct/edis_5.png} 
                  \newline \scriptsize When Barack Obama visited Berlin two years ago, he charmed a city.
                  \\
            \midrule
                  multi-task
                  \newline (CLIP$_{\text{SF}}$)
                  \newline \ding{55} inst
                  &  
                  \Large \textcolor{red}{\texttimes}
                  \scriptsize President Obama stands next to German Chancellor Angela Merkel in front of Brandenburg Gate in Berlin on June 19.
                  &  
                  \Large \textcolor{red}{\texttimes}
                  \scriptsize President Obama and German Chancellor Angela Merkel in 2011.
                  &  
                  \large \textcolor{green}{\checkmark}
                  \includegraphics[width=20mm,height=20mm]{tables_and_figures/suppl/retrieval_results/CLIP_SF/NoInstruct/edis_3.png}
                  \scriptsize When Barack Obama visited Berlin two years ago, he charmed a city.
                  & 
                  \Large \textcolor{red}{\texttimes}
                  \scriptsize  Barack Obama with German Chancellor Angela Merkel at the G20 summit in November.
                  & 
                  \Large \textcolor{red}{\texttimes}
                  \scriptsize US president Barack Obama at the Brandenburg Gate.
                  \\
            \midrule
                  UniIR
                  \newline (BLIP$_{\text{FF}}$)
                  \newline \checkmark inst
                  &  
                  \large \textcolor{green}{\checkmark}
                  \includegraphics[width=20mm,height=20mm]{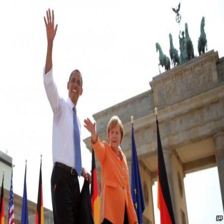}
                  \newline \scriptsize When Barack Obama visited Berlin two years ago, he charmed a city.
                  &  
                  \Large \textcolor{red}{\texttimes}
                  \includegraphics[width=20mm,height=20mm]{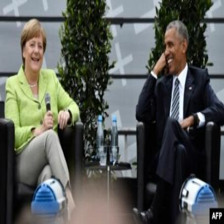} 
                  \newline \scriptsize Obama Gets Warm Welcome in Berlin, Praises Merkel.
                  &  
                  \Large \textcolor{red}{\texttimes}                  
                  \includegraphics[width=20mm,height=20mm]{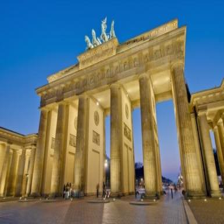} 
                  \newline \scriptsize Obama to give speech at Berlin's Brandenburg Gate.
                  & 
                  \Large \textcolor{red}{\texttimes}                  
                  \includegraphics[width=20mm,height=20mm]{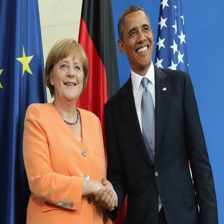} 
                  \newline \scriptsize As President Obama's team grapples with complaints from other countries about U.
                  & 
                  \large \textcolor{green}{\checkmark}
                  \includegraphics[width=20mm,height=20mm]{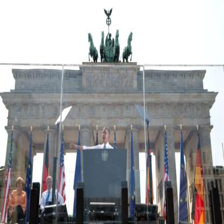}
                  \newline \scriptsize President Obama Speaks to the People of Berlin from the Brandenburg Gate.
                  \\
            \midrule
                  multi-task
                  \newline (BLIP$_{\text{FF}}$)
                  \newline \ding{55} inst
                  &  
                  \Large \textcolor{red}{\texttimes}
                  \scriptsize President Obama stands next to German Chancellor Angela Merkel in front of Brandenburg Gate in Berlin on June 19.
                  &  
                  \Large \textcolor{red}{\texttimes}
                  \scriptsize  President Obama and German Chancellor Angela Merkel in 2011.
                  &  
                  \large \textcolor{green}{\checkmark}
                  \includegraphics[width=20mm,height=20mm]{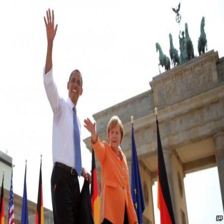}
                  \scriptsize When Barack Obama visited Berlin two years ago, he charmed a city.
                  & 
                  \Large \textcolor{red}{\texttimes}
                  \scriptsize Angela Merkel greets Barack Obama May 2012 Merkel Finally the Germans get to make love not war.
                  & 
                  \Large \textcolor{red}{\texttimes}
                  \tiny  US President Barack Obama waves next to German Chancellor Angela Merkel before they deliver speeches to invited guests in front of the Brandenburg Gate at Pariser Platz in Berlin on June 19 2013 during the official visit of the US President Barack Obama walks in John F Kennedy s footsteps on his first visit to Berlin as US president but encounter a more powerful...
                  \\
            \midrule
                  Zero-shot
                  \newline (BLIP2)
                  \newline \ding{55} inst
                  &  
                  \Large \textcolor{red}{\texttimes}
                  \tiny President Obama stands next to German Chancellor Angela Merkel in front of Brandenburg Gate in Berlin on June 19.
                  &  
                  \Large \textcolor{red}{\texttimes}
                  \tiny US President Barack Obama waves next to German Chancellor Angela Merkel before they deliver speeches to invited guests in front of the Brandenburg Gate at Pariser Platz in Berlin on June 19 2013 during the official visit of the US President Barack Obama walks in ...
                  &  
                  \Large \textcolor{red}{\texttimes}
                  \tiny Barack Obama with German chancellor Angela Merkel at the G20 summit in November.
                  & 
                  \Large \textcolor{red}{\texttimes}
                  \tiny BERLIN GERMANY JUNE 19 US President Barack Obama meets German Chancellor Angela Merkel for bilateral talks at the Chancellery on June 19 2013 in Berlin Germany Obama is visiting Berlin for the first time during his presidency and his speech at the Brandenburg Gate...
                  & 
                  \Large \textcolor{red}{\texttimes}
                  \tiny President Obama walks with Germany s Chancellor Angela Merkel in St Petersburg Russia on Sept 6.
                  \\
            \bottomrule
    \end{tabular}
    \caption{Visualization of the Top 5 Candidates Retrieved from M-BEIR (5.6M candidates) by Different Models for an EDIS Query}
    \label{fig:retrieval_results_edis}
    \end{center}
\end{figure*}

\begin{figure*}[t!]
\centering
\scriptsize
\begin{tabular}{>{\columncolor{lightgreen}}m{1.5cm} >{\columncolor{lightgreen}}m{1cm} m{2.2cm} m{3.8cm} m{2.4cm} m{3.8cm}}
    \toprule
    \textbf{Dataset} & \textbf{Domain} & \textbf{Task} & \textbf{Query Instruction  $q_{\texttt{inst}}$} & \textbf{Query Image $q_i$} & \textbf{Query Text $q_t$} \\
    \midrule
    WebQA 
    & Wikipedia
    & 3. $q_t$ $\to$ ($c_i, c_t$)
    & Find a Wikipedia image that answers this question.
    & -
    & What supports the statues standing on top of a structure at the burial site of Saint Peter?
\end{tabular}
\centering
    \scriptsize
    \begin{center}
    \begin{tabular}{>{\columncolor{lightgreen}}m{1.2cm}m{2.7cm}m{2.7cm}m{2.7cm}m{2.7cm}m{2.7cm}}
            \toprule
            \textbf{Model} & \textbf{Rank 1 ($c_i, c_t$)} & \textbf{Rank 2 ($c_i, c_t$)} & \textbf{Rank 3 ($c_i, c_t$)} & \textbf{Rank 4 ($c_i, c_t$)} & \textbf{Rank 5 ($c_i, c_t$)} \\
            \midrule
                  UniIR
                  \newline (CLIP$_{\text{SF}}$)
                  \newline \checkmark inst
                  &  
                  \Large \textcolor{red}{\texttimes}
                  \includegraphics[width=20mm,height=20mm]{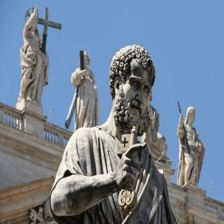}
                  \newline \scriptsize Statue of St. Peter in St. Peter's Square at the Vatican.
                  &  
                  \Large \textcolor{red}{\texttimes}
                  \includegraphics[width=20mm,height=20mm]{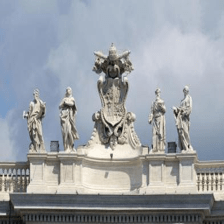} 
                  \newline \scriptsize Symbol and statues in St. Peter symbol and statues in St. Peter.
                  &  
                  \Large \textcolor{red}{\texttimes}
                  \includegraphics[width=20mm,height=20mm]{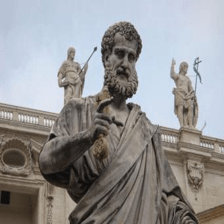} 
                  \newline \scriptsize St-peter St. Peter.
                  & 
                  \Large \textcolor{red}{\texttimes}
                  \includegraphics[width=20mm,height=20mm]{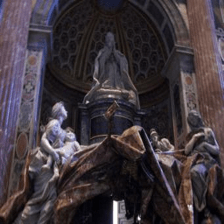} 
                  \newline \scriptsize Statue in St. Peter's Basilica.
                  & 
                  \Large \textcolor{red}{\texttimes}
                  \includegraphics[width=20mm,height=20mm]{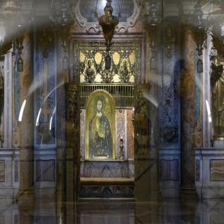} 
                  \newline \scriptsize The floor above Saint Peter's tomb.
                  \\
            \midrule
                  multi-task
                  \newline (CLIP$_{\text{SF}}$)
                  \newline \ding{55} inst
                  &  
                  \Large \textcolor{red}{\texttimes}
                  \newline \scriptsize Saint Peter's tomb is supposedly directly below the high altar of the basilica. For this reason, many popes have been interred at St. Peter's since the Early Christian period.
                  &  
                  \Large \textcolor{red}{\texttimes}
                  \newline \tiny Since the crucifixion and burial of Saint Peter in 64 AD, the spot was thought to be the location of the tomb of Saint Peter, where there stood a small shrine. With its increasing prestige the church became richly decorated with statues, furnishings and elaborate chandeliers, and side tombs and altars were continuously added.
                  &  
                  \Large \textcolor{red}{\texttimes}
                  \newline \scriptsize Saint Peter's tomb is a site under St. Peter's Basilica that includes several graves and a structure said by Vatican authorities to have been built to memorialize the location of Saint Peter's grave.
                  & 
                  \Large \textcolor{red}{\texttimes}
                  \newline \tiny Catholic tradition holds that the basilica is the burial site of Saint Peter, chief among Jesus 's apostles and also the first Bishop of Rome (Pope). Saint Peter's tomb is supposedly directly below the high altar of the basilica. For this reason, many popes have been interred at St. Peter's since the Early Christian period.
                  & 
                  \Large \textcolor{red}{\texttimes}
                  \includegraphics[width=20mm,height=20mm]{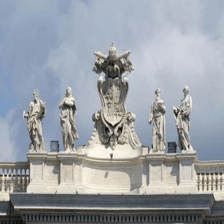}  
                  \newline \scriptsize Symbol and statues in St. Peter symbol and statues in St. Peter.
                  \\
            \midrule
                  UniIR
                  \newline (BLIP$_{\text{FF}}$)
                  \newline \checkmark inst
                  &  
                  \Large \textcolor{red}{\texttimes}
                  \includegraphics[width=20mm,height=20mm]{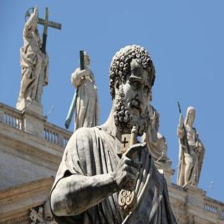}
                  \newline \scriptsize Statue of St. Peter in St. Peter's Square at the Vatican.
                  &  
                  \Large \textcolor{red}{\texttimes}
                  \includegraphics[width=20mm,height=20mm]{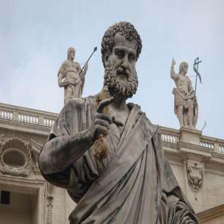} 
                  \newline \scriptsize St-peter St. Peter.
                  &  
                  \large \textcolor{green}{\checkmark}
                  \includegraphics[width=20mm,height=20mm]{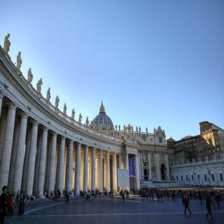} 
                  \newline \scriptsize Saint Peter's Basilica, Vatican City, Rome Burial site of Saint Peter, the "First Pope"!
                  & 
                  \Large \textcolor{red}{\texttimes}                  
                  \includegraphics[width=20mm,height=20mm]{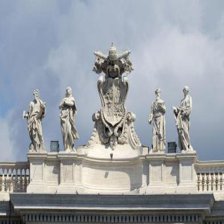} 
                  \newline \scriptsize Symbol and statues in St. Peter symbol and statues in St. Peter.
                  & 
                  \Large \textcolor{red}{\texttimes}   
                  \scriptsize St. Peter's Basilica, believed to be the burial site of St. Peter, seen from the River Tiber.
                  \\
            \midrule
                  multi-task
                  \newline (BLIP$_{\text{FF}}$)
                  \newline \ding{55} inst
                  &  
                  \Large \textcolor{red}{\texttimes}
                  \includegraphics[width=20mm,height=20mm]{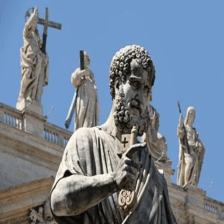}
                  \newline \scriptsize Statue of St. Peter in St. Peter's Square at the Vatican.
                  &  
                  \Large \textcolor{red}{\texttimes}
                  \includegraphics[width=20mm,height=20mm]{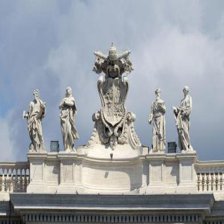} 
                  \newline \scriptsize Symbol and statues in St. Peter symbol and statues in St. Peter.
                  &  
                  \Large \textcolor{red}{\texttimes}                  
                  \includegraphics[width=20mm,height=20mm]{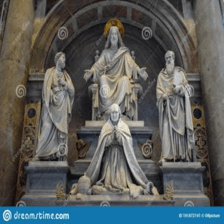} 
                  \newline \scriptsize A Sculpture Featuring Jesus, Saint Paul, Saint Peter and a Pope, St. Peter Basilica, Vatican, Italy Editorial Photo.
                  & 
                  \Large \textcolor{red}{\texttimes}                  
                  \includegraphics[width=20mm,height=20mm]{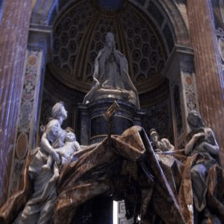} 
                  \newline \scriptsize Statue in St. Peter's Basilica.
                  & 
                  \Large \textcolor{red}{\texttimes}   
                  \scriptsize Saint Peter's tomb is a site under St. Peter's Basilica that includes several graves and a structure said by Vatican authorities to have been built to memorialize the location of Saint Peter's grave.
                  \\
            \midrule
                  Zero-shot
                  \newline (BLIP2)
                  \newline \ding{55} inst
                  &  
                  \Large \textcolor{red}{\texttimes}
                  \tiny Also a bronze statue of the apostle St. Paul was placed on the top platform, to go with that of St. Peter on Trajan's Column (27 October 1588). (Originally the top platform probably had a statue of Marcus Aurelius, but it had been already lost by the 16th century.).
                  &  
                  \Large \textcolor{red}{\texttimes}
                  \tiny Pope Sixtus V was determined to erect the obelisk in front of St Peter's, of which the nave was yet to be built.
                  &  
                  \Large \textcolor{red}{\texttimes}
                  \tiny Along the roofline, by contrast, there is a line of statues, many in their own small pavilions, culminating in Saint Mark flanked by six angels in the centre, above a large gilded winged lion (his symbol, and that of Venice).
                  & 
                  \Large \textcolor{red}{\texttimes}
                  \tiny Saint Peter's tomb is a site under St. Peter's Basilica that includes several graves and a structure said by Vatican authorities to have been built to memorialize the location of Saint Peter's grave. St. Peter's tomb is near the west end of a...
                  & 
                  \Large \textcolor{red}{\texttimes}
                  \tiny Saint Peter's tomb is a site under St. Peter's Basilica that includes several graves and a structure said by Vatican authorities to have been built to memorialize the location of Saint Peter's grave.
                  \\
            \bottomrule
    \end{tabular}
    \caption{Visualization of the Top 5 Candidates Retrieved from M-BEIR (5.6M candidates) by Different Models for a WebQA Query}
    \label{fig:retrieval_results_webqa}
    \end{center}
\end{figure*}

\begin{figure*}[t!]
\centering
\scriptsize
\begin{tabular}{>{\columncolor{lightpurple}}m{1.5cm} >{\columncolor{lightpurple}}m{1cm} m{2.2cm} m{3.8cm} m{2.4cm} m{3.8cm}}
    \toprule
    \textbf{Dataset} & \textbf{Domain} & \textbf{Task} & \textbf{Query Instruction  $q_{\texttt{inst}}$} & \textbf{Query Image $q_i$} & \textbf{Query Text $q_t$} \\
    \midrule
    NIGHTS 
    & Common
    & 5. $q_t$ $\to$  $c_t$
    & Which everyday image is the most similar to the reference image?
    & \includegraphics[width=20mm,height=20mm]{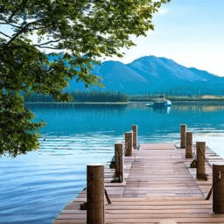}
    & -
\end{tabular}
\centering
    \scriptsize
    \begin{center}
    \begin{tabular}{>{\columncolor{lightpurple}}m{1.2cm}m{2.7cm}m{2.7cm}m{2.7cm}m{2.7cm}m{2.7cm}}
            \toprule
            \textbf{Model} & \textbf{Rank 1 ($c_t$)} & \textbf{Rank 2 ($c_t$)} & \textbf{Rank 3 ($c_t$)} & \textbf{Rank 4 ($c_t$)} & \textbf{Rank 5 ($c_t$)} \\
            \midrule
                  UniIR
                  \newline (CLIP$_{\text{SF}}$)
                  \newline \checkmark inst
                  &  
                  \Large \textcolor{red}{\texttimes}
                  \includegraphics[width=20mm,height=20mm]{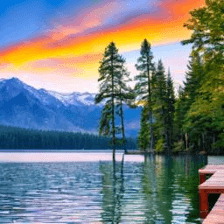}
                  &  
                  \Large \textcolor{red}{\texttimes}
                  \includegraphics[width=20mm,height=20mm]{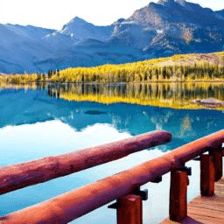} 
                  &  
                  \Large \textcolor{red}{\texttimes}
                  \includegraphics[width=20mm,height=20mm]{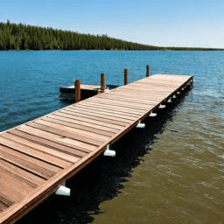} 
                  & 
                  \Large \textcolor{red}{\texttimes}
                  \includegraphics[width=20mm,height=20mm]{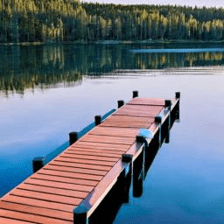} 
                  & 
                  \large \textcolor{green}{\checkmark}
                  \includegraphics[width=20mm,height=20mm]{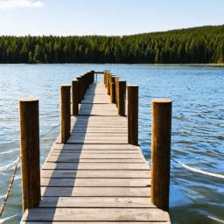} 
                  \\
            \midrule
                  multi-task
                  \newline (CLIP$_{\text{SF}}$)
                  \newline \ding{55} inst
                  &  
                  \large \textcolor{green}{\checkmark}
                  \includegraphics[width=20mm,height=20mm]{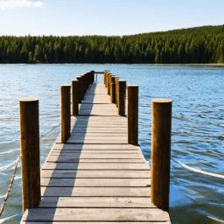}  
                  &  
                  \Large \textcolor{red}{\texttimes}
                  \includegraphics[width=20mm,height=20mm]{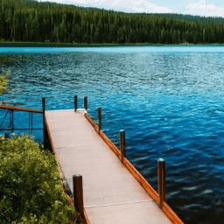} 
                  &  
                  \Large \textcolor{red}{\texttimes}
                  \includegraphics[width=20mm,height=20mm]{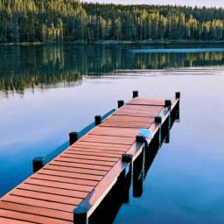}  
                  & 
                  \Large \textcolor{red}{\texttimes}
                  \includegraphics[width=20mm,height=20mm]{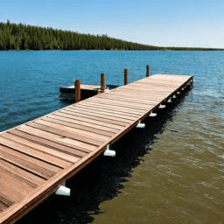} 
                  & 
                  \Large \textcolor{red}{\texttimes}
                  \includegraphics[width=20mm,height=20mm]{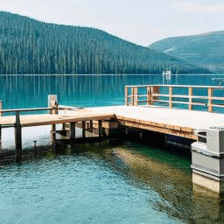}  
                  \\
            \midrule
                  UniIR
                  \newline (BLIP$_{\text{FF}}$)
                  \newline \checkmark inst
                  &  
                  \Large \textcolor{red}{\texttimes}
                  \includegraphics[width=20mm,height=20mm]{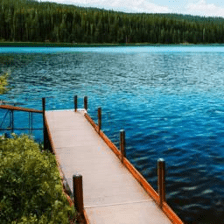}
                  &  
                  \Large \textcolor{red}{\texttimes}
                  \includegraphics[width=20mm,height=20mm]{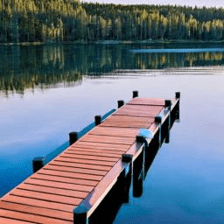} 
                  &  
                  \large \textcolor{green}{\checkmark}
                  \includegraphics[width=20mm,height=20mm]{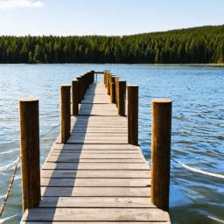} 
                  & 
                  \Large \textcolor{red}{\texttimes}                  
                  \includegraphics[width=20mm,height=20mm]{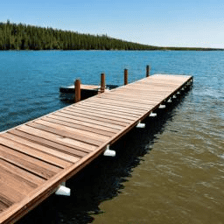} 
                  & 
                  \Large \textcolor{red}{\texttimes}   
                  \includegraphics[width=20mm,height=20mm]{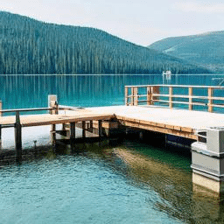}
                  \\
            \midrule
                  multi-task
                  \newline (BLIP$_{\text{FF}}$)
                  \newline \ding{55} inst
                  &  
                  \Large \textcolor{red}{\texttimes}
                  \includegraphics[width=20mm,height=20mm]{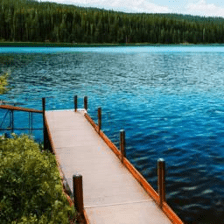}  
                  &  
                  \Large \textcolor{red}{\texttimes}
                  \includegraphics[width=20mm,height=20mm]{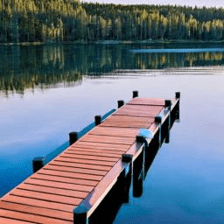} 
                  &  
                  \large \textcolor{green}{\checkmark}
                  \includegraphics[width=20mm,height=20mm]{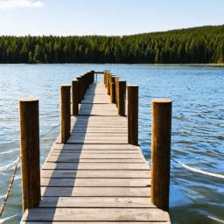} 
                  & 
                  \Large \textcolor{red}{\texttimes}
                  \includegraphics[width=20mm,height=20mm]{tables_and_figures/suppl/retrieval_results/CLIP_SF/NoInstruct/nights_4.png} 
                  & 
                  \Large \textcolor{red}{\texttimes}
                  \includegraphics[width=20mm,height=20mm]{tables_and_figures/suppl/retrieval_results/CLIP_SF/NoInstruct/nights_5.png}  
                  \\
            \midrule
                  Zero-shot
                  \newline (BLIP2)
                  \newline \ding{55} inst
                  &  
                  \Large \textcolor{red}{\texttimes}
                  \includegraphics[width=20mm,height=20mm]{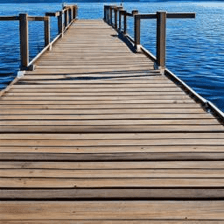}
                  &  
                  \Large \textcolor{red}{\texttimes}
                  \includegraphics[width=20mm,height=20mm]{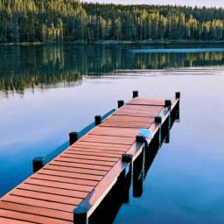}
                  &  
                  \large \textcolor{green}{\checkmark}
                  \includegraphics[width=20mm,height=20mm]{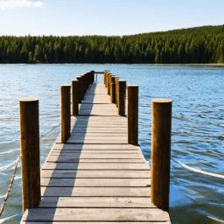}
                  & 
                  \Large \textcolor{red}{\texttimes}
                  \includegraphics[width=20mm,height=20mm]{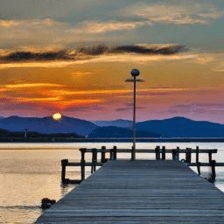} 
                  & 
                  \Large \textcolor{red}{\texttimes}
                  \includegraphics[width=20mm,height=20mm]{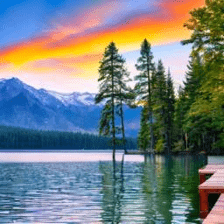}
                  \\
            \bottomrule
    \end{tabular}
    \caption{Visualization of the Top 5 Candidates Retrieved from M-BEIR (5.6M candidates) by Different Models for a NIGHTS Query}
    \label{fig:retrieval_results_nights}
    \end{center}
\end{figure*}

\begin{figure*}[t!]
\centering
\scriptsize
\begin{tabular}{>{\columncolor{lightred}}m{1.5cm} >{\columncolor{lightred}}m{1cm} m{2.2cm} m{3.8cm} m{2.4cm} m{3.8cm}}
    \toprule
    \textbf{Dataset} & \textbf{Domain} & \textbf{Task} & \textbf{Query Instruction  $q_{\texttt{inst}}$} & \textbf{Query Image $q_i$} & \textbf{Query Text $q_t$} \\
    \midrule
    FashionIQ 
    & Fashion
    & 7. ($q_i, q_t$) $\to$ $c_i$
    & Find a fashion image that aligns with the reference image and style note.
    & \includegraphics[width=20mm,height=20mm]{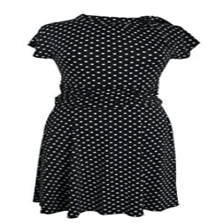}
    & Is white with dots and a black belt.
\end{tabular}
\centering
    \scriptsize
    \begin{center}
    \begin{tabular}{>{\columncolor{lightred}}m{1.2cm}m{2.7cm}m{2.7cm}m{2.7cm}m{2.7cm}m{2.7cm}}
            \toprule
            \textbf{Model} & \textbf{Rank 1 $c_t$} & \textbf{Rank 2 $c_t$} & \textbf{Rank 3 $c_t$} & \textbf{Rank 4 $c_t$} & \textbf{Rank 5 $c_t$} \\
            \midrule
                  UniIR
                  \newline (CLIP$_{\text{SF}}$)
                  \newline \checkmark inst
                  &  
                  \Large \textcolor{red}{\texttimes}
                  \includegraphics[width=20mm,height=20mm]{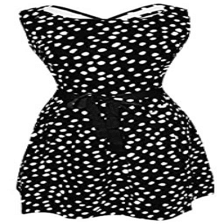}
                  &  
                  \Large \textcolor{red}{\texttimes}
                  \includegraphics[width=20mm,height=20mm]{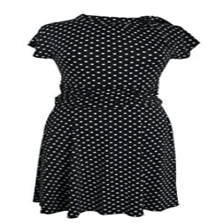} 
                  &  
                  \Large \textcolor{red}{\texttimes}
                  \includegraphics[width=20mm,height=20mm]{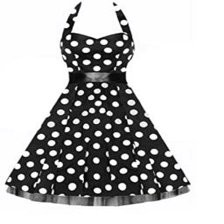} 
                  & 
                  \Large \textcolor{red}{\texttimes}
                  \includegraphics[width=20mm,height=20mm]{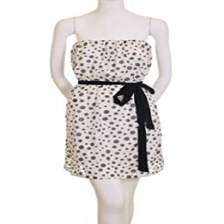} 
                  & 
                  \Large \textcolor{red}{\texttimes}
                  \includegraphics[width=20mm,height=20mm]{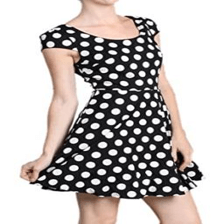} 
                  \\
            \midrule
                  multi-task
                  \newline (CLIP$_{\text{SF}}$)
                  \newline \ding{55} inst
                  &  
                  \Large \textcolor{red}{\texttimes}
                  \includegraphics[width=20mm,height=20mm]{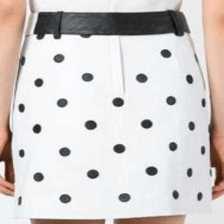}  
                  &  
                  \Large \textcolor{red}{\texttimes}
                  \includegraphics[width=20mm,height=20mm]{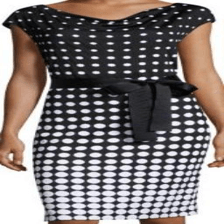} 
                  &  
                  \Large \textcolor{red}{\texttimes}
                  \includegraphics[width=20mm,height=20mm]{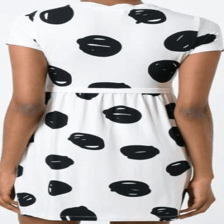}  
                  & 
                  \Large \textcolor{red}{\texttimes}
                  \includegraphics[width=20mm,height=20mm]{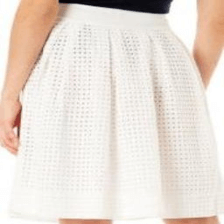} 
                  & 
                  \Large \textcolor{red}{\texttimes}
                  \includegraphics[width=20mm,height=20mm]{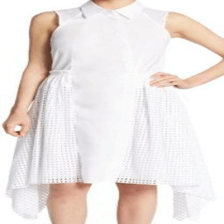}  
                  \\
            \midrule
                  UniIR
                  \newline (BLIP$_{\text{FF}}$)
                  \newline \checkmark inst
                  &  
                  \Large \textcolor{red}{\texttimes}
                  \includegraphics[width=20mm,height=20mm]{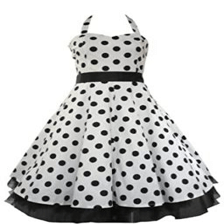}
                  &  
                  \Large \textcolor{red}{\texttimes}
                  \includegraphics[width=20mm,height=20mm]{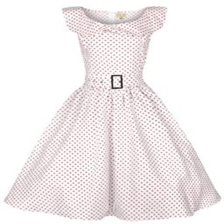} 
                  &  
                  \large \textcolor{green}{\checkmark}
                  \includegraphics[width=20mm,height=20mm]{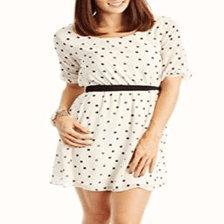} 
                  & 
                  \Large \textcolor{red}{\texttimes}                  
                  \includegraphics[width=20mm,height=20mm]{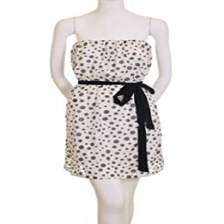} 
                  & 
                  \Large \textcolor{red}{\texttimes}   
                  \includegraphics[width=20mm,height=20mm]{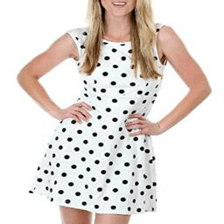}
                  \\
            \midrule
                  multi-task
                  \newline (BLIP$_{\text{FF}}$)
                  \newline \ding{55} inst
                  &  
                  \Large \textcolor{red}{\texttimes}
                  \includegraphics[width=20mm,height=20mm]{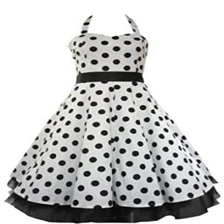}  
                  &  
                  \Large \textcolor{red}{\texttimes}
                  \includegraphics[width=20mm,height=20mm]{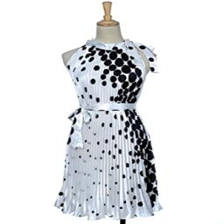} 
                  &  
                  \large \textcolor{green}{\checkmark}
                  \includegraphics[width=20mm,height=20mm]{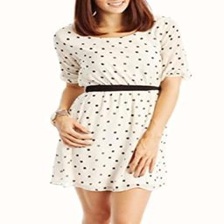} 
                  & 
                  \Large \textcolor{red}{\texttimes}
                  \includegraphics[width=20mm,height=20mm]{tables_and_figures/suppl/retrieval_results/CLIP_SF/NoInstruct/fashioniq_4.png} 
                  & 
                  \Large \textcolor{red}{\texttimes}
                  \includegraphics[width=20mm,height=20mm]{tables_and_figures/suppl/retrieval_results/CLIP_SF/NoInstruct/fashioniq_5.png}  
                  \\
            \midrule
                  Zero-shot
                  \newline (BLIP2)
                  \newline \ding{55} inst
                  &  
                  \Large \textcolor{red}{\texttimes}
                  \includegraphics[width=20mm,height=20mm]{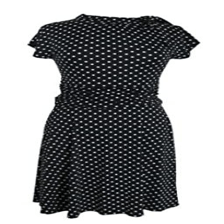}
                  &  
                  \Large \textcolor{red}{\texttimes}
                  \includegraphics[width=20mm,height=20mm]{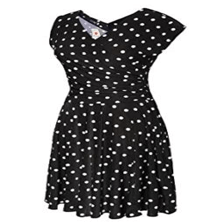}
                  &  
                  \Large \textcolor{red}{\texttimes}
                  \includegraphics[width=20mm,height=20mm]{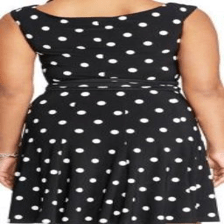}
                  & 
                  \Large \textcolor{red}{\texttimes}
                  \includegraphics[width=20mm,height=20mm]{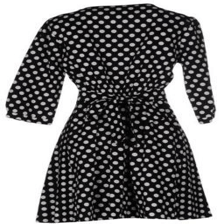} 
                  & 
                  \Large \textcolor{red}{\texttimes}
                  \includegraphics[width=20mm,height=20mm]{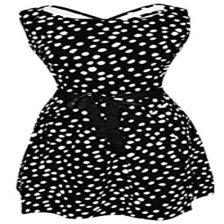}
                  \\
            \bottomrule
    \end{tabular}
    \caption{Visualization of the Top 5 Candidates Retrieved from M-BEIR (5.6M candidates) by Different Models for a FashionIQ Query}
    \label{fig:retrieval_results_fashioniq}
    \end{center}
\end{figure*}

\begin{figure*}[t!]
\centering
\scriptsize
\begin{tabular}{>{\columncolor{lightpurple}}m{1.5cm} >{\columncolor{lightpurple}}m{1cm} m{2.2cm} m{3.8cm} m{2.4cm} m{3.8cm}}
    \toprule
    \textbf{Dataset} & \textbf{Domain} & \textbf{Task} & \textbf{Query Instruction  $q_{\texttt{inst}}$} & \textbf{Query Image $q_i$} & \textbf{Query Text $q_t$} \\
    \midrule
    CIRR 
    & Common
    & 7. ($q_i, q_t$) $\to$ $c_i$
    & I'm looking for a similar everyday image with the described changes.
    & \includegraphics[width=20mm,height=20mm]{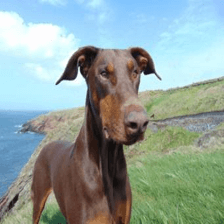}
    & The target photo is of a black and brown dog laying in the grass with a small red ball at his feet.
\end{tabular}
\centering
    \scriptsize
    \begin{center}
    \begin{tabular}{>{\columncolor{lightpurple}}m{1.2cm}m{2.7cm}m{2.7cm}m{2.7cm}m{2.7cm}m{2.7cm}}
            \toprule
            \textbf{Model} & \textbf{Rank 1 $c_t$} & \textbf{Rank 2 $c_t$} & \textbf{Rank 3 $c_t$} & \textbf{Rank 4 $c_t$} & \textbf{Rank 5 $c_t$} \\
            \midrule
                  UniIR
                  \newline (CLIP$_{\text{SF}}$)
                  \newline \checkmark inst
                  &  
                  \Large \textcolor{red}{\texttimes}
                  \includegraphics[width=20mm,height=20mm]{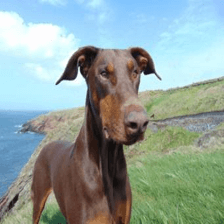}
                  &  
                  \Large \textcolor{red}{\texttimes}
                  \includegraphics[width=20mm,height=20mm]{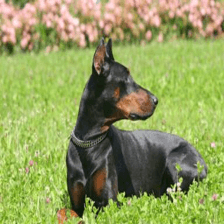} 
                  &  
                  \Large \textcolor{red}{\texttimes}
                  \includegraphics[width=20mm,height=20mm]{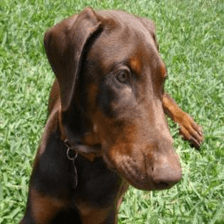} 
                  & 
                  \Large \textcolor{red}{\texttimes}
                  \includegraphics[width=20mm,height=20mm]{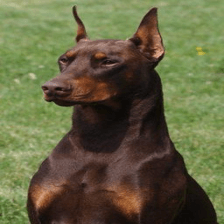} 
                  & 
                  \Large \textcolor{red}{\texttimes}
                  \includegraphics[width=20mm,height=20mm]{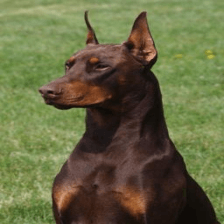} 
                  \\
            \midrule
                  multi-task
                  \newline (CLIP$_{\text{SF}}$)
                  \newline \ding{55} inst
                  &  
                  \Large \textcolor{red}{\texttimes}
                  \includegraphics[width=20mm,height=20mm]{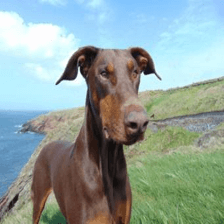}  
                  &  
                  \Large \textcolor{red}{\texttimes}
                  \includegraphics[width=20mm,height=20mm]{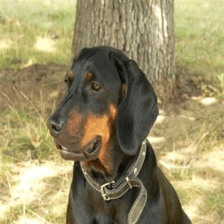} 
                  &  
                  \Large \textcolor{red}{\texttimes}
                  \includegraphics[width=20mm,height=20mm]{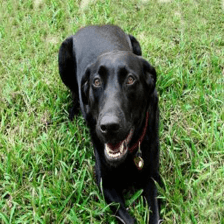}  
                  & 
                  \Large \textcolor{red}{\texttimes}
                  \includegraphics[width=20mm,height=20mm]{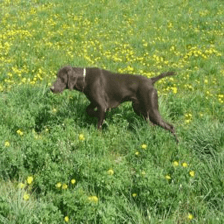} 
                  & 
                  \Large \textcolor{red}{\texttimes}
                  \includegraphics[width=20mm,height=20mm]{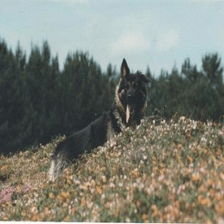}  
                  \\
            \midrule
                  UniIR
                  \newline (BLIP$_{\text{FF}}$)
                  \newline \checkmark inst
                  &  
                  \large \textcolor{green}{\checkmark}
                  \includegraphics[width=20mm,height=20mm]{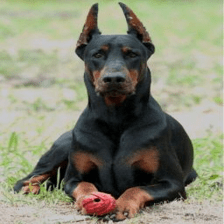}
                  &  
                  \Large \textcolor{red}{\texttimes}
                  \includegraphics[width=20mm,height=20mm]{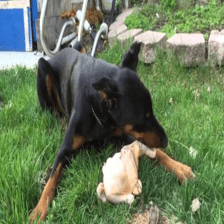} 
                  &  
                  \Large \textcolor{red}{\texttimes}                  
                  \includegraphics[width=20mm,height=20mm]{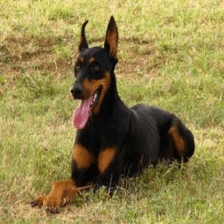} 
                  & 
                  \Large \textcolor{red}{\texttimes}                  
                  \includegraphics[width=20mm,height=20mm]{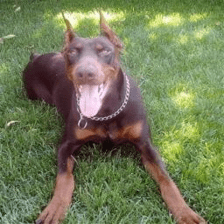} 
                  & 
                  \Large \textcolor{red}{\texttimes}   
                  \includegraphics[width=20mm,height=20mm]{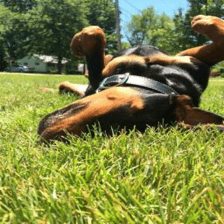}
                  \\
            \midrule
                  multi-task
                  \newline (BLIP$_{\text{FF}}$)
                  \newline \ding{55} inst
                  &  
                  \large \textcolor{green}{\checkmark}
                  \includegraphics[width=20mm,height=20mm]{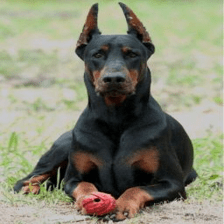}  
                  &  
                  \Large \textcolor{red}{\texttimes}
                  \includegraphics[width=20mm,height=20mm]{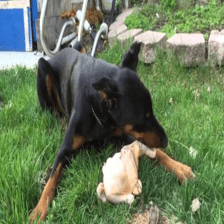} 
                  &  
                  \Large \textcolor{red}{\texttimes}
                  \includegraphics[width=20mm,height=20mm]{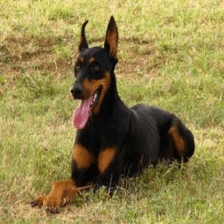} 
                  & 
                  \Large \textcolor{red}{\texttimes}
                  \includegraphics[width=20mm,height=20mm]{tables_and_figures/suppl/retrieval_results/CLIP_SF/NoInstruct/cirr_4.png} 
                  & 
                  \Large \textcolor{red}{\texttimes}
                  \includegraphics[width=20mm,height=20mm]{tables_and_figures/suppl/retrieval_results/CLIP_SF/NoInstruct/cirr_5.png}  
                  \\
            \midrule
                  Zero-shot
                  \newline (BLIP2)
                  \newline \ding{55} inst
                  &  
                  \Large \textcolor{red}{\texttimes}
                  \includegraphics[width=20mm,height=20mm]{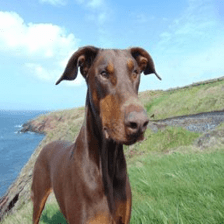}
                  &  
                  \Large \textcolor{red}{\texttimes}
                  \includegraphics[width=20mm,height=20mm]{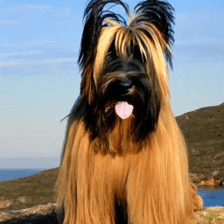}
                  &  
                  \Large \textcolor{red}{\texttimes}
                  \includegraphics[width=20mm,height=20mm]{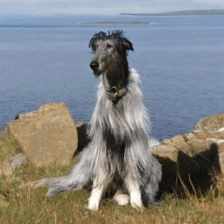}
                  & 
                  \Large \textcolor{red}{\texttimes}
                  \includegraphics[width=20mm,height=20mm]{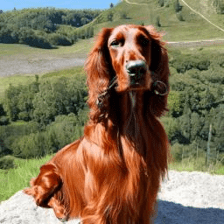} 
                  & 
                  \Large \textcolor{red}{\texttimes}
                  \includegraphics[width=20mm,height=20mm]{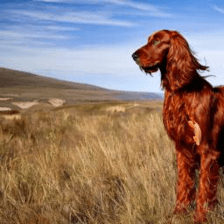}
                  \\
            \bottomrule
    \end{tabular}
    \caption{Visualization of the Top 5 Candidates Retrieved from M-BEIR (5.6M candidates) by Different Models for a CIRR Query}
    \label{fig:retrieval_results_cirr}
    \end{center}
\end{figure*}

\begin{figure*}[t!]
\centering
\scriptsize
\begin{tabular}{>{\columncolor{lightgreen}}m{1.5cm} >{\columncolor{lightgreen}}m{1cm} m{2.2cm} m{3.8cm} m{2.4cm} m{3.8cm}}
    \toprule
    \textbf{Dataset} & \textbf{Domain} & \textbf{Task} & \textbf{Query Instruction  $q_{\texttt{inst}}$} & \textbf{Query Image $q_i$} & \textbf{Query Text $q_t$} \\
    \midrule
    OVEN 
    & Wikipedia
    & 8. ($q_i, q_t$) $\to$ ($c_i, q_t$)
    & Determine the Wikipedia image-snippet pair that clarifies the entity in this picture.
    & \includegraphics[width=20mm,height=20mm]{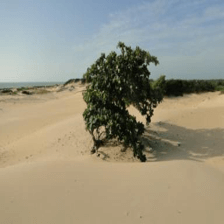}
    & What is the name of this place?
\end{tabular}
\centering
    \scriptsize
    \begin{center}
    \begin{tabular}{>{\columncolor{lightgreen}}m{1.2cm}m{2.7cm}m{2.7cm}m{2.7cm}m{2.7cm}m{2.7cm}}
            \toprule
            \textbf{Model} & \textbf{Rank 1 ($c_i, c_t$)} & \textbf{Rank 2 ($c_i, c_t$)} & \textbf{Rank 3 ($c_i, c_t$)} & \textbf{Rank 4 ($c_i, c_t$)} & \textbf{Rank 5 ($c_i, c_t$)} \\
            \midrule
                  UniIR
                  \newline (CLIP$_{\text{SF}}$)
                  \newline \checkmark inst
                  &  
                  \Large \textcolor{red}{\texttimes}
                  \includegraphics[width=20mm,height=20mm]{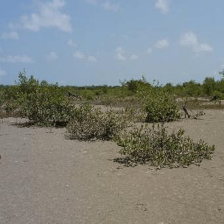}
                  \newline \tiny  Pará mangroves. other than the city of Belém and its surroundings. The mangroves are used by artisanal fishermen as a source of crabs and wood.Threats include subsistence agriculture and livestock raising, tourism, logging, mining and urban development.
                  &  
                  \Large \textcolor{red}{\texttimes}
                  \includegraphics[width=20mm,height=20mm]{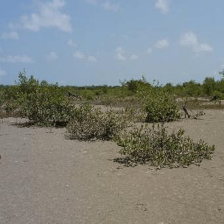} 
                  \newline \tiny Pará mangroves. The Pará mangroves (NT1427) is an ecoregion along the Atlantic coast of the state of Pará in Brazil.They constitute the western extension of the Maranhão mangroves ecoregion.The mangroves are relatively intact, although they are under...
                  &  
                  \large \textcolor{green}{\checkmark}
                  \includegraphics[width=20mm,height=20mm]{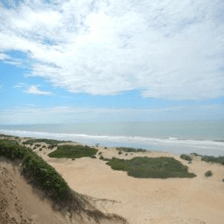} 
                  \newline \tiny Itaúnas State Park. by squatters. Activities. As of 2015 the opening hours were from 8:30 to 17:30.The park is suitable for ecotourism, hiking and swimming in the river and the ocean.Visitors may take guided horseback tours, or may explore by kayak, canoe, jeep...
                  & 
                  \large \textcolor{green}{\checkmark}
                  \includegraphics[width=20mm,height=20mm]{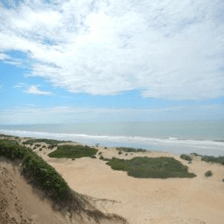} 
                  \newline \tiny Itaúnas State Park. archaeological sites with traces of human settlements such as chipped stones, indigenous pottery and artifacts from the colonial era.Vegetation includes coastal forest, with fragments of forest endangered in Espírito Santo...
                  & 
                  \Large \textcolor{red}{\texttimes}
                  \includegraphics[width=20mm,height=20mm]{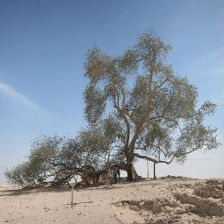} 
                  \newline \tiny Tree of Life (Bahrain). the list.In October 2010, archaeologists unearthed 500-year-old pottery and other artefacts in the vicinity of the tree. A soil and dendrochronology analysis conducted in the 1990s concluded that the tree was an "Acacia"...
                  \\
            \midrule
                  multi-task
                  \newline (CLIP$_{\text{SF}}$)
                  \newline \ding{55} inst
                  &  
                  \Large \textcolor{red}{\texttimes}
                  \includegraphics[width=20mm,height=20mm]{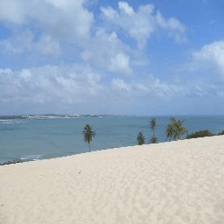}  
                  \newline \tiny Natal Dunes State Park. The Natal Dunes State Park "Journalist Luiz Maria Alves" (), or simply the Dunes Park () is a state park in the state of Rio Grande do Norte in the Northeast Region of Brazil. It protects an area of dunes and native ...
                  &  
                  \Large \textcolor{red}{\texttimes}
                  \includegraphics[width=20mm,height=20mm]{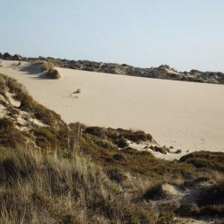} 
                  \newline \tiny Cresmina Dune. Cresmina Dune from the north.
                  &  
                  \Large \textcolor{red}{\texttimes}
                  \includegraphics[width=20mm,height=20mm]{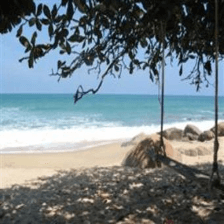}   
                  \newline \tiny Kijal. Kijal (est. pop. (2000 census): 4,375) is a mukim in Kemaman District, Terengganu, Malaysia. The town is well known for the Awana Kijal Golf and Beach Resort, a luxury resort which is owned by the...
                  & 
                  \Large \textcolor{red}{\texttimes}
                  \tiny Maspalomas Dunes. The Maspalomas Dunes () are sand dunes located on the south coast of the island of Gran Canaria, Province of Las Palmas, in the Canary Islands. A 404 (ha) area of the municipality of San Bartolomé de Tirajana, they have been protected...
                  & 
                  \Large \textcolor{red}{\texttimes}
                  \tiny Maspalomas Dunes. They were one of the contenders in the 12 Treasures of Spain competition.They were formed by sand from the now subdued marine shelf, when it was laid dry during the last ice age and the wind blew the sand towards the coast of the island...
                  \\
            \midrule
                  UniIR
                  \newline (BLIP$_{\text{FF}}$)
                  \newline \checkmark inst
                  &  
                  \Large \textcolor{red}{\texttimes}
                  \includegraphics[width=20mm,height=20mm]{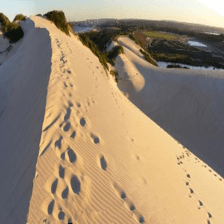}
                  \newline \tiny Cronulla sand dunes. context of its setting, intactness, aesthetic qualities and social significance is held in high community esteem. The water bodies formed by sand removal surrounding the Cronulla Sand Dune have been identified ...
                  &  
                  \Large \textcolor{red}{\texttimes}
                  \includegraphics[width=20mm,height=20mm]{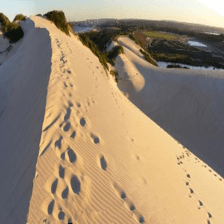} 
                  \newline \tiny Cronulla sand dunes. referred to as the Kurnell sand dune is estimated to be about 15,000 years old. It was formed when the sea reached its present level and began to stabilise, between 9000 and 6000 ...
                  &  
                  \Large \textcolor{red}{\texttimes}
                  \includegraphics[width=20mm,height=20mm]{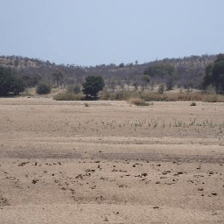} 
                  \newline \tiny Tel Tanninim. Tel Tanninim (), in Arabic Tell al-MilÄt (lit. 'Mortar Mound') is an ancient tell (archaeological mound) on the shore of the Mediterranean, near the mouth of Nahal Tanninim ('Crocodiles Stream') ...
                  & 
                  \Large \textcolor{red}{\texttimes}
                  \includegraphics[width=20mm,height=20mm]{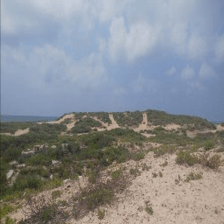} 
                  \newline \tiny Thuli Parks and Wildlife Land. black rhinoceros, lion, Cape wild dog, Namibian cheetah and African leopard, many of which move freely between the protected area and neighbouring Botswana and South Africa ...
                  & 
                  \large \textcolor{green}{\checkmark}
                  \includegraphics[width=20mm,height=20mm]{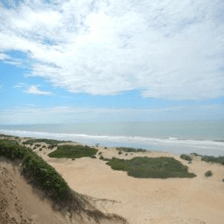} 
                  \newline \tiny Itaúnas State Park. archaeological sites with traces of human settlements such as chipped stones, indigenous pottery and artifacts from the colonial era. Vegetation includes coastal forest, with fragments of forest ...
                  \\
            \midrule
                  multi-task
                  \newline (BLIP$_{\text{FF}}$)
                  \newline \ding{55} inst
                  &  
                  \Large \textcolor{red}{\texttimes}
                  \includegraphics[width=20mm,height=20mm]{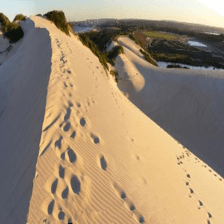}  
                  \newline \tiny Cronulla sand dunes. in New South Wales for social, cultural or spiritual reasons.The site has historic and cultural significance for the Aboriginal community. The Cronulla Sand Dune is of cultural heritage and spiritual significance to the La Perouse ...
                  &  
                  \Large \textcolor{red}{\texttimes}
                  \includegraphics[width=20mm,height=20mm]{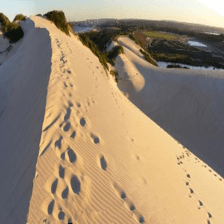} 
                  \newline \tiny Cronulla sand dunes. men in the colony. During the 1860s Holt consolidated his landholdings on the Peninsular (which included Captain Cook's landing place). His accumulated holdings represented 5261 (ha) of what is the present day...
                  &  
                  \Large \textcolor{red}{\texttimes}
                  \includegraphics[width=20mm,height=20mm]{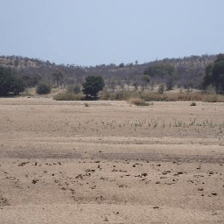} 
                  \newline \tiny Samalayuca Dune Fields. areas of El Paso del Norte in the north and the village of Carrizal in the south. Experiences of travelers in the 1800s.In order to avoid the delay of traveling around the dune fields on the detour, many travelers on the trail...
                  & 
                  \Large \textcolor{red}{\texttimes}
                  \newline \tiny Maspalomas Dunes. The Maspalomas Dunes () are sand dunes located on the south coast of the island of Gran Canaria, Province of Las Palmas, in the Canary Islands. A 404 (ha) area of the municipality of San BartolomÃ© de Tirajana, they have been protected....
                  & 
                  \Large \textcolor{red}{\texttimes}
                  \newline \tiny Maspalomas Dunes. The Maspalomas Dunes () They were one of the contenders in the 12 Treasures of Spain competition. They were formed by sand from the now subdued marine shelf, when it was laid dry during the last ice age and the wind blew the sand towards the...
                  \\
            \midrule
                  Zero-shot
                  \newline (BLIP2)
                  \newline \ding{55} inst
                  &  
                  \Large \textcolor{red}{\texttimes}
                  \includegraphics[width=20mm,height=20mm]{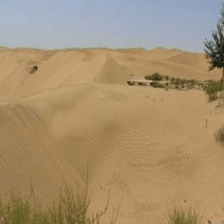}
                  &  
                  \Large \textcolor{red}{\texttimes}
                  \includegraphics[width=20mm,height=20mm]{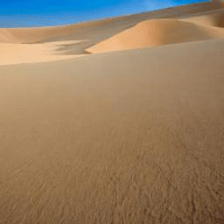}
                  &  
                  \Large \textcolor{red}{\texttimes}
                  \includegraphics[width=20mm,height=20mm]{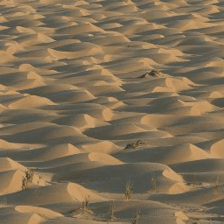}
                  \tiny Ād. Imam of Grand Mosque of Mecca, along with the officials of Ministry of Islamic Affairs, Dawah and Guidance also agreed the verse were speaking about ...
                  & 
                  \Large \textcolor{red}{\texttimes}
                  \includegraphics[width=20mm,height=20mm]{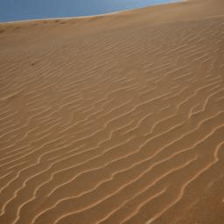} 
                  & 
                  \Large \textcolor{red}{\texttimes}
                  \includegraphics[width=20mm,height=20mm]{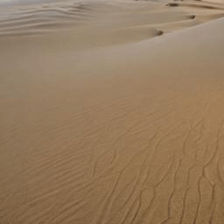}
                  \\
            \bottomrule
    \end{tabular}
    \caption{Visualization of the Top 5 Candidates Retrieved from M-BEIR (5.6M candidates) by Different Models for a OVEN Query}
    \label{fig:retrieval_results_oven}
    \end{center}
\end{figure*}

\begin{figure*}[t!]
\centering
\scriptsize
\begin{tabular}{>{\columncolor{lightgreen}}m{1.5cm} >{\columncolor{lightgreen}}m{1cm} m{2.2cm} m{3.8cm} m{2.4cm} m{3.8cm}}
    \toprule
    \textbf{Dataset} & \textbf{Domain} & \textbf{Task} & \textbf{Query Instruction  $q_{\texttt{inst}}$} & \textbf{Query Image $q_i$} & \textbf{Query Text $q_t$} \\
    \midrule
    InfoSeek 
    & Wikipedia
    & 8. ($q_i, q_t$) $\to$ ($c_i, c_t$)
    & Retrieve a Wikipedia image-description pair that provides evidence for the question of this image.
    & \includegraphics[width=20mm,height=20mm]{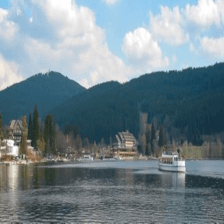}
    & Where is the lake outflow to?
\end{tabular}
\centering
    \scriptsize
    \begin{center}
    \begin{tabular}{>{\columncolor{lightgreen}}m{1.2cm}m{2.7cm}m{2.7cm}m{2.7cm}m{2.7cm}m{2.7cm}}
            \toprule
            \textbf{Model} & \textbf{Rank 1 ($c_i, c_t$)} & \textbf{Rank 2 ($c_i, c_t$)} & \textbf{Rank 3 ($c_i, c_t$)} & \textbf{Rank 4 ($c_i, c_t$)} & \textbf{Rank 5 ($c_i, c_t$)} \\
            \midrule
                  UniIR
                  \newline (CLIP$_{\text{SF}}$)
                  \newline \checkmark inst
                  &  
                  \Large \textcolor{red}{\texttimes}
                  \includegraphics[width=20mm,height=20mm]{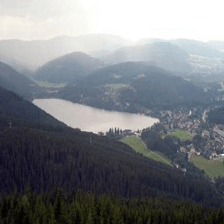}
                  \newline \tiny Titisee. the promenade. Furthermore, plenty of open-air events are organised around the lake each summer.In winter, the 1.2 km long Saig-Titisee toboggan is open. The largest natural ski jump in Germany, the Hochfirstschanze, is also located at ...
                  &  
                  \Large \textcolor{red}{\texttimes}
                  \includegraphics[width=20mm,height=20mm]{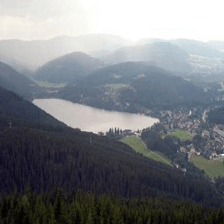} 
                  \newline \tiny Titisee. and burbot on the lake bed. In addition there are small fish varieties such as sunbleak, minnow and brook lamprey. Around the lake, grey heron may be seen.The shores of the Titisee are home to two rare types of quillworts, the spring quillwort ...
                  &  
                  \Large \textcolor{red}{\texttimes}
                  \includegraphics[width=20mm,height=20mm]{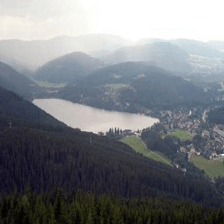} 
                  \newline \tiny Titisee. could be the origin of the name, even though it no longer occurs by the Titisee. Tourism.At the north shore of the Titisee lies a popular spa town of the same name. Lots of wellness and health hotels are settled there. Spa therapy offers ...
                  & 
                  \Large \textcolor{red}{\texttimes}
                  \includegraphics[width=20mm,height=20mm]{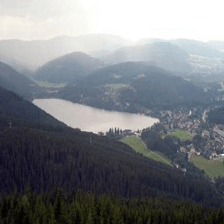} 
                  \newline \tiny Titisee. a tractor with a snowplough was clearing the landing strip of snow when it broke through the ice and sank to the bottom of the lake, taking the driver, Walter Wilde (29), with it. His body was only recovered 2 weeks later. Fauna and flora ...
                  & 
                  \Large \textcolor{red}{\texttimes}
                  \includegraphics[width=20mm,height=20mm]{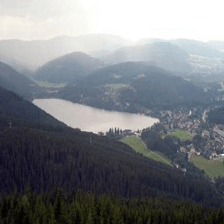} 
                  \newline \tiny Titisee. great fire. In the 18th century, the clockmaking trade developed in the town to become a major part of the economy.During World War I and shortly thereafter, a dearth of staple foods prevailed. In May 1919 the first municipa ...
                  \\
            \midrule
                  multi-task
                  \newline (CLIP$_{\text{SF}}$)
                  \newline \ding{55} inst
                  &  
                  \Large \textcolor{red}{\texttimes}
                  \includegraphics[width=20mm,height=20mm]{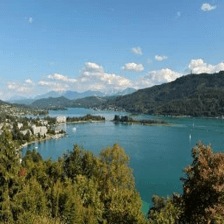}  
                  \newline 
                  \tiny  Wörthersee. ("Coregonus lavaretus"). Etymology.First mentioned as "Werdse" in an 1143 deed, the lake's name originates from the islands in the lake, most notably Maria Wörth, a peninsula since the building of the Glanfurt outlet in 1770...
                  &  
                  \Large \textcolor{red}{\texttimes}
                  \includegraphics[width=20mm,height=20mm]{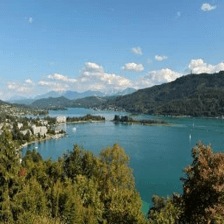} 
                  \newline 
                  \tiny Wörthersee. The lake's water is of a distinctive blue-green colour and transparent. Lake Wörth and its basin in the central Carinthian foothills were largely formed by glaciers during the last ice age. The lake is divided into three basins by several...
                  &  
                  \Large \textcolor{red}{\texttimes}
                  \includegraphics[width=20mm,height=20mm]{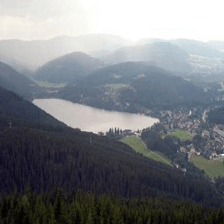} 
                  \newline 
                  \tiny Titisee. the High Black Forest had been unsettled during the first millennium. Origin of the name.There are various theories about the origin of the unusual name "Titisee": In the Alemannic dialect "Teti" means "little child" or "baby"...
                  & 
                  \Large \textcolor{red}{\texttimes}
                  \includegraphics[width=20mm,height=20mm]{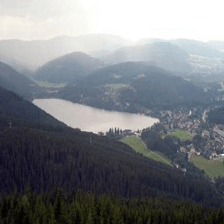} 
                  \newline 
                  \tiny  Titisee. and burbot on the lake bed. In addition there are small fish varieties such as sunbleak, minnow and brook lamprey. Around the lake, grey heron may be seen.The shores of the Titisee are home to two rare types of quillworts, the ...
                  & 
                  \Large \textcolor{red}{\texttimes}
                  \tiny Atwood Lake. Atwood Lake is a reservoir located in Tuscarawas and Carroll counties in east central Ohio. The lake is formed by Atwood Dam across Indian Fork, a tributary of Conotton Creek. The lake is named for the community of Atwood which was purchased...
                  \\
            \midrule
                  UniIR
                  \newline (BLIP$_{\text{FF}}$)
                  \newline \checkmark inst
                  &  
                  \Large \textcolor{red}{\texttimes}
                  \includegraphics[width=20mm,height=20mm]{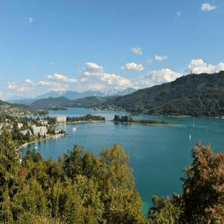}
                  \newline \tiny Wörthersee. ("Coregonus lavaretus"). Etymology.First mentioned as "Werdse" in an 1143 deed, the lake's name originates from the islands in the lake, most notably Maria Wörth, a peninsula since the building of the Glanfurt outlet in 1770...
                  &  
                  \large \textcolor{green}{\checkmark}
                  \includegraphics[width=20mm,height=20mm]{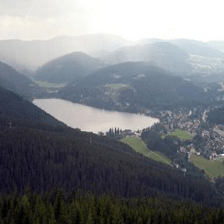} 
                  \newline \tiny Titisee. The Titisee is a lake in the southern Black Forest in Baden-Württemberg... The lake's outflow, at 840 (m) above sea level, is the \textcolor{green}{\textbf{River Gutach}}, which merges with the Haslach stream below Kappel to form the Wutach. The waters of ...
                  &  
                  \Large \textcolor{red}{\texttimes}
                  \includegraphics[width=20mm,height=20mm]{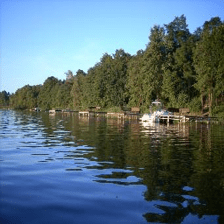} 
                  \newline \tiny Mikorzyńskie Lake. Mikorzyńskie Lake is a lake in central Poland known as Jezioro Mikorzyńskie in Polish. It is located in Konin County in Gmina Ślesin, near the town of Licheń Stary. It is a post-glacial ribbon lake with a strongly developed shoreline...
                  & 
                  \Large \textcolor{red}{\texttimes}
                  \includegraphics[width=20mm,height=20mm]{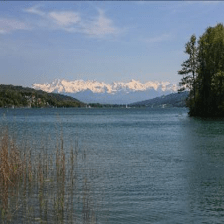} 
                  \newline \tiny Lake Hallwil. Lake Hallwil (German: "Hallwilersee") is a lake largely in the Canton of Aargau, Switzerland, located at . It is the largest lake in Aargau and lies mostly in the districts of Lenzburg and Kulm on the southern edge of the canton...
                  & 
                  \Large \textcolor{red}{\texttimes}
                  \includegraphics[width=20mm,height=20mm]{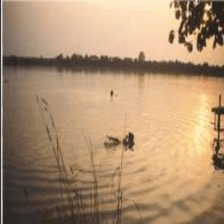} 
                  \newline \tiny Głuszyńskie Lake. Głuszyńskie is a lake in Poland near the city of Radziejów in the Kuyavian-Pomeranian Voivodship...
                  \\
            \midrule
                  multi-task
                  \newline (BLIP$_{\text{FF}}$)
                  \newline \ding{55} inst
                  &  
                  \Large \textcolor{red}{\texttimes}
                  \includegraphics[width=20mm,height=20mm]{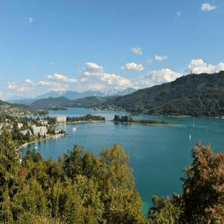}  
                  \newline \tiny Titisee. and burbot on the lake bed. In addition there are small fish varieties such as sunbleak, minnow and brook lamprey. Around the lake, grey heron may be seen. The shores of the Titisee are home to two rare types of quillworts, the...
                  &  
                  \Large \textcolor{red}{\texttimes}
                  \includegraphics[width=20mm,height=20mm]{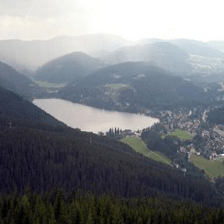} 
                  \newline \tiny Wdzydze Lake. Wdzydze (Kashubian "Jezoro Wdzydzcz"), also known as the "Kashubian Sea", "Big Water" and "Szerzawa", is a lake in the Tuchola Forest in the Kocierzyna County (Pomeranian Voivodeship), in the region of South Kashubia...
                  &  
                  \Large \textcolor{red}{\texttimes}
                  \includegraphics[width=20mm,height=20mm]{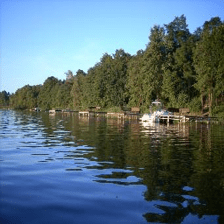} 
                  \newline \tiny Loch Lomond. Powan are one of the commonest fish species in the loch, which has more species of fish than any other loch in Scotland, including lamprey, lampern, brook trout, perch, loach, common roach and flounder...
                  & 
                  \Large \textcolor{red}{\texttimes}
                  \tiny Mountain Lake (Ontario). Mountain Lake (Ontario) may refer to one of a number of lakes in Ontario, Canada with this precise name or to others with similar names. Mountain Lake (Algoma County)...
                  & 
                  \Large \textcolor{red}{\texttimes}
                  \tiny Atwood Lake. Atwood Lake is a reservoir located in Tuscarawas and Carroll counties in east central Ohio. The lake is formed by Atwood Dam across Indian Fork, a tributary of Conotton Creek. The lake is named for the community of Atwood...
                  \\
            \midrule
                  Zero-shot
                  \newline (BLIP2)
                  \newline \ding{55} inst
                  &  
                  \Large \textcolor{red}{\texttimes}
                  \includegraphics[width=20mm,height=20mm]{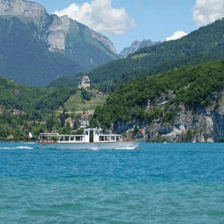}
                  \newline 
                  \tiny Lake Annecy. Human-Modified Channel Systems to Lake Annecy, SE France" (2004) Vol 20 (4) Journal of Coastal Research pp...
                  &  
                  \Large \textcolor{red}{\texttimes}
                  \includegraphics[width=20mm,height=20mm]{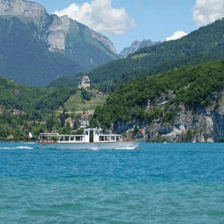}
                  \newline 
                  \tiny Lake Annecy. Lake Annecy (, ) is a perialpine lake in Haute-Savoie in France. It is named after the city of Annecy...
                  &  
                  \Large \textcolor{red}{\texttimes}
                  \includegraphics[width=20mm,height=20mm]{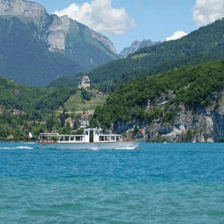}
                  \newline 
                  \tiny Lake Annecy. time the large alpine glaciers melted. It is fed by many small rivers from the surrounding mountains...
                  & 
                  \Large \textcolor{red}{\texttimes}
                  \includegraphics[width=20mm,height=20mm]{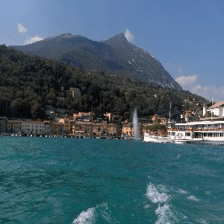} 
                  \newline 
                  \tiny Monte Pizzocolo. Monte Pizzocolo is a mountain of Lombardy, Italy. It has an elevation of 1,581 metres...
                  & 
                  \Large \textcolor{red}{\texttimes}
                  \includegraphics[width=20mm,height=20mm]{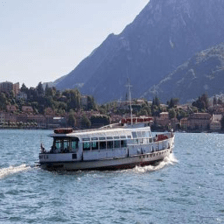}
                  \newline 
                  \tiny Lake Como, Wikimania 2016, MP 126 Lake Como. Wikimania 2016, Esino Lario, Italy...
                  \\
            \bottomrule
    \end{tabular}
    \caption{Visualization of the Top 5 Candidates Retrieved from M-BEIR (5.6M candidates) by Different Models for an Infoseek Query}
    \label{fig:retrieval_results_infoseek}
    \end{center}
\end{figure*}

\definecolor{lightblue}{rgb}{233, 237, 201}

\definecolor{lightgray}{RGB}{237, 237, 233} 

\begin{figure*}[t!]
\centering
    \scriptsize
    \begin{center}
    \tabcolsep 2pt
    \begin{tabular}{m{1.8cm}m{1.8cm}m{3.5cm}m{2.1cm}>{\columncolor{lightgray}}m{3.5cm}>{\columncolor{lightgray}}m{2cm}}
            \toprule
             \textbf{Task} & \textbf{Dataset} & \textbf{Query Text} & \textbf{Query Image} & \textbf{Retrieved Text} & \textbf{Retrieved Image} \\
            \midrule
            \multirow{13}{*}{1. $q_t \to c_i$}
                  & VisualNews~\citep{liu2020visual}
                  & \textit{Identify the news-related image in line with the described event.} \newline
                  Apple CEO Tim Cook talks about the new iPhones at an event in California 9 September 2015.
                  & -
                  & -
                  & \includegraphics[width=15mm,height=15mm]{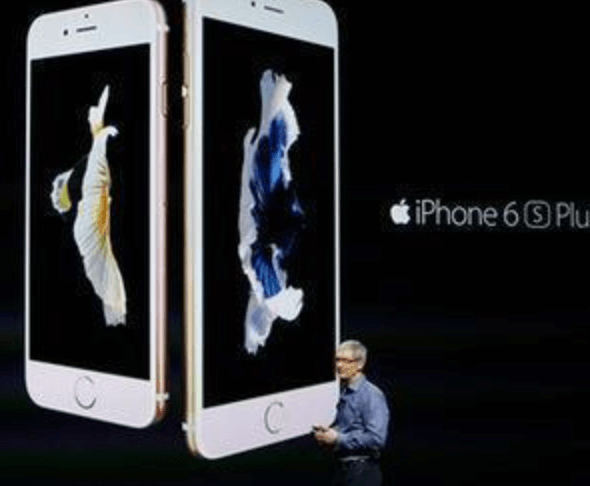}\\
            \cmidrule{2-6}
                  & MSCOCO~\citep{lin2014microsoft}
                  & \textit{Find me an everyday image that matches the given caption.} \newline A teddy bear that is on top of a desk.
                  & -
                  & -
                  & \includegraphics[width=15mm,height=15mm]{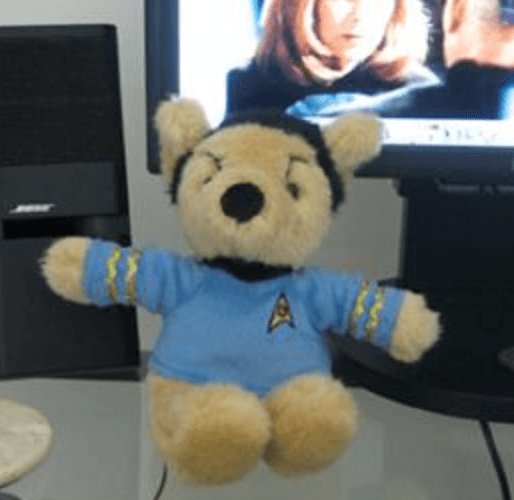}\\
            \cmidrule{2-6}
                  & Fashion200K~\citep{han2017automatic}
                  & \textit{Based on the following fashion description, retrieve the best matching image.} \newline Blue nomad suede side zip mini skirt.
                  & -
                  & -
                  & \includegraphics[width=15mm,height=15mm]{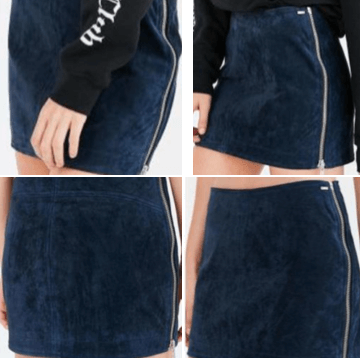}\\
            \midrule
            $2. q_t \to c_t$
                  & WebQA~\citep{chang2022webqa}
                  & \textit{Retrieve passages from Wikipedia that provide answers to the following question.} \newline Who was the first emperor to visit the the city that is at the head of the Adriatic and the seat of an ancient episcopal see?
                  & -
                  & Augustus was the first of a number of emperors to visit Aquileia, notably during the Pannonian wars in 12–10 BC. It was the birthplace of Tiberius ' son by Julia, in the latter year. The Roman poet Martial praised Aquileia as his hoped for haven and resting place in his old age.
                  & -\\
            \midrule
            \multirow{15}{*}{3. $q_t \to$ ($c_i, c_t$)}
                  & &&&&\\
                  & EDIS~\citep{liu2023edis}
                  & \textit{Find a news image that matches the provided caption.} \newline Bulgarian President Rosen Plevneliev speaks during the 68th session of the UN General Assembly.
                  & -
                  &
                  
                   National SecurityIran leader open to meeting Obama at later date.
                  & 
                  \includegraphics[width=15mm,height=15mm]{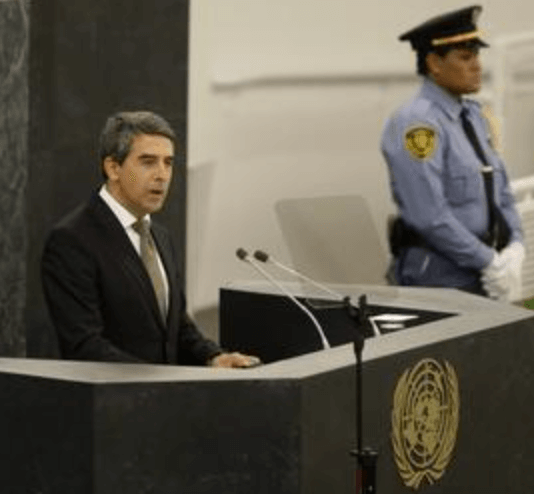}\\
            \cmidrule{2-6}
                  & &&&&\\
                  & WebQA~\citep{chang2022webqa} 
                  & \textit{Find a Wikipedia image that answers this question.} \newline Do both the Hays County Courthouse in San Marcos, Texas and the Ike Wood House at 227 Mitchell Street in San Marcos, Texas have six columns on their front entrance?   
                  & -
                  & Hays County Courthouse (2018), San Marcos, TX The Hays County Courthouse in San Marcos, Texas. Listed on the National Register of Historic Places. 
                  227 Mitchell, San Marcos, Texas  Ike Wood House at 227 Mitchell Street in San Marcos, Texas.
                  &
                  \includegraphics[width=15mm,height=15mm]{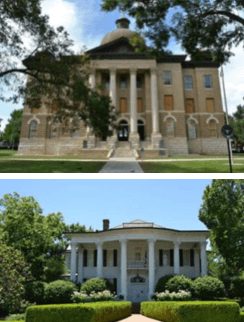}\\
            \midrule
            \multirow{13}{*}{4. $q_i \to c_t$}
                  & VisualNews~\citep{liu2020visual}
                  & \textit{Find a caption for the news in the given photo.}
                  & 
                  \includegraphics[width=15mm,height=15mm]{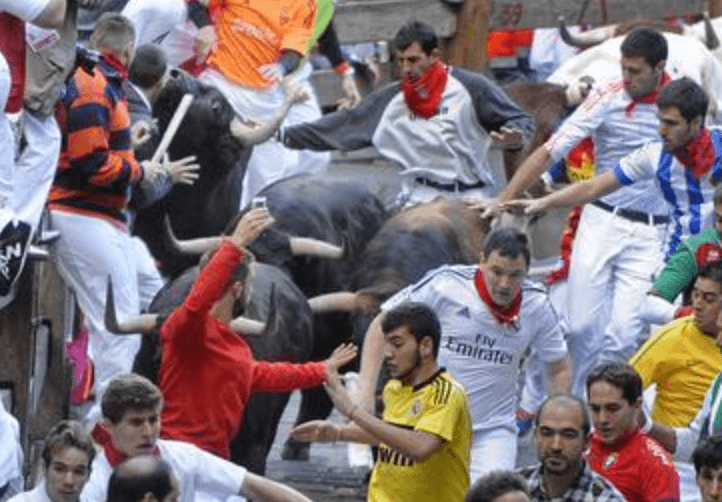}
                  & Participants run in front of Jandilla s bulls during the fifth bullrun of the San Fermin Festival in Pamplona northern Spain on July 11.
                  & -\\
            \cmidrule{2-6}
                  & MSCOCO~\citep{lin2014microsoft}
                  & \textit{Find an image caption describing the following everyday image.}
                  & 
                  \includegraphics[width=15mm,height=15mm]{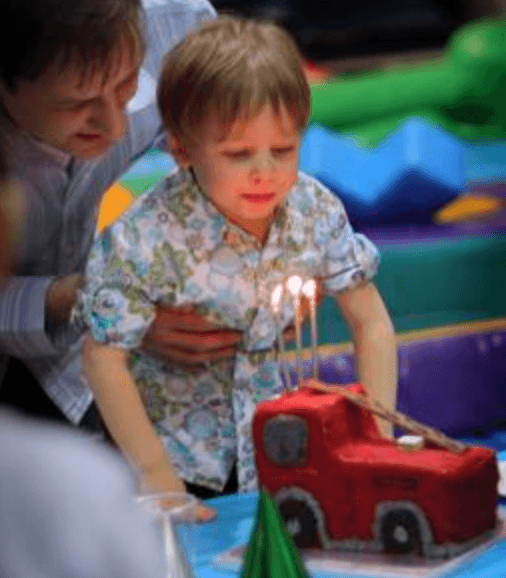}
                  & A young boy blows on candles on a fire truck cake.
                  & -\\
            \cmidrule{2-6}
                  & Fashion200K~\citep{han2017automatic}
                  & \textit{Find a product description for the fashion item in the image.}
                  & 
                  \includegraphics[width=15mm,height=15mm]{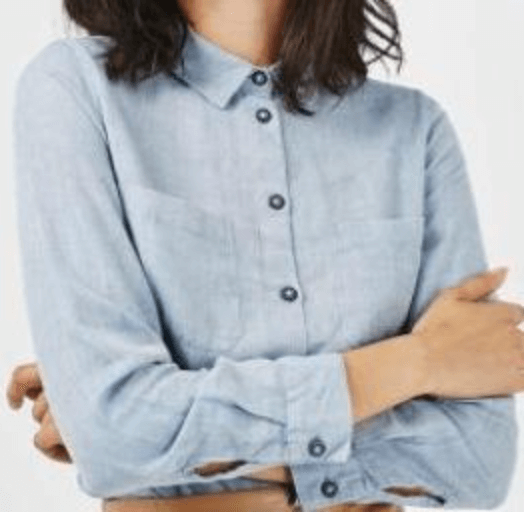}
                  & Blue washed chambray shirt.
                  & -\\
            \midrule
            \bottomrule
    \end{tabular}
    \caption{Examples of datasets in MBEIR. \textit{Query instructions} are written in italic font style.}
    \end{center}
\end{figure*}

\begin{figure*}[t!]
\centering
    \scriptsize
    \begin{center}
    \tabcolsep 2pt
    \begin{tabular}{m{2.6cm}m{1.8cm}m{3.2cm}m{2.1cm}>{\columncolor{lightgray}}m{3cm}>{\columncolor{lightgray}}m{2cm}}
            \toprule
             \textbf{Task} & \textbf{Dataset} & \textbf{Query Text} & \textbf{Query Image} & \textbf{Retrieved Text} & \textbf{Retrieved Image} \\
            \midrule
            5. $q_i \to c_i$
                  & NIGHTS~\citep{fu2023learning}
                  & \textit{Find a day-to-day image that looks similar to the provided image.}
                  & \includegraphics[width=15mm,height=15mm]{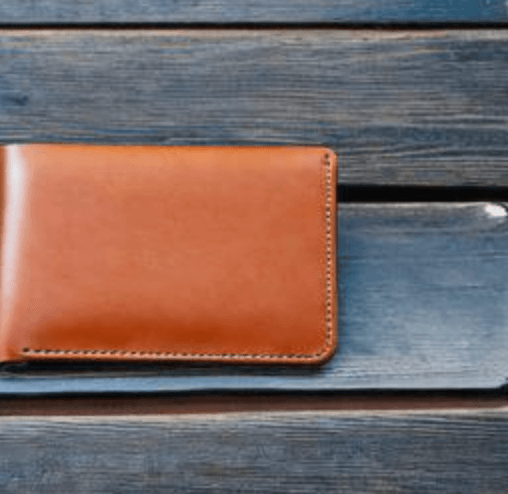}
                  & -
                  & \includegraphics[width=15mm,height=15mm]{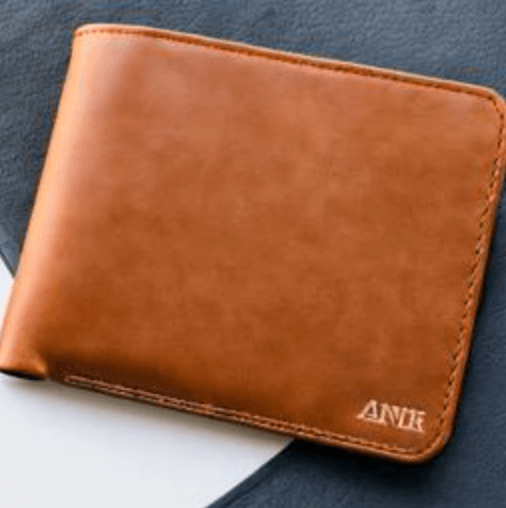}
                  \\
            \midrule
            \multirow{8}{*}{6. ($q_i, q_t$) $\to c_t$}
                  & OVEN~\citep{hu2023open}
                  & \textit{Retrieve a Wikipedia paragraph that provides an answer to the given query about the image.} \newline What is the name of this bridge?
                  & \includegraphics[width=15mm,height=15mm]{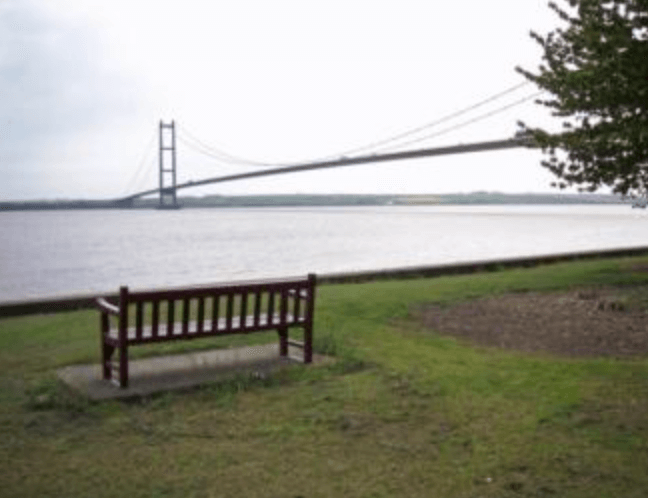}
                  & The \textbf{Humber Bridge}, near Kingston upon Hull, East Riding of Yorkshire, England, is a 2.22 km (2,430 yd; 7,300 ft; 1.38 mi) single-span road suspension bridge, which opened to traffic on 24 June 1981. 
                  & -
                  \\
            \cmidrule{2-6}
                  & InfoSeek~\citep{chen2023infoseek}
                  & \textit{Retrieve a Wikipedia paragraph that provides an answer to the given query about the image.} \newline What country does this bridge belong to?
                  & \includegraphics[width=15mm,height=15mm]{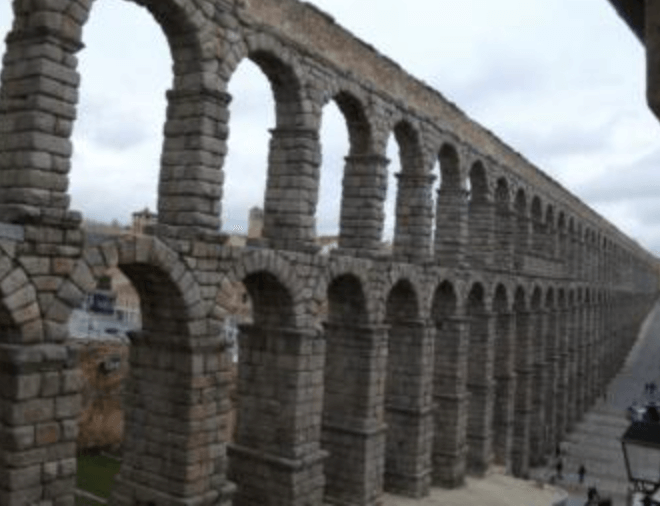}
                  & Aqueduct of Segovia. The Aqueduct of Segovia () is a Roman aqueduct in Segovia, \textbf{Spain}. It is one of the best-preserved elevated Roman aqueducts and the foremost symbol of Segovia...
                  & -
                  \\
            \midrule
            \multirow{8}{*}{7. ($q_i, q_t$) $\to c_i$}
                  & FashionIQ~\citep{wu2021fashion}
                  & \textit{Find a fashion image that aligns with the reference image and style note.} \newline Does not have writing in it and has less font and simpler graphics.
                  & \includegraphics[width=15mm,height=15mm]{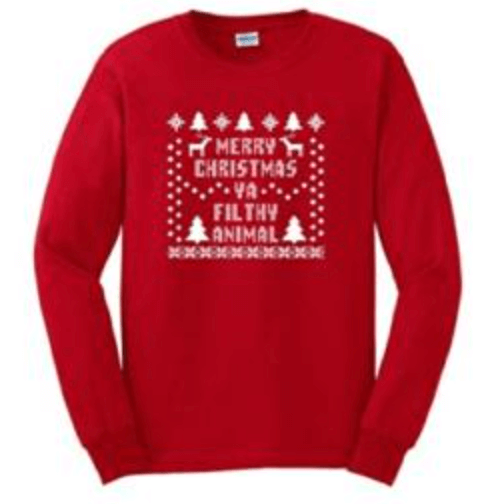}
                  & -
                  & \includegraphics[width=16mm,height=16mm]{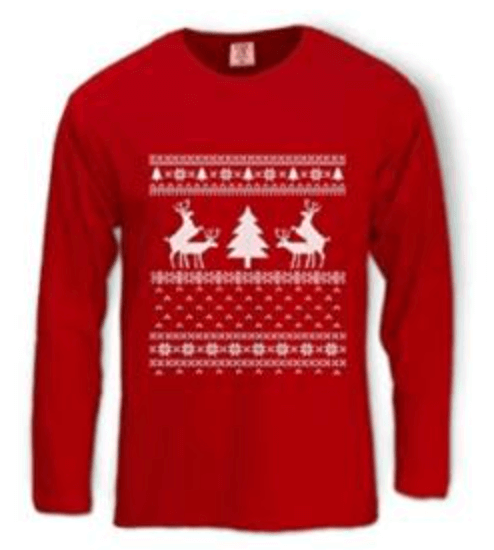}
                  \\
            \cmidrule{2-6}
                  & CIRR~\citep{liu2021image}
                  & \textit{Retrieve a day-to-day image that aligns with the modification instructions of the provided image.} \newline White fuzzy animal sits by the food bowl.
                  & \includegraphics[width=15mm,height=15mm]{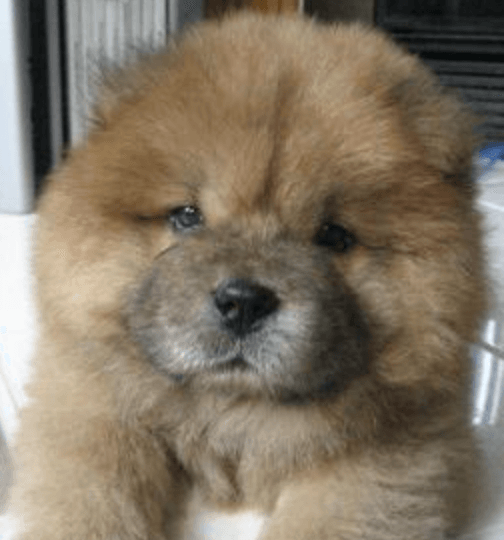}
                  & -
                  & \includegraphics[width=15mm,height=15mm]{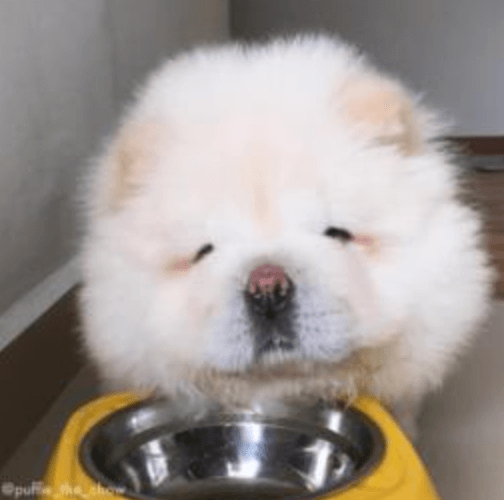}
                  \\
            \midrule
            \multirow{20}{*}{\parbox{2.5cm}{8. ($q_i, q_t$) $\to$ ($c_i, c_t$)}}
                  & &&&& \\
                  & OVEN~\citep{hu2023open}
                  &\textit{Retrieve a Wikipedia image-description pair that provides evidence for the question of this image.} \newline What is the model of this aircraft?
                  & \includegraphics[width=15mm,height=15mm]{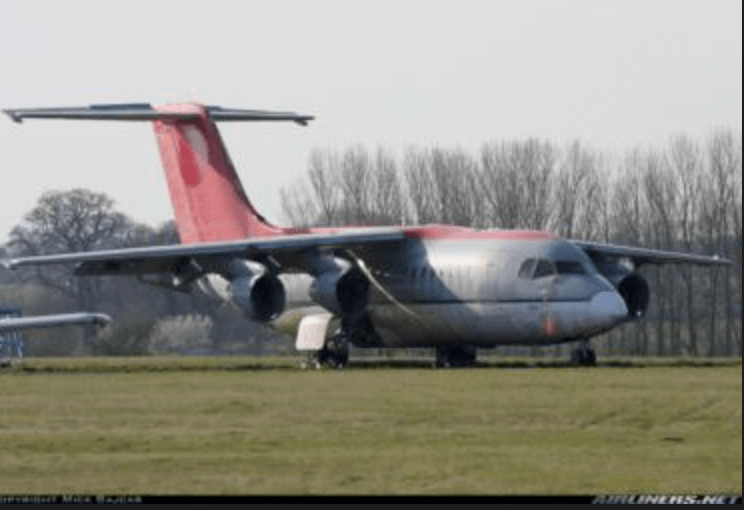}
                  & 
                  Identify the news-related image that aligns with the described event. The \textbf{British Aerospace 146} (also BAe 146) is a short-haul and regional airliner that was manufactured in the United Kingdom by British Aerospace, later part of BAE Systems. 
                  & \includegraphics[width=15mm,height=15mm]{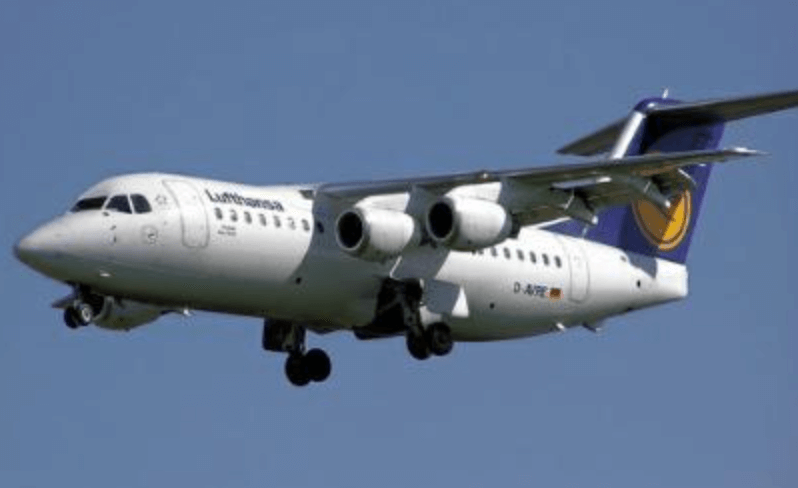}
                  \\
            \cmidrule{2-6}
                    & &&&&\\
                  & InfoSeek~\citep{chen2023infoseek}
                  & \textit{Retrieve a Wikipedia image-description pair that provides evidence for the question of this image.} \newline What is the country of origin of this animal?
                  & \includegraphics[width=15mm,height=15mm]{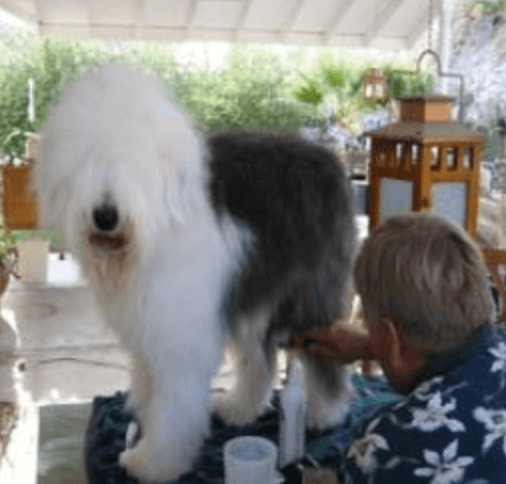}
                  &
                  The Old English Sheepdog is a large breed of dog that emerged in \textbf{England} from early types of herding dog. 
                  & \includegraphics[width=15mm,height=15mm]{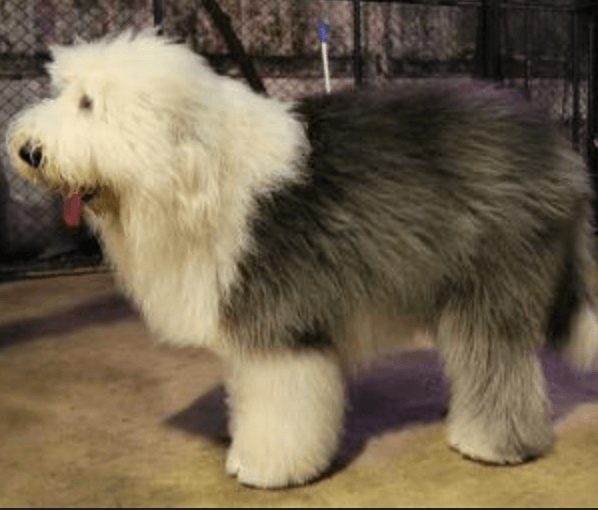}
                  \\
            \bottomrule
    \end{tabular}
    \caption{Additional examples of datasets in MBEIR. \textit{Query instructions} are written in italic font style.
    }
    \end{center}
\end{figure*}
\begin{table*}
    \centering
    \scriptsize
    \begin{tabular}{l l l}
    \toprule
        \textbf{Task}        & \textbf{Dataset}             & \textbf{Query Instruction}                                               \\
        \midrule
         \multirow{12}{*}{1. $q_{t}$ $\to$ $c_i$} 
         & \multirow{4}{*}{VisualNews}                &  Identify the news-related image in line with the described event. \\
         & & Display an image that best captures the following caption from the news. \\
         & & Based on the caption, provide the most fitting image for the news story. \\
         & & I want you to retrieve an image of this news caption. \\
         \cmidrule(lr){2-3}
         & \multirow{4}{*}{MSCOCO}                  & Find me an everyday image that matches the given caption. \\
         && Identify the image showcasing the described everyday scene. \\
         && I want you to retrieve an image of this daily life description. \\
         && Show me an image that best captures the following common scene description. \\
         \cmidrule(lr){2-3}
         & \multirow{4}{*}{Fashion200K}            & Based on the following fashion description, retrieve the best matching image.  \\
         && Match the provided description to the correct fashion item photo.  \\
         && Identify the fashion image that aligns with the described product.  \\
         && You need to identify the image that corresponds to the fashion product description provided.  \\
         \midrule
         \multirow{4}{*}{2. $q_{t}$ $\to$ $c_t$} &      & Retrieve passages from Wikipedia that provide answers to the following question. \\
         &WebQA&You have to find a Wikipedia paragraph that provides the answer to the question. \\
         &&I want to find an answer to the question. Can you find some snippets that provide evidence from Wikipedia? \\
         &&RI'm looking for a Wikipedia snippet that answers this question. \\
         \midrule
         \multirow{8}{*}{3. $q_t$ $\to$ ($c_i$, $c_t$)}
         & \multirow{4}{*}{EDIS}                      &  Find a news image that matches the provided caption. \\
         &&  Identify the news photo for the given caption. \\
         &&  Can you pair this news caption with the right image? \\
         &&  I'm looking for an image that aligns with this news caption. \\
         \cmidrule(lr){2-3}
         & \multirow{4}{*}{WebQA}                     & Find a Wikipedia image that answers this question. \\
         && Provide with me an image from Wikipedia to answer this question. \\
         && I want to know the answer to this question. Please find the related Wikipedia image for me. \\
         && You need to retrieve an evidence image from Wikipedia to address this question. \\
         \midrule
         \multirow{12}{*}{4. $q_{i}$ $\to$ $c_t$} 
         & \multirow{4}{*}{VisualNews}                &  Find a caption for the news in the given photo. \\
         & & Based on the shown image, retrieve an appropriate news caption. \\
         & & Provide a news-related caption for the displayed image. \\
         & & I want to know the caption for this news image. \\
         \cmidrule(lr){2-3}
         & \multirow{4}{*}{MSCOCO}                  & Find an image caption describing the following everyday image. \\
         && Retrieve the caption for the displayed day-to-day image. \\
         && Can you find a caption talking about this daily life image? \\
         && I want to locate the caption that best describes this everyday scene image. \\
         \cmidrule(lr){2-3}
         & \multirow{4}{*}{Fashion200K}            & Find a product description for the fashion item in the image.  \\
         && Based on the displayed image, retrieve the corresponding fashion description.  \\
         && Can you retrieve the description for the fashion item in the image?  \\
         && I want to find a matching description for the fashion item in this image.  \\
         \bottomrule
    \end{tabular}
    \caption{M-BEIR query instructions.}
    \label{tab:instructions}
\end{table*}

\begin{table*}
    \centering
    \scriptsize
    \begin{tabular}{l l l}
    \toprule
        \textbf{Task}        & \textbf{Dataset}             & \textbf{Query Instruction}                                               \\
        \midrule
         \multirow{4}{*}{5. $q_{i}$ $\to$ $c_t$} &      & Find a day-to-day image that looks similar to the provided image. \\
         &NIGHTS&Which everyday image is the most similar to the reference image? \\
         &&Find a daily life image that is identical to the given one. \\
         &&You need to identify the common scene image that aligns most with this reference image. \\
         \midrule
         \multirow{8}{*}{6. ($q_i, q_t$) $\to$ $c_t$} 
         & \multirow{4}{*}{OVEN}                          & Retrieve a Wikipedia paragraph that provides an answer to the given query about the image. \\
         && Determine the Wikipedia snippet that identifies the visual entity in the image. \\
         && I want to find a paragraph from Wikipedia that answers my question about this image. \\
         && You have to find a Wikipedia segment that identifies this image's subject. \\
         \cmidrule(lr){2-3}
         & \multirow{4}{*}{InfoSeek}                & Retrieve a Wikipedia paragraph that provides an answer to the given query about the image. \\
         &     & Determine the Wikipedia snippet that matches the question of this image. \\
         &     & I want to find a paragraph from Wikipedia that answers my question about this image. \\
         &     & You have to find a Wikipedia segment that answers the question about the displayed image. \\
         \midrule
         \multirow{8}{*}{7. ($q_i, q_t$) $\to$ $c_i$} 
         & \multirow{4}{*}{FashionIQ}                          & Find a fashion image that aligns with the reference image and style note. \\
         && With the reference image and modification instructions, find the described fashion look. \\
         && Given the reference image and design hint, identify the matching fashion image. \\
         && I'm looking for a similar fashion product image with the described style changes. \\
         \cmidrule(lr){2-3}
         & \multirow{4}{*}{CIRR}                & Retrieve a day-to-day image that aligns with the modification instructions of the provided image. \\
         &     & Pull up a common scene image like this one, but with the modifications I asked for. \\
         &     & Can you help me find a daily image that meets the modification from the given image? \\
         &     & I'm looking for a similar everyday image with the described changes. \\
         \midrule
         \multirow{8}{*}{8. ($q_i, q_t$) $\to$ ($c_i, c_t$)} 
         & \multirow{4}{*}{OVEN}                          & Retrieve a Wikipedia image-description pair that provides evidence for the question of this image. \\
         && Determine the Wikipedia image-snippet pair that clarifies the entity in this picture. \\
         && I want to find an image and subject description from Wikipedia that answers my question about this image. \\
         && I want to know the subject in the photo. Can you provide the relevant Wikipedia section and image? \\
         \cmidrule(lr){2-3}
         & \multirow{4}{*}{InfoSeek}                & Retrieve a Wikipedia image-description pair that provides evidence for the question of this image. \\
         &     & Determine the Wikipedia image-snippet pair that matches my question about this image. \\
         &     & I want to find an image and subject description from Wikipedia that answers my question about this image. \\
         &     & I want to address the query about this picture. Please pull up a relevant Wikipedia section and image. \\
         \bottomrule
    \end{tabular}
    \caption{M-BEIR query instructions.}
    \label{tab:instructions2}
\end{table*}
\begin{table*}
    \centering
    \LARGE
    \resizebox{\textwidth}{!}{%
    \begin{tabular}{lllcccc|ccccc|ccccc}
    \toprule
     &&&\multicolumn{4}{c}{\textbf{Zero-shot}}& \multicolumn{5}{c}{\textbf{Multi-task} (\ding{55} instruction)}& \multicolumn{5}{c}{\textbf{UniIR} (\checkmark instruction)}\\
     \cmidrule(lr){4-7}\cmidrule(lr){8-12}\cmidrule(lr){13-17}
Task & Dataset & Metric &  CLIP & SigLIP & BLIP & BLIP2 & CLIP$_{\text{SF}}$ & CLIP$_{\text{FF}}$ & BLIP$_{\text{SF}}$ & BLIP$_{\text{FF}}$ &
    BLIP$_{\text{FF, 384}}$
     & CLIP$_{\text{SF}}$ & CLIP$_{\text{FF}}$& BLIP$_{\text{SF}}$ & BLIP$_{\text{FF}}$&
    BLIP$_{\text{FF, 384}}$ \\
        \midrule
\multirow{9}{*}{1. $q_t \to c_i$}
         &               & R@1 & 0.0 & 0.0 & 0.0 & 0.0 & 0.0 & 0.0 & 0.0 &0.0 & 0.0 &22.3 & 12.8 &8.4 &9.9 &11.8\\
         & VisualNews              & R@5 & 0.0 & 0.0 & 0.0 & 0.0 & 12.7 & 8.8 & 5.0 &8.3 & 10.5 &42.6 & 28.8 &20.9&23.0 &26.5\\
         &               & R@10 & 0.0 & 0.0 & 0.0 & 0.0 & 22.3 & 15.8 & 9.4 &14.5 & 17.2 &51.9 & 37.1 &27.4&29.9 &34.0\\
         \cmidrule(lr){2-17}
         &                 & R@1 & 0.0 & 0.0 & 0.0 & 0.0 & 0.0 & 0.0 & 0.0 &0.0 & 0.0 &45.8 & 47.5 &23.3&37.4 &20.4\\
         & MSCOCO                & R@5 & 0.0 & 0.0 & 0.0 & 0.0 & 27.3 & 24.6 & 22.9 &27.7 & 35.2 &77.9 & 74.7 &71.6&75.6 &75.7\\
         &                 & R@10& 0.0 & 0.0 & 0.0 & 0.0 & 47.3 & 42.8 & 41.0 &45.8 & 53.9 &86.0 & 83.5 &81.1&84.3 &84.6\\
         \cmidrule(lr){2-17}
         &              & R@10& 0.0 & 0.0 & 0.0 & 0.0 & 5.9 & 5.9 & 5.7 &9.0 & 13.1 &17.8 & 15.5 &24.3&25.4 &26.7\\
         & Fashion200K             & R@20& 0.0 & 0.0 & 0.0 & 0.0 & 12.1 & 10.9 & 12.4 &17.4 & 22.2 &24.5 & 21.6 &32.8&35.4 &34.3\\
         &              & R@50& 0.0 & 0.0 & 0.0 & 0.0 & 25.6 & 21.0 & 26.7 &30.3 & 36.4 &36.2 & 32.8 &45.4&48.3 &48.1\\
         \midrule
         \multirow{3}{*}{2. $q_t \to c_t$}
         &      & R@1 & 17.4 & 17.0 & 21.2 & 20.7 & 56.4 & 42.4 & 48.3 &49.0 & 48.5 &58.3 & 51.6 &51.6&51.6 &52.8\\
         & WebQA     & R@5& 32.1 & 34.0 & 38.1 & 35.2 & 82.3 & 67.9 & 74.4 &76.1 & 76.9 &84.7 & 78.4 &78.9&79.5 &79.2\\
         &      & R@10& 38.9 & 41.5 & 45.7 & 41.5 & 87.5 & 75.6 & 81.2 &82.5 & 83.2 &89.5 & 85.4 &84.9&85.2 &85.5\\
         \midrule
         \multirow{6}{*}{3. $q_t \to$ ($c_i, c_t$)}
         &   &  R@1& 1.4 & 0.1 & 0.0 & 0.0 & 15.2 & 12.8 & 13.5 &14.0 & 15.7 &30.1 & 24.5 &23.3&25.2 &25.7\\
         & EDIS  & R@5 & 6.7 & 1.1 & 0.0 & 0.0 & 41.1 & 38.3 & 33.6 &36.0 & 38.5 &59.4 & 50.0 &47.2&50.3 &51.4\\
         &   &  R@10& 9.8 & 2.1 & 0.0 & 0.0 & 54.4 & 51.2 & 43.4 &45.9 & 49.8 &70.4 & 60.8 &56.8&60.9 &63.2\\
         \cmidrule(lr){2-17}
         &  &  R@1& 1.7 & 0.4 & 0.0 & 0.0 & 38.9 & 34.6 & 43.2 &43.7 & 46.4 &50.1 & 48.8 &48.5&50.9 &51.5\\
         & WebQA &  R@5& 5.5 & 2.1 & 0.0 & 0.0 & 68.2 & 62.5 & 73.2 &74.7 & 75.2 &78.8 & 75.3 &76.8&79.7 &79.4\\
         &  &  R@10& 7.8 & 3.6 & 0.0 & 0.0 & 78.8 & 72.8 & 82.9 &83.0 & 83.5 &86.5 & 83.5 &85.3&87.1 &87.1\\
         \midrule
         \multirow{9}{*}{4. $q_i \to c_t$}
         &               & R@1& 0.0 & 0.0 & 0.0 & 0.0 & 0.0 & 0.0 & 0.0 &0.0 & 0.0 &22.6 & 12.6 &5.2&7.4 &9.2\\
         & VisualNews              & R@5& 0.0 & 0.0 & 0.0 & 0.0 & 12.1 & 8.2 & 4.8 &4.9 & 6.0 &42.8 & 28.6 &19.4&21.1 &22.9\\
         &               & R@10& 0.0 & 0.0 & 0.0 & 0.0 & 22.4 & 15.4 & 9.3 &9.6 & 11.4 &52.2 & 37.0 &26.9&28.7 &30.4\\
         \cmidrule(lr){2-17}
         &                 & R@1& 0.0 & 0.0 & 0.0 & 0.0 & 0.0 & 0.0 & 0.0 &0.0 & 0.0 &73.8 & 66.9 &39.7&53.4 &62.4\\
         & MSCOCO                & R@5& 0.0 & 0.0 & 0.0 & 0.0 & 84.6 & 80.8 & 74.9 &76.9 & 81.4 &92.3 & 89.0 &88.2&88.8 &90.1\\
         &                 & R@10& 0.0 & 0.0 & 0.0 & 0.0 & 92.3 & 90.2 & 85.9 &87.5 & 91.1 &96.0 & 94.5 &94.1&94.8 &95.7\\
         \cmidrule(lr){2-17}
         &           & R@10& 0.0 & 0.0 & 0.0 & 0.0 & 1.2 & 1.3 & 2.6 &3.6 & 4.0 &17.9 & 13.7 &24.3&27.6 &28.4\\
         & Fashion200K          & R@20& 0.0 & 0.0 & 0.0 & 0.0 & 2.9 & 3.0 & 5.9 &7.5 & 8.5 &25.7 & 20.4 &33.5&37.4 &38.3\\
         &           & R@50& 0.0 & 0.0 & 0.0 & 0.0 & 7.3 & 6.5 & 13.9 &16.6 & 17.9 &38.1 & 32.6 &46.2&51.6 &52.3\\
         \midrule
         \multirow{3}{*}{5. $q_i \to c_i$}
         &      & R@1 & 6.5 & 7.5 & 6.7 & 6.3 & 8.3 & 7.5 & 9.0 &8.5 & 9.1 &8.3 & 7.3 &8.5&8.2 &8.8\\
         & NIGHTS     & R@5& 25.3 & 28.7 & 25.1 & 24.0 & 31.0 & 30.8 & 32.9 &31.3 & 32.5 &32.0 & 31.9 &33.4&33.0 &33.7\\
         &      & R@10& 42.2 & 46.0 & 40.1 & 38.1 & 52.4 & 52.9 & 52.8 &52.8 & 55.5 &53.7 & 52.1 &53.8&53.7 &54.4\\
         \midrule
         \multirow{6}{*}{6. ($q_i, q_t$) $\to c_t$}
         &                        & R@1& 0.0 & 0.0 & 0.0 & 0.0 & 22.2 & 19.5 & 20.0 &24.6 & 28.4 &26.1 & 22.4 &21.3&25.6 &27.4\\
         & OVEN                       & R@5& 0.0 & 0.0 & 0.0 & 0.0 & 36.8 & 31.6 & 33.2 &37.7 & 39.2 &39.2 & 34.7 &35.2&38.7 &40.7\\
         &                        & R@10& 0.0 & 0.0 & 0.0 & 0.0 & 43.2 & 37.6 & 39.3 &43.7 & 44.3 &45.1 & 40.5 &41.6&44.8 &46.4\\
         \cmidrule(lr){2-17}
         &                & R@1& 0.0 & 0.0 & 0.0 & 0.0 & 8.4 & 7.0 & 4.6 &7.6 & 8.7 &11.4 & 8.5 &7.5&9.2 &9.5\\
         & InfoSeek               & R@5& 0.0 & 0.0 & 0.0 & 0.0 & 18.3 & 15.4 & 11.9 &17.8 & 17.1 &24.0 & 17.5 &16.7&19.7 &19.2\\
         &                & R@10& 0.0 & 0.0 & 0.0 & 0.0 & 25.0 & 21.8 & 16.9 &23.5 & 22.6 &31.2 & 23.2 &22.4&25.9 &25.2\\
         \midrule
         \multirow{6}{*}{7. ($q_i, q_t$) $\to c_i$}
         &                & R@10& 4.4 & 4.8 & 2.2 & 3.9 & 22.8 & 19.7 & 26.1 &28.1 & 29.0 &24.3 & 20.5 &26.2&28.5 &29.8\\
         & FashionIQ               & R@20& 6.7 & 6.5 & 3.7 & 6.2 & 30.7 & 26.4 & 33.6 &35.9 & 36.2 &32.5 & 27.8 &34.1&36.2 &36.5\\
         &                & R@50& 11.6 & 9.7 & 6.1 & 10.1 & 42.1 & 36.7 & 45.6 &47.6 & 47.1 &43.2 & 39.5 &46.1&46.9 &47.1\\
         \cmidrule(lr){2-17}
         &                     &  R@1& 0.7 & 0.5 & 0.8 & 0.9 & 4.1 & 10.9 & 11.5 &20.7 & 23.5 &8.2 & 13.1 &11.9&24.4 &25.0\\
         & CIRR                    & R@5 & 5.4 & 7.1 & 7.4 & 6.2 & 32.0 & 32.7 & 36.7 &45.1 & 47.4 &43.9 & 40.9 &43.0&51.4 &51.1\\
         &                     &  R@10& 8.2 & 10.8 & 11.3 & 9.0 & 43.2 & 42.9 & 47.7 &55.8 & 58.2 &56.8 & 52.7 &54.8&61.8 &61.4\\
         \midrule
         \multirow{6}{*}{8. ($q_i, q_t$) $\to$ ($c_i, c_t$)}
         &                        & R@1& 14.9 & 18.1 & 4.4 & 7.8 & 44.1 & 37.0 & 36.4 &37.6 & 40.5 &45.9 & 42.2 &36.0&44.5 &46.9\\
         & OVEN                       & R@5& 24.5 & 27.2 & 10.1 & 13.8 & 58.7 & 50.1 & 51.0 &51.6 & 53.1 &60.2 & 55.8 &51.8&57.8 &59.5\\
         &                        & R@10& 29.5 & 32.0 & 13.5 & 17.4 & 63.8 & 55.7 & 57.0 &57.4 & 58.2 &65.4 & 61.2 &58.6&62.8 &64.4\\
         \cmidrule(lr){2-17}
         &                & R@1& 7.0 & 8.7 & 1.7 & 3.4 & 22.4 & 16.0 & 11.4 &12.3 & 12.6 &25.7 & 18.6 &11.4&13.7 &16.6\\
         & InfoSeek               & R@5& 22.1 & 24.3 & 7.9 & 11.4 & 42.3 & 31.5 & 23.0 &25.4 & 25.2 &44.6 & 36.8 &25.4&27.7 &31.1\\
         &                & R@10& 29.7 & 31.4 & 11.9 & 16.0 & 50.9 & 39.5 & 30.4 &32.8 & 32.2 &53.8 & 45.2 &32.9&35.9 &38.8\\
         \midrule
         \multirow{3}{*}{-}
         &      & R@1 & 3.4 & 3.6 & 2.3 & 2.7 & 15.6 & 13.4 & 14.5 &16.2 & 17.5 &30.5 & 26.7 &23.2&27.7 &28.3\\
         & Average     & R@5 & 8.0 & 8.2 & 5.8 & 6.1 & 37.1 & 32.7 & 33.1 &35.9 & 37.8 &50.3 & 44.5 &44.3&47.2 &48.1\\
         &      & R@10 & 11.1 & 11.1 & 8.0 & 8.3 & 47.4 & 42.4 & 42.7 &45.6 & 47.7 &59.8 & 53.8 &53.6&56.4 &57.4\\

         \bottomrule
    \end{tabular}
    }
    \caption{Benchmarking information retrieval recall@1/5/10 on M-BEIR. For Fashion200K and FashionIQ, we report recall@10/20/50 following original work.}
    \label{tab:union}
\end{table*}
\begin{table*}[]
\centering
\resizebox{\textwidth}{!}{%
\begin{tabular}{lll|cccc|cccc}
\toprule
\textbf{} &
  \textbf{} &
  \textbf{} &
  \multicolumn{4}{c}{Multi-task (\ding{55} instruction)} &
  \multicolumn{4}{c}{UniIR (\checkmark instruction)} \\ \cmidrule(l){4-7}\cmidrule(l){8-11} 
 &
   &
   &
  \multicolumn{2}{c}{CLIP$_{\text{SF}}$} &
  \multicolumn{2}{c}{CLIP$_{\text{FF}}$} &
  \multicolumn{2}{c}{CLIP$_{\text{SF}}$} &
  \multicolumn{2}{c}{CLIP$_{\text{FF}}$} \\ \cmidrule(l){4-5}\cmidrule(l){6-7} \cmidrule(l){8-9}\cmidrule(l){10-11} 
 &
  Dataset &
  Metric &
  \multicolumn{1}{l}{M-BEIR$_{\text{local}}$} &
  \multicolumn{1}{l}{M-BEIR} &
  \multicolumn{1}{l}{M-BEIR$_{\text{local}}$} &
  \multicolumn{1}{l}{M-BEIR} &
  \multicolumn{1}{l}{M-BEIR$_{\text{local}}$} &
  \multicolumn{1}{l}{M-BEIR} &
  \multicolumn{1}{l}{M-BEIR$_{\text{local}}$} &
  \multicolumn{1}{l}{M-BEIR} \\ 
  \midrule
\multirow{9}{*}{1. $q_t \to c_i$} &  & R@1 & 10.4 & 0.0 & 6.4 & 0.0 & 11.8 & 11.5 & 7.0 & 6.9\\
 & VisualNews & R@5 & 24.1 & 2.9 & 16.6 & 3.5 & 26.3 & 26.1 & 18.1 & 18.1\\
 &  & R@10 & 31.0 & 6.0 & 22.6 & 7.5 & 34.1 & 34.0 & 25.1 & 25.0\\
 \cmidrule{2-11}
 &  & R@1 & 41.8 & 0.0 & 38.0 & 0.0 & 43.8 & 36.1 & 39.5 & 36.3\\
 & MSCOCO & R@5 & 70.3 & 11.6 & 67.0 & 12.8 & 72.1 & 67.2 & 68.9 & 64.6\\
 &  & R@10 & 80.1 & 27.0 & 77.8 & 28.0 & 82.0 & 77.2 & 79.6 & 75.2\\
 \cmidrule{2-11}
 &  & R@10 & 10.2 & 3.3 & 9.6 & 3.9 & 9.7 & 9.6 & 9.9 & 9.9\\
 & Fashion200K & R@20 & 15.7 & 7.2 & 14.9 & 7.5 & 14.4 & 14.4 & 13.4 & 13.4\\
 &  & R@50 & 26.4 & 16.9 & 23.9 & 16.9 & 22.8 & 22.8 & 22.8 & 22.8\\
 \midrule
\multirow{3}{*}{2. $q_t 	\to c_t$} &  & R@1 & 55.1 & 53.7 & 49.0 & 47.8 & 54.0 & 54.0 & 51.3 & 51.3\\
 & WebQA & R@5 & 82.0 & 80.5 & 74.1 & 72.9 & 81.1 & 81.1 & 79.1 & 79.1\\
 &  & R@10 & 87.9 & 86.0 & 81.7 & 80.3 & 87.8 & 87.8 & 85.6 & 85.6\\
 \midrule
\multirow{6}{*}{3. $q_t 	\to$ ($c_i, c_t$)} &  & R@1 & 22.7 & 12.5 & 22.7 & 11.4 & 25.7 & 25.8 & 23.1 & 23.1\\
 & EDIS & R@5 & 46.8 & 36.4 & 46.8 & 34.6 & 50.9 & 50.8 & 46.3 & 46.2\\
 &  & R@10 & 57.3 & 47.6 & 56.7 & 45.5 & 60.9 & 60.9 & 56.2 & 56.2\\
 \cmidrule{2-11}
 &  & R@1 & 46.4 & 37.5 & 43.2 & 34.7 & 46.8 & 46.9 & 46.4 & 46.8\\
 & WebQA & R@5 & 74.3 & 67.0 & 71.5 & 63.4 & 74.5 & 74.7 & 74.4 & 74.4\\
 &  & R@10 & 83.6 & 77.3 & 81.0 & 73.9 & 84.0 & 83.9 & 83.8 & 83.8\\
 \midrule
\multirow{9}{*}{4. $q_i 	\to c_t$} &  & R@1 & 10.4 & 0.0 & 6.5 & 0.0 & 12.5 & 12.2 & 7.4 & 7.4\\
 & VisualNews & R@5 & 23.8 & 4.0 & 16.2 & 4.5 & 27.1 & 26.8 & 18.4 & 18.4\\
 &  & R@10 & 31.0 & 8.1 & 22.2 & 8.6 & 34.9 & 34.6 & 25.0 & 25.0\\
 \cmidrule{2-11}
 &  & R@1 & 55.6 & 0.0 & 50.3 & 0.0 & 60.8 & 60.7 & 55.5 & 55.5\\
 & MSCOCO & R@5 & 80.7 & 67.5 & 77.9 & 69.0 & 84.0 & 84.0 & 81.8 & 81.8\\
 &  & R@10 & 88.8 & 79.6 & 86.8 & 81.0 & 90.6 & 90.6 & 89.5 & 89.5\\
 \cmidrule{2-11}
 &  & R@10 & 11.6 & 0.7 & 9.7 & 0.8 & 10.3 & 10.2 & 9.1 & 9.1\\
 & Fashion200K & R@20 & 16.7 & 1.5 & 14.5 & 2.2 & 15.3 & 15.3 & 13.8 & 13.8\\
 &  & R@50 & 26.8 & 4.0 & 24.5 & 5.4 & 25.1 & 25.0 & 22.8 & 22.8\\
 \midrule
\multirow{3}{*}{5. $q_i 	\to c_i$} &  & R@1 & 8.1 & 8.1 & 7.3 & 7.3 & 8.0 & 8.0 & 8.1 & 8.1\\
 & NIGHTS & R@5 & 29.6 & 29.2 & 27.1 & 26.9 & 31.9 & 31.9 & 29.1 & 29.0\\
 &  & R@10 & 49.2 & 48.4 & 46.1 & 45.3 & 48.8 & 48.7 & 48.3 & 48.2\\
 \midrule
\multirow{6}{*}{6. ($q_i, q_t$) $	\to c_t$} &  & R@1 & 16.9 & 15.0 & 15.9 & 13.9 & 16.4 & 16.4 & 15.2 & 15.5\\
 & OVEN & R@5 & 33.9 & 26.8 & 31.9 & 24.2 & 32.6 & 28.2 & 31.6 & 26.6\\
 &  & R@10 & 41.6 & 32.6 & 39.4 & 29.6 & 40.1 & 34.0 & 39.2 & 32.4\\
 \cmidrule{2-11}
 &  & R@1 & 7.0 & 4.4 & 7.1 & 3.8 & 7.0 & 5.4 & 6.6 & 5.3\\
 & InfoSeek & R@5 & 16.8 & 10.6 & 16.3 & 9.7 & 16.9 & 13.3 & 15.6 & 12.9\\
 &  & R@10 & 22.2 & 14.5 & 21.9 & 13.6 & 22.7 & 18.4 & 20.5 & 17.1\\
 \midrule
\multirow{6}{*}{7. ($q_i, q_t$) $	\to c_i$} &  & R@10 & 16.3 & 15.7 & 13.2 & 12.8 & 17.3 & 16.5 & 14.1 & 14.0\\
 & FashionIQ & R@20 & 22.5 & 21.7 & 19.0 & 18.6 & 23.4 & 22.5 & 20.0 & 19.8\\
 &  & R@50 & 33.0 & 32.2 & 28.7 & 27.9 & 34.3 & 32.8 & 30.5 & 30.1\\
 \cmidrule{2-11}
 &  & R@1 & 2.4 & 2.2 & 7.4 & 6.0 & 2.6 & 2.6 & 9.7 & 9.6\\
 & CIRR & R@5 & 34.3 & 23.2 & 31.2 & 23.0 & 35.6 & 33.2 & 35.1 & 34.1\\
 &  & R@10 & 45.4 & 32.3 & 42.4 & 32.0 & 48.0 & 45.9 & 48.5 & 47.5\\
 \midrule
\multirow{6}{*}{8. ($q_i, q_t$) $	\to$ ($c_i, c_t$)} & 
 & 
R@1 & 
36.1 & 
29.1 & 
31.9 & 
27.3 & 
34.5 & 
32.6 & 
31.4 & 
30.6\\
 & OVEN & R@5 & 55.5 & 42.1 & 50.7 & 39.8 & 54.5 & 46.5 & 50.9 & 43.7\\
 &  & R@10 & 62.7 & 47.8 & 58.2 & 45.3 & 61.8 & 52.5 & 58.2 & 49.6\\
 \cmidrule{2-11}
 &  & R@1 & 16.3 & 11.6 & 13.1 & 10.0 & 16.8 & 14.8 & 13.7 & 11.8\\
 & InfoSeek & R@5 & 32.6 & 24.3 & 27.0 & 21.2 & 33.1 & 29.1 & 28.8 & 25.6\\
 &  & R@10 & 41.0 & 31.2 & 34.4 & 26.9 & 41.2 & 36.9 & 36.8 & 33.0\\
 \midrule
\multirow{3}{*}{-} &  & R@1 & 25.3 & 13.4 & 23.0 & 12.5 & 26.2 & 25.1 & 24.2 & 23.7\\
 & Average & R@5 & 46.5 & 32.8 & 42.6 & 31.2 & 47.7 & 45.6 & 44.5 & 42.7\\
 &  & R@10 & 47.5 & 34.9 & 44.0 & 33.5 & 48.4 & 46.4 & 45.6 & 43.8\\
\bottomrule
\end{tabular}%
}
\caption{Benchmarking information retrieval recall@1/5/10 on M-BEIR for CLIP Base models. For Fashion200K and FashionIQ, we report recall@10/20/50 following the original work.}
\label{tab:clip_base}
\end{table*}
\begin{table*}[]
\centering
\resizebox{\textwidth}{!}{%
\begin{tabular}{lll|cccc|cccc}
\toprule
\textbf{} &
  \textbf{} &
  \textbf{} &
  \multicolumn{4}{c}{Multi-task (\ding{55} instruction)} &
  \multicolumn{4}{c}{UniIR (\checkmark instruction)} \\ \cmidrule(l){4-7}\cmidrule(l){8-11} 
 &
   &
   &
  \multicolumn{2}{c}{BLIP$_{\text{SF}}$} &
  \multicolumn{2}{c}{BLIP$_{\text{FF}}$} &
  \multicolumn{2}{c}{BLIP$_{\text{SF}}$} &
  \multicolumn{2}{c}{BLIP$_{\text{FF}}$} \\ \cmidrule(l){4-5}\cmidrule(l){6-7} \cmidrule(l){8-9}\cmidrule(l){10-11} 
 &
  Dataset &
  Metric &
  \multicolumn{1}{l}{M-BEIR$_{\text{local}}$} &
  \multicolumn{1}{l}{M-BEIR} &
  \multicolumn{1}{l}{M-BEIR$_{\text{local}}$} &
  \multicolumn{1}{l}{M-BEIR} &
  \multicolumn{1}{l}{M-BEIR$_{\text{local}}$} &
  \multicolumn{1}{l}{M-BEIR} &
  \multicolumn{1}{l}{M-BEIR$_{\text{local}}$} &
  \multicolumn{1}{l}{M-BEIR} \\ 
  \midrule
\multirow{9}{*}{1. $q_t \to c_i$} &  & R@1 & 7.1 & 0.0 & 7.8 & 0.0 & 6.9 & 5.5 & 7.5 & 6.9\\
 & VisualNews & R@5 & 16.9 & 2.3 & 18.2 & 3.4 & 17.0 & 15.9 & 18.0 & 17.6\\
 &  & R@10 & 22.9 & 5.0 & 24.3 & 6.9 & 22.8 & 21.9 & 24.3 & 23.8\\
 \cmidrule{2-11}
 &  & R@1 & 47.3 & 0.0 & 48.1 & 0.0 & 47.8 & 11.3 & 49.5 & 23.8\\
 & MSCOCO & R@5 & 74.6 & 15.4 & 75.7 & 10.5 & 75.6 & 65.1 & 76.6 & 70.2\\
 &  & R@10 & 83.4 & 32.9 & 84.5 & 24.4 & 84.5 & 76.5 & 85.1 & 80.2\\
 \cmidrule{2-11}
 &  & R@10 & 22.1 & 3.5 & 22.3 & 0.9 & 21.6 & 21.2 & 22.8 & 22.1\\
 & Fashion200K & R@20 & 29.7 & 7.5 & 29.6 & 3.0 & 28.2 & 27.8 & 30.6 & 30.3\\
 &  & R@50 & 42.5 & 18.2 & 41.8 & 8.7 & 40.0 & 39.7 & 41.5 & 41.1\\
 \midrule
\multirow{3}{*}{2. $q_t 	\to c_t$} &  & R@1 & 50.1 & 47.9 & 49.2 & 46.9 & 51.4 & 51.2 & 50.1 & 49.7\\
 & WebQA & R@5 & 77.0 & 74.2 & 76.2 & 74.6 & 76.7 & 76.7 & 76.4 & 76.4\\
 &  & R@10 & 84.0 & 82.0 & 83.4 & 81.9 & 83.3 & 83.2 & 82.7 & 82.7\\
 \midrule
\multirow{6}{*}{3. $q_t 	\to$ ($c_i, c_t$)} &  & R@1 & 20.7 & 12.0 & 23.8 & 12.2 & 22.0 & 21.2 & 23.5 & 22.6\\
 & EDIS & R@5 & 40.2 & 30.0 & 47.5 & 30.4 & 44.5 & 44.1 & 46.2 & 45.9\\
 &  & R@10 & 50.1 & 38.7 & 57.7 & 38.9 & 53.9 & 53.7 & 56.7 & 56.4\\
 \cmidrule{2-11}
 &  & R@1 & 49.0 & 45.0 & 49.1 & 45.7 & 49.9 & 49.6 & 49.5 & 49.1\\
 & WebQA & R@5 & 77.3 & 73.5 & 78.1 & 75.0 & 77.5 & 77.2 & 78.1 & 77.9\\
 &  & R@10 & 86.6 & 83.4 & 85.9 & 82.6 & 86.4 & 86.2 & 85.6 & 85.5\\
 \midrule
\multirow{9}{*}{4. $q_i 	\to c_t$} &  & R@1 & 7.1 & 0.0 & 7.4 & 0.0 & 6.6 & 3.7 & 7.2 & 5.0\\
 & VisualNews & R@5 & 17.1 & 3.2 & 17.1 & 2.3 & 16.8 & 14.8 & 17.1 & 15.8\\
 &  & R@10 & 23.2 & 6.7 & 22.8 & 5.0 & 22.6 & 21.4 & 23.6 & 22.5\\
 \cmidrule{2-11}
 &  & R@1 & 58.3 & 0.0 & 63.7 & 0.0 & 62.4 & 39.2 & 64.6 & 51.7\\
 & MSCOCO & R@5 & 83.0 & 71.6 & 87.0 & 71.7 & 85.9 & 84.5 & 88.1 & 86.9\\
 &  & R@10 & 90.3 & 83.3 & 93.1 & 83.9 & 92.0 & 91.5 & 93.5 & 93.1\\
 \cmidrule{2-11}
 &  & R@10 & 22.4 & 1.6 & 23.6 & 0.9 & 20.9 & 18.2 & 23.8 & 22.4\\
 & Fashion200K & R@20 & 30.4 & 4.0 & 32.0 & 2.4 & 29.7 & 27.7 & 32.7 & 31.5\\
 &  & R@50 & 43.8 & 10.2 & 44.7 & 6.8 & 42.3 & 40.6 & 45.6 & 45.1\\
 \midrule
\multirow{3}{*}{5. $q_i 	\to c_i$} &  & R@1 & 8.2 & 8.2 & 8.0 & 8.0 & 7.7 & 7.7 & 7.7 & 7.7\\
 & NIGHTS & R@5 & 30.0 & 29.7 & 31.8 & 31.6 & 30.7 & 30.6 & 30.3 & 30.3\\
 &  & R@10 & 49.8 & 49.1 & 51.0 & 50.5 & 51.2 & 51.2 & 49.5 & 49.5\\
 \midrule
\multirow{6}{*}{6. ($q_i, q_t$) $	\to c_t$} &  & R@1 & 17.6 & 16.9 & 19.6 & 20.4 & 13.9 & 16.0 & 17.9 & 20.8\\
 & OVEN & R@5 & 33.2 & 29.2 & 36.7 & 32.8 & 28.9 & 28.2 & 34.5 & 33.1\\
 &  & R@10 & 40.7 & 35.2 & 44.4 & 38.9 & 36.1 & 34.1 & 42.1 & 39.0\\
 \cmidrule{2-11}
 &  & R@1 & 7.6 & 3.7 & 8.6 & 5.6 & 6.3 & 5.0 & 7.9 & 6.9\\
 & InfoSeek & R@5 & 17.3 & 9.5 & 21.2 & 13.6 & 16.5 & 13.3 & 18.9 & 16.5\\
 &  & R@10 & 23.6 & 14.0 & 28.1 & 19.3 & 23.3 & 19.6 & 25.4 & 22.5\\
 \midrule
\multirow{6}{*}{7. ($q_i, q_t$) $	\to c_i$} &  & R@10 & 22.5 & 22.1 & 25.4 & 24.9 & 20.8 & 20.1 & 23.7 & 23.0\\
 & FashionIQ & R@20 & 29.9 & 29.2 & 33.2 & 32.5 & 27.5 & 26.6 & 31.0 & 29.8\\
 &  & R@50 & 41.2 & 40.3 & 44.8 & 43.9 & 38.2 & 36.6 & 42.1 & 40.5\\
 \cmidrule{2-11}
 &  & R@1 & 11.0 & 9.1 & 20.3 & 18.7 & 13.1 & 12.9 & 20.2 & 19.4\\
 & CIRR & R@5 & 39.0 & 31.8 & 45.1 & 42.0 & 42.2 & 40.7 & 46.4 & 45.1\\
 &  & R@10 & 50.3 & 42.7 & 56.0 & 52.2 & 54.4 & 52.3 & 57.5 & 55.8\\
 \midrule
\multirow{6}{*}{8. ($q_i, q_t$) $	\to$ ($c_i, c_t$)} &  & R@1 & 28.3 & 32.5 & 31.1 & 33.2 & 28.3 & 33.0 & 29.1 & 37.1\\
 & OVEN & R@5 & 47.0 & 46.8 & 49.9 & 47.5 & 47.3 & 48.2 & 48.2 & 50.6\\
 &  & R@10 & 54.6 & 52.5 & 57.5 & 53.1 & 54.7 & 54.2 & 55.6 & 55.9\\
 \cmidrule{2-11}
 &  & R@1 & 11.2 & 9.4 & 13.7 & 10.9 & 12.2 & 10.8 & 13.0 & 11.8\\
 & InfoSeek & R@5 & 25.5 & 20.7 & 29.1 & 22.7 & 26.4 & 23.6 & 27.0 & 23.5\\
 &  & R@10 & 33.6 & 27.9 & 37.3 & 29.4 & 34.2 & 30.7 & 35.0 & 30.4\\
 \midrule
\multirow{3}{*}{-} &  & R@1 & 24.9 & 14.2 & 26.9 & 15.5 & 25.3 & 20.6 & 26.8 & 24.1\\
 & Average & R@5 & 44.5 & 33.7 & 47.2 & 35.2 & 45.1 & 43.3 & 46.6 & 45.4\\
 &  & R@10 & 47.5 & 36.3 & 49.8 & 37.1 & 47.7 & 46.0 & 49.2 & 47.8\\
 \bottomrule
\end{tabular}%
}
\caption{Benchmarking information retrieval recall@1/5/10 on M-BEIR for BLIP Base models. For Fashion200K and FashionIQ, we report recall@10/20/50 following the original work.}
\label{tab:blip_base}
\end{table*}

\end{document}